\documentclass{article}
\usepackage{arxiv}
\usepackage[numbers]{natbib}

\usepackage{times}
\usepackage{amsmath,amsthm,amssymb}

\usepackage{graphicx}
\usepackage{booktabs}
\usepackage[cm]{fullpage}
\usepackage{enumitem}
\usepackage[utf8]{inputenc}
\usepackage{textgreek}
\usepackage[colaction]{multicol}
\usepackage{comment}
\usepackage[T1]{fontenc}
\usepackage[dvipsnames]{xcolor}
\usepackage{subcaption}
\usepackage{wasysym}
\usepackage{float}
\usepackage[hypertexnames=false]{hyperref}
\usepackage{spverbatim}

\newcommand{\x}{\boldsymbol{x}}
\newcommand{\ba}{\boldsymbol{a}}

\newcommand{\bx}{\boldsymbol{x}}

\title{\bf Insightful analysis of historical sources at scales beyond human capabilities using unsupervised Machine Learning and XAI}

\newcommand{\correspondingauthor}{\textasteriskcentered}
\author{
	Oliver Eberle\textsuperscript{\textnormal{1,2}},
	Jochen Büttner\textsuperscript{\textnormal{2,3}}, 
	Hassan El-Hajj\textsuperscript{\textnormal{2,3}}, 
	Grégoire Montavon\textsuperscript{\textnormal{4,2,1}},\\
	Klaus-Robert Müller\textsuperscript{\textnormal{1,2,5,6,\correspondingauthor}},
	Matteo Valleriani\textsuperscript{\textnormal{1,2,3,7,8,\correspondingauthor}} 
}

\date{\normalsize
$^1$ Machine Learning Group, Technische Universit\"at Berlin, Marchstr. 23, 10587 Berlin, Germany \\
$^2$ BIFOLD -- Berlin Institute for the Foundations of Learning and Data, 10587 Berlin, Germany \\
$^3$ Max Planck Institute for the History of Science, Boltzmannstr. 22, 14195 Berlin, Germany \\
$^4$ Department of Mathematics and Computer Science, Freie Universit\"at Berlin, Arnimallee 14, 14195 Berlin, Germany \\
$^5$ Department of Artificial Intelligence, Korea University, Seoul 136-713, South Korea \\
$^6$ Max Planck Institute for Informatics, Stuhlsatzenhausweg 4, 66123 Saarbrücken, Germany\\
$^7$ Institute of History and Philosophy of Science, Technology, and Literature, Faculty I - Humanities and Educational Sciences, Technische Universit\"at Berlin, Straße des 17. Juni 135, 10623 Berlin, Germany\\
$^8$ The Cohn Institute for the History and Philosophy of Science and Ideas, Faculty of Humanities, Tel Aviv University, P.O.B. 39040, Ramat Aviv, Tel Aviv 6139001, Israel\\
$^\text{\textasteriskcentered}$ To whom correspondence should be addressed: valleriani@mpiwg-berlin.mpg.de, klaus-robert.mueller@tu-berlin.de. \\
}

\begin{document} 

\maketitle

\begin{abstract}
\normalsize
Historical materials are abundant. Yet, piecing together how human knowledge has evolved and spread both diachronically and synchronically remains a challenge that can so far only be very selectively addressed. The vast volume of materials precludes comprehensive studies, given the restricted number of human specialists. However, as large amounts of historical materials are now available in digital form there is a promising opportunity for AI-assisted historical analysis.
In this work, we take a pivotal step towards analyzing vast historical corpora by employing innovative machine learning (ML) techniques, enabling in-depth historical insights on a grand scale. Our study centers on the evolution of knowledge within the `Sacrobosco Collection' -- a digitized collection of 359 early modern printed editions of textbooks on astronomy used at European universities between 1472 and 1650 -- roughly 76,000 pages, many of which contain astronomic, computational tables. An ML based analysis of these tables helps to unveil important facets of the spatio-temporal evolution of knowledge and innovation in the field of mathematical astronomy in the period, as taught at European universities.
\end{abstract}

\smallskip

\begin{multicols}{2}

\section{Introduction}

When investigating the early modern period, traditional history of science mainly focused on what is commonly termed the Scientific Revolution. This is frequently portrayed as a cumulative sequence pieced together from singular events, most of which are associated with the publication of significant works by heroic figures. A prime example of such a narrative is the lineage from Nikolaus Copernicus via Galileo Galilei and Johannes Kepler to Isaac Newton, which is often seen as quintessentially capturing the nature of the revolution in astronomy during this period \cite{Kyore1939,Kyore1973,Kyore1957,Pedersen1993,Westfall1971,Cohen2015,Kuhn1957}.\par

An alternative to this traditional approach is a history of science that delves into a broader range of historical sources to more comprehensively grasp the intellectual context within which these celebrated "heroes" of science worked and produced their intellectual insights. Thomas Kuhn's influential \emph{The Structure of Scientific Revolutions} of 1962 marked a pivotal redirection in this respect: emphasizing the  role of scientific paradigms, it shifts from spotlighting individual contributors to viewing scientific progress as a collective achievement of the wider scientific community \cite{Kuhn1962}. Today, such a perspective has evolved even further. History of science more broadly perceived as a "history of knowledge" intends to harness every conceivable historical source that might offer insights \cite{Burke2015,Daston2017,OestlingLarsson2020}. However, a significant, practical limitation obstructs such endeavor: The sheer volume of available  sources  surpasses our current capacity to accomplish historical investigation. In the following, we suggest an approach based on Machine Learning (ML) and Explainable Artificial Intelligence (XAI) techniques, conceived to overcome this limit. \par

In this study we focus focus on the core knowledge of the period, i.e., the set of widely accepted theories, methods, and results. A prime source for reconstructing this broader core knowledge are university textbooks, which informed the broader student population and \emph{intelligentia} \cite{RN2956}. Historians have previously shown interest in textbooks \cite{LundBensVin2000,Vicedo2012}. However, a comprehensive analysis has remained elusive due to the great amount of available material. Our research is uniquely poised in this context, as we leverage the "Sacrobosco Collection"\cite{SphaeraAuthors,VallerianiOttone2022,mva19,SRN2020,Zamani2023} (Supplementary Note \ref{text:suppl_sacro_collection} in Section Materials and Methods). This very large and significant collection encompasses textbooks introducing geocentric astronomy to students across Europe from the final quarter of the 15th century up to 1650.\par

The collection contains approximately 76,000 pages of scientific content from 359 editions of different textbooks, which were published starting in 1472, the year of the first print (and of the first ever print of a scientific, mathematical text). The year 1650, on the other hand, marks the end of the slow decline of geocentric astronomy initiated almost a 100 years  earlier by Nikolaus Copernicus  who  in his \textit{De revolutionibus orbium coelestium} of 1543 introduced a mathematical system based on a heliocentric worldview. For each edition in the corpus only one exemplar has been collected and considered as representative of the entire print-run. Accepting the current view according to which academic textbooks on mathematical subject were printed at the time with an average print-run of ca. 1000 copies, the Sacrobosco Collection thus can be considered as representative for about 350.000 textbooks that were circulating and used in Europe during the period considered \cite{Gingerich1988,Gingerich1990}.\par

Our study specifically addresses the mathematical education and culture possessed by students and the educated populace (i.e., the potential readers). The impact of cutting-edge innovations in mathematical astronomy hinged significantly on their reception and comprehension by a broader audience. As a case in point, Copernicus's work remained largely overlooked for an extended period \cite{Gingerich2004}. To discern if this neglect stemmed from challenges in grasping its mathematical underpinnings, we must ascertain the scope and depth of mathematical knowledge prevalent in society at large. This entails understanding where this knowledge originated, the motivations behind its dissemination, and the modes of its circulation. The present study introduces a new method to enable the historical analysis of the mathematical education in astronomy all over Europe and its transformation during the ca. 180 years considered, while the question as to whether Copernicus's work was neglected because of the characteristics of the mathematical education of the time will be investigated in further studies.\par

A central element of the mathematical apparatus of early modern astronomy are computational astronomic tables. Such tables can be understood as the sequential representation of input and output values of mathematical relations akin to equations. Yet, the formulaic algebraic language was only beginning to be used towards the end of period considered.  Before that, the meaning of the mathematical relations represented by tables was described in  the associated texts \cite{ChabasGoldstein2003,ChabasGoldstein2012}.\par

To investigate astronomical tables one needs a method to identify the corresponding content in the historical material, to group the tables according to a semantically meaningful similarity (Supplementary note \ref{supplement:sec:scaling_tables}), and finally to analyze the dynamics of their development throughout space and time. As it turns out,  approximately 10,000 pages of the Sacrobosco Collection feature computational tables rendering a  standard historical analysis based on close reading practically impossible. In this work, we introduce an approach that employs ML and XAI to assist historians in analyzing early modern computational numerical tables on an unprecedented scale. Furthermore, we argue that this approach can be adapted to other types of sources besides numerical tables as well, such as visual or textual elements.\par 

In recent years, ML and specifically deep learning has established itself as a key enabler in industry and the sciences for efficient and insightful exploration of large corpora of structured or unstructured data (cf.~\cite{lecun2015deep,hochreiter1997long, schmidhuber2015deep, radford2019language}). This has led to unprecedented progress in technical disciplines such as speech recognition \cite{hinton2012deep, Graves2013SpeechRW, speechrecog_chiu_2018, ott2019fairseq}, natural language processing \cite{vaswani2017attention,devlin-etal-2019-bert,radford2019language,brown2020language, lambda2022}, control and planning \cite{mnih-atari-2013, mnih2015humanlevel, alphago_2016, DBLP:journals/corr/LillicrapHPHETS15, won2020adaptive}, and computer vision \cite{lecun98, yolo2016, resnet2016}, as well as in the sciences and medicine where novel insights could be gained, e.g.~\cite{Baldi2014,Schuett2017,doi:10.1021/acs.chemrev.1c00107, Reichstein2019DeepLA, samek2021explaining,binder2021morphological, Jumper2021HighlyAP}. All of these disciplines can harvest large collections of well-structured digitized data that have become available in the respective fields.\par

In the context of the digital humanities, deep learning is being used increasingly to process data and generate insights from historical corpora. The relevance of this approach is growing, especially in the field of historical document analysis alongside the proliferation of well-curated image datasets and benchmarks of historical sources \cite{Papadopoulos2013, Fischer2020, Nikolaidou2022, Buttner2022, grasshoff2021kepler}. In particular, the availability of such datasets encouraged the usage of neural networks, such as U-Net \cite{journals/corr/RonnebergerFB15}, YOLO \cite{yolo2016}, Faster R-CNN \cite{FasterRCNN2015},  to extract relevant visual elements (e.g., illustrations, drawings, images, etc.) from large corpora, using them as proxy for understanding their accompanying texts \cite{monnier2020docExtractor, Abhishek2021, Buttner2022}. When it comes to text, deep learning approaches based on Recurrent Neural Networks (RNN) \cite{Tsochatzidis2021, Fischer2020}, and more recently Transformer-based architectures \cite{Wick2021, Li2021TrOCR, strobel2022transformerbased} have been developed for Optical Character Recognition (OCR) and Handwritten Text Recognition (HTR). Multimodal approaches have further enabled the exploration of large document datasets using both language and image modalities \cite{Smits2023}.\par

Beyond mere data exploration and extraction, \cite{Assael2019} proposed a sequence-to-sequence RNN to reconstruct ancient Greek inscriptions, which was later followed by a Transformer-based architecture \cite{AssaelEtAlNature2022} to not only restore ancient Greek inscriptions, but also generate local insights about their provenance and dating. Other `ancient' languages also benefited from deep learning approaches, such as Latin \cite{Bamman2020}, Akkadian \cite{Fetaya2020}, and Hieroglyphs \cite{Barucci2021}.

To obtain trustworthy and reliable scientific insights within the digital humanities, explainable artificial intelligence allows to validate results of ML models \cite{Pawlowicz2021} and to further generate insights into humanities datasets \cite{Bell2021, ElHajjEberle_xaidh_2023}. \par

From an ML perspective, the analysis of historical data presents very unique challenges. Previous works have often relied on readily available pre-trained models and large amounts of annotated material, this scenario is typically not applicable to historical data collections, especially with regard to labels of interest to historians such as detailed semantic connections; a scenario that mostly occurs because of the unreasonable requirement for human resources. 

In addition, historical data is typically characterized by extensive heterogeneity and non-stationarity \cite{sugiyama2012machine}, and an overall lack of annotations (Supplementary Note \ref{supplement:sec:heterogeneity} and \ref{supplement:sec:ML_limits} in Section Materials and Method).\par

The historical sources analyzed in this work come from  different times and from different places and were frequently produced following very different standards. With regard to the printed book, the type source from which the tables analyzed in this work are extracted, the heterogeneity is further increased by the intertwined effects induced by processes of scientific knowledge transformation, development of printing technology, and academic book market mechanisms \cite{Gilbert1995, Eisenstein1996, MacLean2009, Nuovo2013, VallerianiOttone2022}, contributing differently to the diverse sources of data variability (Supplementary Note \ref{supplement:sec:heterogeneity} and \ref{supplement:sec:limit_OCR} in Section Materials and Method).\par

These general challenges are further accompanied by specific characteristics of the selected material to be analyzed. In the case of astronomic tables, assessing their complex similarity structure poses challenges for both trained historians and conventional ML approaches, encompassing end-to-end training and the utilization of pre-trained models.

In the case of individual source analysis executed by historians, the required similarity assessments are unfeasible at scale (Supplementary Note \ref{supplement:sec:scaling_tables} in Section Materials and Method), and conventional ML approaches are unfeasible due to the lack of labeled material combined with high data heterogeneity (Supplementary Note \ref{supplement:sec:standard_approaches_hetereogeneity} in Section Materials and Method).

With this work, we address these challenges within a novel `atomization-recomposition' approach, which we intend as a general ML framework in unsupervised settings when only limited and sparse annotations are available as described in Supplementary Note \ref{sec:atomization_recomposition} in Section Materials and Method. \par

We demonstrate this approach by decomposing complex table-page information to enable our ML model to discover semantic similarities between heterogeneous tables with highly variable mathematical content. After validation of the obtained representations using both nominal accuracies and XAI, we extend our analysis to the corpus level. By leveraging the similarity structure of the entire material, its full potential is realized, enabling previously inaccessible historical investigations.
The examination of the geo-temporal evolution of the computational tables provides insights into the widespread diffusion of mathematical education and culture in the frame of astronomy that otherwise remains hidden behind an enormous amount of hitherto inaccessible computational tables. \par

Our approach allows not only for a systematic extraction of data-driven insights in large corpora but it also provides an example for the quantification of historical processes at scale. It thus aids in making more informed selections of historical source material which can then be analyzed using conventional methods of historical inquiry. The presented historical analysis of early modern mathematization thus provides an example of how historical disciplines can benefit from ML and XAI methodologies, which can also assist and elevate the close-reading analysis of individual sources. \par

\section{Results}
\subsection{Representation of historical material via atomization-recomposition.}

\begin{figure*}[t!]
\centering
\includegraphics[width=\textwidth]{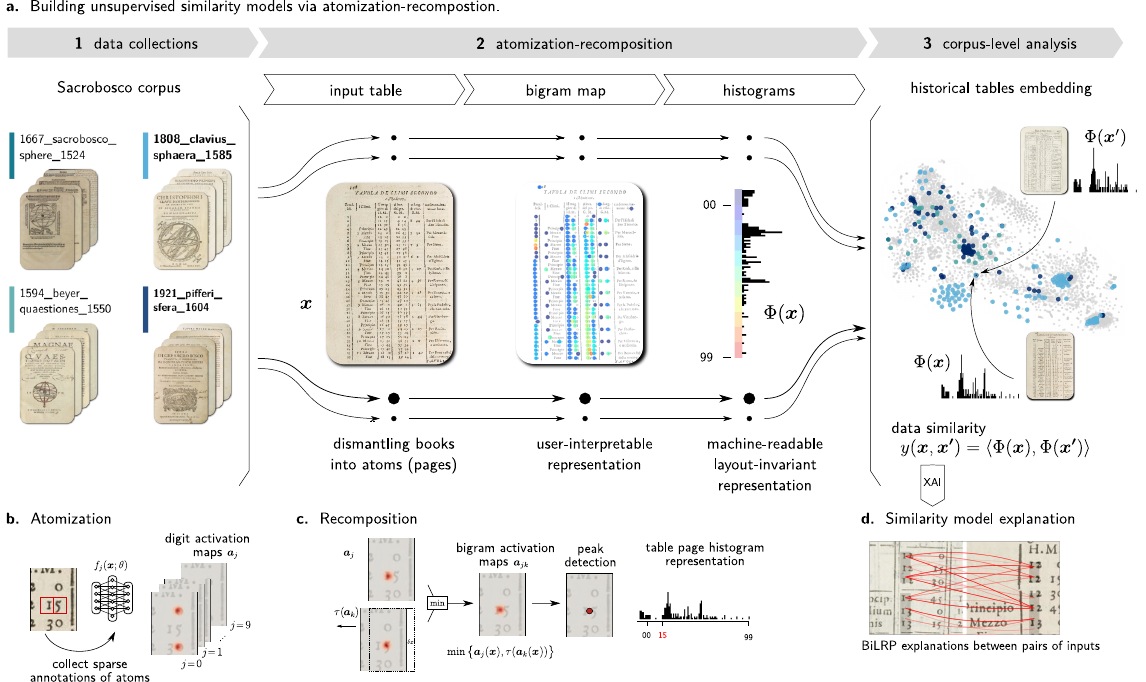}
\caption{\textbf{Atomization-recomposition  framework for model learning under sparse annotation settings.} 
(\textbf{a}) Overall computational workflow starting with an unstructured collection of books (Sacrobosco Collection), atomizing them into tables and single digits that a ML model can detect, recomposing them into user-interpretable bigrams, and generating histograms that enable dataset-wide unsupervised ML-based analyses.
(\textbf{b}) A few hundred sparse single-digit annotations are used to train a digit recognition model which activates where digits are found in the input image. 
(\textbf{c}) The resulting digit activation maps are recomposed into more complex, task-specific representations, here, numerical bigrams, and whole-page histograms.
(\textbf{d}) The similarity scores on which ML-based analyses operate are verified via XAI, specifically the BiLRP technique \cite{eberle2020}, which highlights how the similarity scores arise from the pixel representation.
}
\label{fig:overview_approach}
\end{figure*}

We consider tables as collections of table pages, and these as a collection of numbers, and the numbers themselves as sequences of digits and these finally as a collections of digits. Concordant with this scheme, we built the  \textit{Sacrobosco Tables} corpus, which consists of pages that contain tables with at least one numerical column (Supplementary Note \ref{supplement:subsec:page_classifier} in Section Materials and Method). Our atomization-recomposition approach utilizes this compositional structure. 
The initial atomization step yields a collection of individual digits (0-9) with heterogeneous fonts, print quality, and spatial location. These digits are the most basic building blocks essential to describing the semantics of the tables as shown in Figure \ref{fig:overview_approach}-b. Thereby, we reduce the ML model complexity to that of a single digit recognition model, which can be learned efficiently by collecting only a few hundred labeled digit patches. Each table page $\bx$ can subsequently be passed to the learned ML model, leading to activation maps $\ba_j(\x)$ associated to each digit, with $j$ from $0$ to $9$. \par
\begin{figure*}[t!]
\centering
\includegraphics[width=\textwidth]{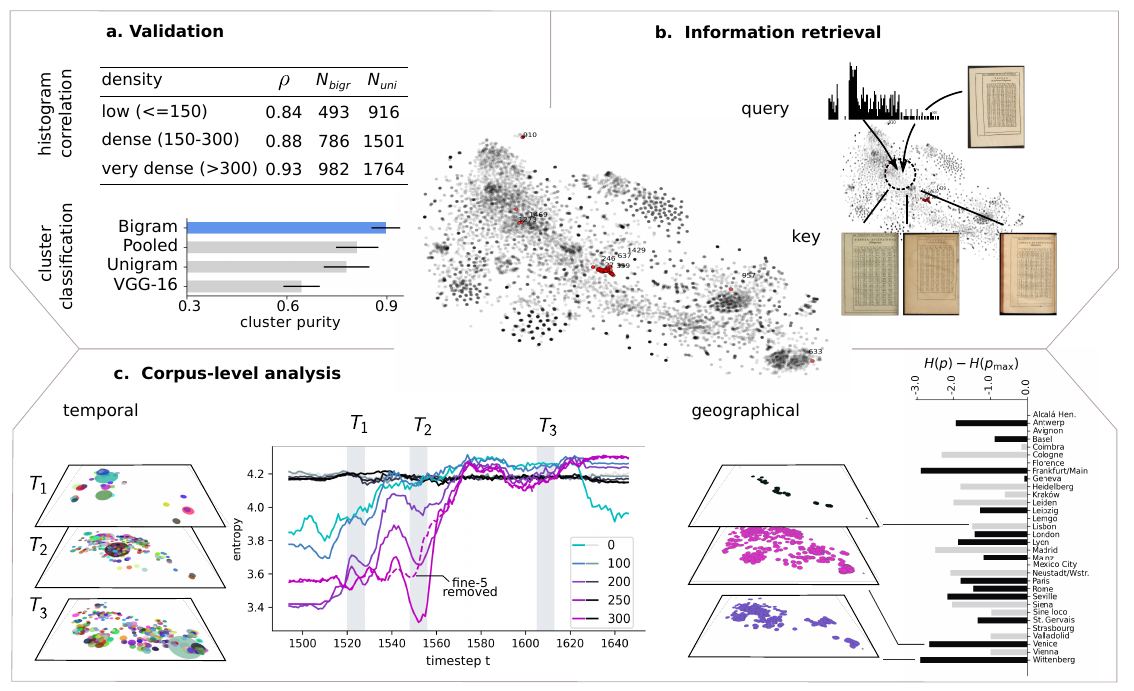}
\caption{\textbf{Extracting historical insights from bigram histograms.}  T-SNE visualization (\textit{center}) of the corpus. A set of hand labeled, semantically identical tables providing the position of the Sun against the zodiac over the course of the year is shown in red.  After performing a $k$-means clustering on the extracted numerical histograms, we show the $k$-means clusters that contain members of this ground truth group (marked by their cluster-id).
\textbf{a}. Validation of different table representations and Pearson correlation scores for different digit densities (number of digits per table page).
\textbf{b}. By providing query histograms or reference pages our approach is able to generate a set of key candidates of tables that are identical or very similar to a given query table.
\textbf{c}. Left and center: Temporal evolution of knowledge displayed by computing the entropy of cluster membership vectors (number of tables in each cluster) for each time step. Gray to black lines correspond to a random embedding baseline, colored lines correspond to the data from our corpus. Different colors indicate a filtering threshold on the digit density per page, i.e. all pages containing at least 100 digits. The clusters are shown as t-SNE visualization for three time intervals indicating active clusters and cluster disk diameter is proportional to cluster size. We observe a marked drop in entropy for tables with extensive numerical content between 1540 and 1560. This drop disappears after removal of the \textit{fine-5} group, a subset of tables that occur in Finé's editions that we identified as the dominant factor driving the entropy change. Right: Geographical analysis of knowledge distribution for each print location in alphabetical order using relative entropy. Low-output cities (\textless=100 tables) are colored in light gray. For three selected cities t-SNE visualization of the distribution of the printed tables is provided.
}
\label{fig:results_plot}
\end{figure*}

In the subsequent recomposition step, a sequence of non-trainable layers are applied to compute increasingly task-specific features. First, we generate bigram activation maps $\ba_{jk}$ as,
\begin{align*}
\ba_{jk}(\x) &= \min\big\{\ba_j(\bx) ,\tau(\ba_k(\bx))\big\},
\end{align*}
with bigrams $jk$ from $00, 01,...,99$, and  $\tau$ being a spatial translation shifting activation maps by a fixed number of pixels as shown in Figure \ref{fig:overview_approach}-c. In addition, we also include isolated single digit numbers into the representation via an extension of this approach (Supplementary Material \ref{supplement:recomposition} in Section Materials and Method). Besides the clear advantage of only having to provide sufficient single digit labels to ensure their robust detection, this approach allows features to be detected that do not occur in the training data. For example, the bigram `25' could be detected on test pages even when the training pages contained only bigrams `12' and `51'. As shown in Figure \ref{fig:overview_approach}-a, the recomposed feature maps additionally provide a suitable interface for a human expert to inspect the inner workings of the ML model and to gain further confidence in its predictions. A second stage of recomposition via spatial pooling then converts this human-readable map representation into a lower-dimensional bag-of-bigrams histogram that is invariant to the exact table layout. We validate the resulting histograms using a diverse subset of fully annotated table pages (Supplementary Note \ref{supplement:sec:fully_annotated} in Section Materials and Method) and achieve average Pearson correlation scores from 0.84 for tables of low digit density to 0.93 for high density tables as shown in Figure \ref{fig:results_plot}-a. Furthermore, we assess the performance of different table page representations to identify clusters of identical table pages. We find that our proposed bigram representation is most effective for retrieving correct cluster members when compared to a direct pooling of bigram activations (pooled), single digit summaries (unigram), or a pre-trained deep neural network representation from VGG-16 (see Figure \ref{fig:results_plot}-a). In addition, explanation techniques are provided that help the user understand why the ML algorithms arrive at a certain similarity assessment for a pair of tables \cite{eberle2020} as shown in Figure \ref{fig:overview_approach}-d (Supplementary Note \ref{supplement:sec:xai} in Section Materials and Method). While we have clearly focused on numerical tables, we emphasize that the similarity of other aspects of historical documents can be readily learned by an analogous extension of our framework. \par

\subsection{Corpus-Level Historical Insights and Case Studies}\par
Our approach allows a) for historical investigations on a general, corpus level as it makes it possible to trace and analyze the geotemporal evolution of the computational tables in the entire corpus and, b), for the identification of particularly interesting clusters of similar tables thus guiding an informed selection of specific case studies, which are ultimately analyzed through standard close-reading. In the following, the results are described of the corpus-level analysis as well as of the identification and investigations concerned with two relevant, mutually interconnected case studies.\par

On a corpus level, we demonstrate that the process of mathematization of the astronomy codified in textbooks and taught at the European universities, occurred alongside a process of acceleration of diffusion of mathematical knowledge that took place during the last decades of the 16th century. This acceleration was ignited and fueled mainly by the competition between two key entities: the French Royal Chair of Mathematics and the \textit{Collegio Romano}, the principal mathematical division within the Jesuit order \cite{Grendler2022} (Supplementary Notes \ref{supplement:text:Histo_TSNE} and  \ref{supplement:text:CorLev_TSNE} in Section Supplementary Text). Spreading mathematical knowledge was among the main goals of  both institutions.\par

This process exhibits a non-linear dynamic that, on closer inspection, turns out to be caused by the necessity to adhere to early modern marketing rules for academic prints. These rules required the rapid introduction of scientific works in various formats to the market, with multiple editions of each work released in close temporal proximity to one another \cite{VallerianiOttoneChap12022,Maclean2022,RN2923}. The most significant episodes of such high frequency publication and republication occurred  within a time frame of five years around 1550 and involved Oronce Finé, the French Royal mathematician at that time (Figure \ref{fig:results_plot}-c) \cite{Axworthy2020} (Supplementary Note \ref{supplement:text:history_temporal} in Section Supplementary Text).\par

The accelerated circulation of mathematical knowledge represented in the corpus of textbooks ultimately led to  a process of homogenization, which means that scientific works were increasingly offering the same mathematical approaches.
By measuring the entropy of cluster membership vectors that represent the number of table pages in each cluster, we show which places of print production contributed to this phenomenon most and which did so to a lesser extent. We demonstrate that mathematical knowledge presented in treatises produced in post-Reformation Wittenberg is particularly homogeneous, presumably due to political control over scientific education \cite{LimbachForth,RN2847,Jackson2013}. On the other hand, of the spectrum treatises from Venice display a variety of scientific approaches, a characteristic that aligns with the central international economic position of Venice's printing industry serving a variety of local markets (Supplementary Note \ref{supplement:text:history_spatial} in Section Supplementary Text).\par

The insights from the corpus-level analysis reveal instances where the process dynamics deviate from established trends. This puts us in position to make informed decisions about specific case studies (Supplementary Note \ref{supplement:text:case_studies} in Section Supplementary Text). To facilitate such studies we have  provided a tool to identify clusters of tables identical and similar to one selected by a domain expert. (For more information, see Supplementary Note \ref{supplement:text:case_studies} in Section Supplementary Text). Classifying closely related tables can also enable the automatic identification of various mathematical approaches to the same topic, with each approach represented by a distinct cluster. In this context, a cluster encapsulates all the necessary materials for a comprehensive historical case study, encompassing the full spectrum of available sources. As a result, clustering facilitates an in-depth exploration of a particular phenomenon across its entire evolutionary trajectory.\par

Two case studies were identified and conducted along the line described above: one dedicated to the method for geometrically subdividing the Earth's surface from the equator to the poles based on the length of the Solar day and, the second, concerned with the calculation workflow necessary to retroactively predict  the position of the Sun on the Zodiac during classical antiquity (Supplementary Note \ref{supplement:text:case_study_climate} and \ref{supplement:text:case_study_zodiac} in Section Supplementary Text, with individual examples). These two case studies, considered together,  allow us to formulate a hypothesis as to how the acceleration of diffusion of mathematical knowledge and the resulting increase of homogenization of scientific knowledge (always referring only to astronomy as taught at the universities) were  interwoven with the process of formation of a European scientific identity (Section \ref{FinalDiscussion}).  \par

Since antiquity the known world was considered as divided into inhabitable and habitable zones. The inhabitable were not considered entirely devoid of people but generally held inhabitable because of the hard life conditions. The habitable zone, covering roughly the longitudinal area of Europe and extending from North Africa northwards  to include Paris, had been traditionally divided into seven `climate zones' since antiquity. A climate zone (land strips parallel to the equator) was defined based on the length of the solar day in those areas on  the summer solstice. This conception was fundamental in a variety of scientific disciplines, such as medicine, and continued to be taught until at least the mid-17th  century \cite{RN1938}. Clearly, however, the early modern journeys of explorations had exposed that this ancient conception of the habitable zone was too limited \cite{RN820}. This situation is reflected in the sources under consideration, which display two different types of climate zone tables: one for seven zones and another that encompasses the entire planetary surface from the equator up to the polar circle and thus conceptualizing 24 zones.\par

Our approach has yielded a series of new insights regarding the concept of the climate zones and its development attainable only by comparing a large number of relevant tables. First of all, we were able to track the dissemination of the pertinent knowledge in detail over the 180 years under consideration. We discovered that the diffusion of the modern conception of 24 zones was surprisingly not detrimental to the ancient one, contrary to what one might expect. (Supplementary Material \ref{supp:link:climate_zones} in Section Supplementary Text, Chains 1 and 3). Rather the opposite is the case: The success of the innovation was, in fact, largely dependent on its link to the traditional, ancient, and authoritative concept and eventually worldview. The peak in the dissemination of the table representing the new conception can primarily be attributed to editions that also included the old table listing the traditional seven zones. Secondly, by accurately assessing the similarity within the subgroup of the relevant 225 pages of tables, our approach enabled the identification of a third variant of climate zone tables. This variant initially expanded the old view, but only to the extent of incorporating European regions at higher northern latitudes, specifically including Wittenberg by adding two zones (a video link for the spread of the climate zone tables can be found in Supplementary Material \ref{supp:link:climate_zones} in Section Supplementary Text). In fact, even though the dissemination of this conception of nine zones remained limited in both time and space, it represented the first significant break from the traditional view. 

The second case study focuses on a scientific specialization, no longer extant, that closely connected mathematical astronomy and history. Starting from the 13th century, when Europeans created the epochal subdivision between antiquity, the Middle Ages, and the new epoch in which they were living, frantic activity began that aimed to reconstruct an exact chronology of  ancient events \cite{Burke1969,Tanaka2019}. This was because, from the perspective of the day, antiquity represented the epoch during which the pinnacle of civilization and knowledge had been reached. In antiquity, the connection between the calendar and the Sun's position within the signs of the Zodiac was already well-established. As a result, by providing the positional values for the Sun, it was possible to calculate the specific day, and vice versa. Consequently, in ancient Greek and Latin works, descriptions of events are often accompanied by specific astronomical observations that can be linked to the position of the Sun in the Zodiac.\par

After Philipp Melanchthon, one of the founding fathers of the Protestant Reformation, had urged young students to study astronomy in 1531 and 1538, warning that without it the history of humanity would be mere chaos  \cite{RN1480,MVBFON2022}, a particular scientific specialization emerged.  This specialization, which aimed to provide precise dates for ancient events, endured until the 19th century, particularly in German universities. Mathematically, the required calculations were challenging both because of the historical changes of the calendar systems and the precession of the equinoxes, which itself was not yet fully understood in the 16th century \cite{Pantin2020}. Also in this case our approach provided us with the necessary selection of the material which allowed us to investigate the first steps of a broad phenomenon of diffusion of mathematical culture in the framework of the teaching of astronomy at the universities. \par

First of all, we have been able to establish that the values of the position of the Sun against the ecliptic were transposed into a handy table for the students for the first time in 1543, and also to show that this table was printed and used only in Northern Germany and France (Supplementary Material \ref{supp:link:sun_zodiac} in Section Supplementary Text contains a video link for the spread of the so called \emph{nostro} tables).
Second, and more relevantly, we were able to identify another table, which essentially provides the same information but pertains to ancient times. To communicate this information, a new table is indeed required since the position of the Sun relative to the zodiacal signs for a given date changes. While the annual change is minimal, the change accumulates to a noticeable difference if longer time periods are considered. This similar but not identical table therefore serves to directly display  the position of the Sun as it was observed by the ancient writers. This table was first conceived in Wittenberg and was created to simplify the calculations otherwise required to convert the current (of the 16th century) position of the Sun into the ancient position, which was necessary to establish a connection to the calendar. It spread, however, only  in Northern Germany 
(Supplementary Material \ref{supp:link:sun_zodiac} in Section Supplementary Text contains a link to a video visualizing the spread of the  sun-zodiac table for the ancient authors (so called \emph{veterum} table)). 
\par

\begin{figure*}[t!]
\centering
\includegraphics[width=\textwidth]{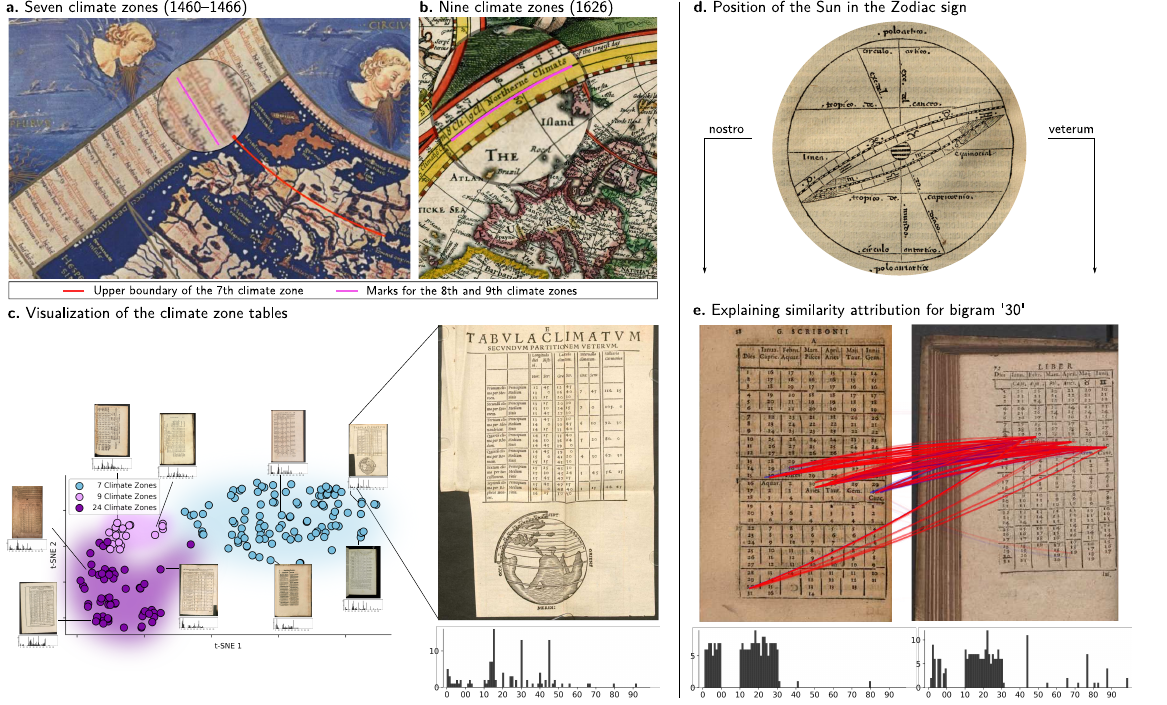}
\caption{\textbf{Historical case studies.} (\textbf{a}) Worldmap as conceived in the Hellenistic era by Ptolemy and drawn for the first time during the 15th century by following the list of coordinates and the metric of Ptolemy. The 7th climate zone clearly excludes all regions north of Paris, including current Great Britain. From: Ptolemy, \textit{Cosmographia}. Map maker: Nicolaus Germanus. Ms. membr., lat., sec. XV, cc. I–II, 124, III–IV. 1460–1466. Biblioteca Nazionale di Napoli.
(\textbf{b}) Robert Walton's Worldmap drawn in 1626. It includes all recently discovered territories on the Earth but considers only nine climate zones as worth being explicitly mentioned. The 9th climate zone includes England but was originally introduced to include Wittenberg. Further zones toward North are only generically mentioned. From: \textit{A New and Accurat Map of the World Drawne according to ye truest Descriptions lastest Discoveries \& best observations yt have beene made by English or Strangers}, 1626. London 1627. The Barry Lawrence Ruderman Map Collection. Courtesy Stanford University Libraries. \protect\hyperlink{http://purl.stanford.edu/cc815fz9830}{http://purl.stanford.edu/cc815fz9830}
(\textbf{c}) T-SNE visualization of the climate zone table histograms colored according to the number of climate zones they consider. (\textbf{d}) Illustration displaying the orbit of the Sun (ecliptic) on the Zodiac subdivided into the twelve signs. From \cite[sign. b-IIII-4]{RN2670}. Augsburg, Staats- und Stadtbibliothek. urn:nbn:de:bvb:12-bsb11218245-6. (\textbf{e}) Examples for two types of Sun-Zodiac tables: the ancient (\textit{veterum}) and the 16th-century variation (\textit{nostro}). The prediction of the similarity model is made explainable by highlighting the most relevant feature interactions, using here one bigram as an example. It is clearly visible that the position columns are shifted by a fixed number of days.
}
\label{fig:case_study_figure}
\end{figure*}

\section{Discussion}\label{FinalDiscussion}
The present study has shown both qualitatively and quantitatively how mathematical knowledge as taught in the frame of the early modern universities in Europe has evolved in a context of institutional competition in Europe. This competition seems to have fostered a sharing process of scientific knowledge in Europe while, as it well known, the latter was being fragmented by religious and political currents.\par

The pattern along which the conception of historical climate zones changed (from 7 to 7+2 towards 24 climate zones) allows to formulate the hypothesis that  the emergence of a shared science in continental Europe,  at least as the generally educated populace is concerned, was related to the development of a global perception beyond politics.

The computation of the position of the Sun with respect to the Zodiac, moreover, seems to indicate the emergence and spread of a societal desire to establish its own intellectual roots, namely a shared chronicle with the past. Consequently, there was a concerted effort to accurately reconstruct the chronology of the events beginning in classical antiquity. 

The development of a global cultural perspective in Europe together with the emerging need to establish the own historical roots might have contributed to the creation of the very intellectual background against which the European scientific and cultural identity was later realized (Supplementary Note \ref{supp:Identity} in Section Supplementary Text).\par

The current investigation could be extended by including, in addition to textbooks, works that were associated with the research frontiers of the time \cite{RN2956}. In this manner the relation between the diffusion of a broad  mathematical culture and those disruptive works usually associated with the idea of a scientific revolution could be studied in more detail.\par

By extending the time interval moreover, for instance by including more recent sources, the evolution of mathematical knowledge could be investigated as it transits from the early modern tabular expression of mathematical functional relations  to the more modern formulaic one. 
By broadening the geographic scope, the same phenomenon could be investigated within a global perspective, potentially allowing for the quantification of the process of European intellectual colonization. Thus spatial and temporal extensions of the source base would first require well-curated dataset of the relevant sources.\par

In the future, our ML-based atomization-recomposition framework holds the potential to unlock intricate historical analyses, such as understanding the complex interplay between various data including visual, textual and numerical elements, information related to the materiality of the sources, and social and institutional embeddings of the historical actors themselves. This approach could lead to the the possibility of generating genealogies between historical sources even before engaging in a close reading analysis (Supplementary Note \ref{supplement:sec:ML_limits} in Section Materials and Method).\par

In our new approach, the historian is assisted by our AI methodology, allowing the examination of large corpora, potentially giving rise to previously unexplored hypotheses in a {\em data-driven} manner. As evidenced by our study, new perspectives can  particularly emerge from the results of unsupervised ML analysis. These results subsequently need to be studied and validated by historians. Importantly, the general limitations presented by data-driven methods, and limited data and label availability for the generation of research hypotheses need to be considered and directly addressed. We have demonstrated how these challenges can be mitigated via efficient modeling that is embedded into a process of scrutiny, independent testing and thorough model evaluation that incorporates XAI to make the underlying ML inference processes transparent and verifiable as further discussed in Supplementary Note \ref{supplement:sec:ML_limits} in Section Materials and Method. Only after these steps can the hypotheses that have emerged be further pursued on the basis of the established methods in history writing: a hypothesis-driven research. This is precisely the path that we have followed.\par

While this ambitious vision presents numerous challenges, we emphasize that computational astronomical tables from the early modern period are exceptionally intricate sources that demand profound expertise for analysis. We have demonstrated that such analysis can be substantially augmented by ML methods. Therefore, we would like to express optimism that our general approach can be adapted and applied to other historical questions and sources.\par

The results achieved in this way may pave the way towards an even more complete integration of ML and XAI into historical disciplines while at the same time enhancing the horizon of the digital humanities. 
Importantly, we believe that the integration of humanities and ML technology needs to be problem specific and highly interwoven between the disciplines. Only through close interaction can a virtuous cycle of scholarly dialogue be achieved, ultimately leading to innovation, insights, and meaningful advancements.
In our study, ML particularly benefited from addressing the challenge of sparseness in historical data, which was solved by the novel atomization-recomposition approach.\par

Ultimately, the aspiration is to establish an AI-based assistant capable of effectively enabling an accelerated science lab for insightful historical research, interpretation, and reconstruction. Such lab would serve a more comprehensive understanding of our historical roots. 

\section*{Materials and Methods}
\subsection*{Data}\label{section:corpus}
The  ``Sacrobosco Collection'' \cite{CorpusTracer18} represents the complex edition history of the astronomy textbook `De sphaera' of Johannes de Sacrobosco, and that provides a corpus of 359 early modern printed editions, roughly 76,000 pages of material \cite{mva20}. These books were used at the European universities for the mandatory introduction to the study of astronomy and geocentric cosmology during the first curricular year. The dates of the editions of the corpus range from 1472 to 1650. This corpus enables the study of important historical questions, such as the evolution and the process of homogenization of knowledge on cosmology.

\subsubsection*{Table pages} 
From all pages of the Sacrobosco Collection, we select 9793 pages bearing one or more numerical tables, which we submit to the table similarity workflow as the Sacrobosco Tables dataset. By numerical table we refer to any tabular arrangement of data in our corpus which has at least one column with (predominantly) numerical content. We specifically exclude tables of content and book indices. The pre-selection was supported by an off-the-shelf CNN (VGG-16 \cite{Simonyan15}) trained to classify pages as bearing such numerical tables or not. The output of this CNN was checked down to a low probability of the assignment of a page as bearing a numerical table. Due to the human post processing the list of of pages with numerical tables should have  close to perfect precision and a very high recall. A list of all pages with numerical tables is provided as  \texttt{spharea\_tables\_meta.csv}, the trained model instrumental in establishing this list is provided as \texttt{sphaera\_tables\_classifier.h5}. The digital images of the pages, that we refer to as the Sphaera Tables dataset can be obtained at \texttt{sphaera\_tables\_images.zip}. 

\subsubsection*{Preparation and acquisition of ground truth } We have prepared four different ground truth datasets to train and test our model at different processing stages, \emph{single digits} and \emph{non digit content} to train the recognition model, \emph{fully annotated numbers} to test the digit recognition and the bigram expansion and \emph{sun zodiac pages} to evaluate the table similarity model. These sets are provided as \texttt{numerical\_patches.csv}, \texttt{contrast\_patches.csv}, \texttt{digit\_page\_annotations.csv} and \texttt{sun\_zodiac.csv} in the code and data  repository.

\smallskip
\paragraph*{\em Single-digits}
To capture the non-standardized print types that occur in historical corpora, we have selected a subset of important printers and have for each of them annotated five individual number patches from five different pages that contain numerical content. A dataset containing a diverse set of single digits was then created. We further have added contrastive non-digit patches that contain text, illustrations, or geometry from non-table pages.

\smallskip
\paragraph*{\em Fully annotated numbers.} We have selected 11 pages and annotated each single digit contained on the pages by a bounding box. In addition we have marked if the individual digit is the first and/or the last digit of a number. With this information, all numbers and thus also all bigrams contained on these pages can straightforwardly be reconstructed. The annotated pages have been selected to cover a wide spectrum of different manifestation of numerical content in terms of writing direction, fonts, fonts' sizes, density of digit placement on the page, etc.

\smallskip
\paragraph*{\em Sun zodiac pages.} To evaluate to what extent our approach can reproduce the salient relations between the tables in our corpus, we have chosen the sun-zodiac tables, which give the positions of the sun into the signs of the zodiac in degrees for each day of the year. This table is well-suited for evaluating our approach as it occurs in varying layouts in our corpus, where the different layouts partition the full table differently. In some cases the entire table is comprised on one page, in other books it is distributed over as many as nine pages. Due to its content, the table only comprises numbers from 1 to 31 (maximum number of 31 days per month, 30 degrees per sign of the zodiac). The table thus populates only a subspace of the feature space that we exploit for our similarity assessments. Since this subspace is more densely populated than would be expected with a uniform distribution of the data over the entire feature space, this table is particularly difficult to discriminate under our approach which makes it a  good test case.    

In our corpus, we find two variants of the sun-zodiac table in this respect: tables for the times of the
`ancient' poets (`veterum poetarum temporibus accommodata') where the sun is 16 degrees into Capricorn on the first of January, and tables for  'contemporary' times ('nostro tempori') where the sun on the first day of the year has advanced 3 degrees and is located 21 degrees into Capricorn. Essentially, this difference amounts to a shift of the columns listing the days of the year with respect to columns giving the angular locations and thus, from the perspective of our similarity model, these two variations represent the same (more abstract) table. 

We have identified 68 instances of the sun-zodiac table,  which cover a total of 250 pages in the corpus. A list of the pages containing the different versions of the sun zodiac tables is provided as   \texttt{sun\_zodiac\_pages.csv}. A ground truth histogram for the digit-features distribution of a prototypical, i.e.  noise-free and complete, sun-zodiac table is provided as \texttt{sun\_zodiac\_hist.csv}.

\smallskip
\paragraph*{\em Clime table pages}
We further collect a subset of material that is concerned with climate zone tables, which divide the surface of the ``inhabited'' world and that can be defined by the length of the solar day. This served as an indication of the overall meteorological conditions, which was in turn a determinant information in the framework of Medieval and early modern medicine. We find three different principle variants of climate zone tables that either use 7, 9 or 24 clime zones. The 225 pages containing these tables are provided as \texttt{clime\_tables.csv}. In each row, the csv file lists the occurrence of an individual clime table, specifying the type and providing metadata for the edition containing this table.\par

\subsection*{Details on the atomization-recomposition model}
\subsubsection*{Digit recognition model} As a first step, our goal is to train a single digit recognition model for which provide optimization and architecture details in the following. 
We built a 7-layer convolutional neural network using the Equivariant Steerable Pyramids framework \cite{e2cnn}, starting with an initial 4-layer equivariant convolutional block with filter sizes  \{3$\times$3, 3$\times$3, 5$\times$5, 5$\times$5\} and 8-rotational groups invariant to translations and rotations  on the $\mathbb{R}^2$-plane. Low-level features required to detect digits (lines, arches, circles) thus generalize over spatial input transformations resulting in increased  data efficiency.  A subsequent pooling layer selects the maximally activating map from the equivariant group. We use a stack of three standard convolution layers of kernel sizes \{5$\times$5, 1$\times$1, 1$\times$1\} which output $10$ activation maps $\{\ba_j(\x)\}_{j=0}^{9}$ for the digits $0$--$9$. Finally, we model variations in scan orientation and size on the page level by identifying the page scaling factor and rotation for which single digit activation maps are maximally activated.

We optimized the model using equal amounts of single-digit and non-digit patches, which resulted in around 8,000 datapoints for training. This data was further augmented using small rotations  ($\pm 10^{\circ}$), translations (0.025$\times\text{img\_width/img\_height}$ in x- and y-direction), scaling ($0.8-1.2\times$) and shearing ($\pm 5^{\circ}$) transformations.

Since numbers can occur in various contexts, e.g. as part of a table but also as a page number, we model local page context and consider a border of 10 pixels around the digit bounding box. We use the Adam optimizer to minimize the mean squared error between true activation maps and model outputs using the loss term $\ell= \ell_{bbox} + 0.3 \cdot \ell_{context}$, and select the model of best performance on the test set.

\subsubsection*{Bigram expansion} In the subsequent recomposition step, we combine these single-digit activation maps to detect digit task-relevant bigram features using a hard-coded sequence of processing layers. We compute the composed feature representations by applying an element-wise `min' operation 
\begin{align*}
\ba_{jk}^{(\tau)}(\x; s, \theta) &= \min\big\{\ba_j(\bx; s, \theta) ,\tau(\ba_k(\bx; s, \theta))\big\},
\end{align*}
which signals the presence of bigrams $jk \in \{00,\dots,99\}$ at image scale $s$ and rotation $\theta$, and can be seen as a continuous `\textsc{and}' \cite{Kauffmann20} operation. In addition, we include additional feature maps that detect isolated single digits $j \in \{\square0\square,\dots,\square9\square\}$ with ``$\square$'' indicating that no digit is detected at the given location. The function $\tau$ represents a translation operation shifting activation maps horizontally by a specified number of pixels $\delta$. To account for variations in spacing between characters, we generate bigram maps with multiple shifts $\delta$ and select at each spatial location the best shift via the max-pooling operation:

\begin{align*}
\ba_{jk}(\x; s, \theta) &= \max_\tau \big\{ \ba_{jk}^{(\tau)}(\x;  s, \theta) \big\}.
\end{align*}
The `max' operation can be interpreted as a continuous `\textsc{or}', and determines at each location whether a bigram has been found for at least one candidate alignment. Further, isolated single digits can be detected by computing neighborhood maps using shifts $\pm \delta$. These neighborhood maps are computed from the single digit maps shifted in left and right horizontal direction and further computing a binary map that signals the absence of digits. Now, a `$\min$' operation over digit map $\ba_{j}$ and both neighborhood maps indicates the presence of isolated single digits. This results in a total of 110 feature maps.

In our experiments, we use a reference page height/width of 1200 pixels, $s \in \{0.5, 0.65, 0.8, 0.95,  1.0\}$, $\theta \in \{-90,0,90\}^{\circ} $ and $\delta \in \{8,10\}$ pixels.
We finally select bigram maps from the sets of scalings, rotations and shifts for which the feature map activity is maximized.

\subsubsection*{Pooling} As a final step, we apply a spatial pooling to implement invariance with respect to the table layout and reduce dimensionality, which gives us a `bag-of-bigrams' representation for each page. We experimented with different pooling strategies and found that a standard peak-detection algorithm resulted in the best task performance, while  allowing for a directly interpretable decoding of numerical features.

For the activity peak-detection of bigrams, we start from a set of 100 bigram maps $\ba_{jk}$ with $jk=\{00,...,99\}$ which are added to 10 maps for isolated digits $\hat{\ba}_{i}$ with $i=\{\_0\_,\ldots,\_9\_\}$ resulting in $\bar{\mathbf{a}} = (\ba_{i}, \ba_{jk} )$. Since, the max-pooling used for the bigrams reduces the overall activity levels in comparison to the isolated digit maps, we introduce a scaling parameter $\alpha$ to the latter $\ba_{i} = \hat{\ba}_{i}/\alpha$. Next, we subtract a bias term $\beta \cdot \max_{(x,y)} \bar{\mathbf{a}}_{(x,y)}$ computed as the product of relative scaling parameter $\beta$ and the maximum pixel value in all maps. Resulting maps are rectified to filter weak background activity. For each of the 110 feature maps, we compute occurring peaks using the center of activity mass and further determine the linkage matrix using the distances between centers to perform a hierarchical clustering grouping close-by activated pixels into groups of pixels that belong to one bigram. To limit the size of clustered regions, we define a maximum distance parameter $d$ and select parameters using histogram Pearson correlation scores on the training patches and set $\alpha=3$, $\beta=0.12$ and $d=15$. The resulting center of mass coordinates finally give the digit location together with the digit label. \par

\subsubsection*{Explaining similarity models} To get insight into similarity predictions, we apply the purposely designed BiLRP method \cite{eberle2020}. The method assumes a similarity model of the type $y = \langle \phi(\bx),\phi(\bx')\rangle$ where $\phi$ is a neural network based feature extractor, and $y$ measures the similarity between $\bx$ and $\bx'$. The method explains the produced similarity score $y$ in terms of contributions of feature pairs $(x_i,x'_{i'})$. Conceptually, the method computes these contributions by performing a backpropagation pass from the top layer to the input layer. Each step of the backpropagation redistributes contribution scores from a given layer to the layer below. The method stops once the input features are reached. In practice, the explanation is computed more efficiently by computing multiple standard LRP explanations \cite{bach2015pixel} (one for each element of the dot-product), and recombining them at the input via a matrix product. To compute each LRP pass, we apply the LRP-0 rule \cite{lrpoverview} and pool resulting explanations over pixel regions of 15$\times$15. 

\subsection*{Evaluation}
The evaluation of the different representations used in our approach using ground truth data annotations is described in the following.
\subsubsection*{Single digit accuracy}
The trained digit encoder is used to predict digit maps on the held-out test set. For each patch the resulting activation map is computed, multiplied with a bounding box region mask and finally sum-pooled which results in a vector of size 1$\times$10. The maximally activating vector index gives the predicted digit used to compute the single digit accuracy.

\subsubsection*{Full-page bigram histograms}\label{methods:fullpage_hists}
We use the digit model to compute  110  single-digit and bigram activation maps from which we extract histogram summaries by applying peak-detection or spatial sum-pooling. Ground truth histograms are computed by identifying and counting all bigram and isolated single-digit occurrences. Each bigram count $h_{jk}$ is optionally mapped to its square root to better handle the difference of scales between frequently occurring and rare digits and bigrams respectively, and finally, Pearson correlation between ground truth and computed histograms is computed for each page. 

\subsubsection*{Cluster classification}
To validate the resulting clusters, we use a subset of the full corpus that contains one and two-page instances of the sun-zodiac tables. The corresponding 71 table pages containing more than 45,000 single digits are split into train-test (50/50) sets and a nearest-neighbor distance model is fitted on the training set. For all remaining data points, we assign the class label according to different distance models and compute the cluster purity of the test split over ten random seeds. We have compared different ways of extracting page representations: (i) Bigram: Bigram histogram counts were obtained using the bigram model with peak detection and square root mapping.  (ii) Pooled: Activity maps were obtained as in (i), but instead of peak detection, we directly applied spatial sum-pooling to the bigram maps.  (iii) Unigram: Instead of computing bigram maps, we built a ten-dimensional unigram count histogram using peak detection. (iv) VGG-16: We used the pretrained encoder of the deep image classification network VGG-16 \cite{Simonyan15} and extracted spatially-pooled output feature maps after the last of five convolutional blocks.

\subsection*{Historical corpus-level analyses}
\subsubsection*{Temporal analysis} \label{methods:temporal}
The editions of the Sacrobosco collection that contain at least one page of tables were printed during a time span of 153 years (1494-1647) over which publication rates changed considerably. Thus, we apply a sampling based temporal analysis. For each time step $t_i$, we assign a sampling probability to each book page containing a table from a truncated normal distribution $\mathcal{N}(t_i, \sigma^2 )$, which sets probabilities for data points outside the interval $(t_i - \sigma, t_i + \sigma)$ to zero. At every step, we sample $N=80$ data points, determine their cluster membership label, construct the cluster count histogram of size $1\times k$ with $k$ the number of clusters, and compute the entropy for each histogram vector. Clusters are computed using using $k$-means clustering \cite{macqueen1967} with  $k=1500$ clusters. We have further studied the robustness of our results to the choice of hyperparameter in Supplementary Material \ref{supplement:text:history_temporal}. The temporal evolution of entropy scores is computed for digit density thresholds of $\{0, 100, 200, 250, 300 \}$, which refer to the maximum number of digits detected on a page, and average entropy curves over 20 runs for each threshold.
	
\subsubsection*{Geographical analysis}
To study the varying knowledge production expressed by the tables printed across 32 different printing centers, we compute the difference in entropy between the $k$-means cluster distributions and an uninformed uniformly distributed production process \mbox{$H(p)-H(p_{\max})$}, where $p_k$ represents the probability of assigning a table to cluster $k$ with $k=1500$. The term $H(p_{\max}) = log(N_c)$ with $N_c$ the number of tables printed in city $c$ captures the maximum entropy that a cluster distribution for each print location can achieve.  Consequently, the difference in entropy is minimized for cities that  output low entropy distributions, i.e. by repeatedly printing the same material.

\bibliographystyle{unsrt-abbrv}
\footnotesize
\bibliography{bibliography}

\begin{thebibliography}{100}

\bibitem{Kyore1939}
A.~Koyré.
\newblock {\em Études galiléennes}.
\newblock Hermann, Paris, 1939.

\bibitem{Kyore1973}
A.~Koyré.
\newblock {\em The astronomical revolution. Copernicus, Kepler, Borelli}.
\newblock Cornell University Press, Ithaca, 1973.

\bibitem{Kyore1957}
A.~Koyré.
\newblock {\em From the Closed World to the Infinite Universe}.
\newblock Johns Hopkins Press, Baltimore, 1957.

\bibitem{Pedersen1993}
O.~Perdersen.
\newblock {\em Early Physics and Astronomy: A Historical Introduction}.
\newblock Cambridge University Press, Cambridge, MA, 1993.

\bibitem{Westfall1971}
R.~Westfall.
\newblock {\em The Construction of Modern Science}.
\newblock John Wiley and Sons, New York, 1971.

\bibitem{Cohen2015}
F.~H. Cohen.
\newblock {\em The Rise of Modern Science Explained: A Comparative History}.
\newblock Cambridge University Press, Cambridge, MA, 2015.

\bibitem{Kuhn1957}
T.~S. Kuhn.
\newblock {\em The Copernican Revolution}.
\newblock Harvard University Press, Cambridge, MA, 1957.

\bibitem{Kuhn1962}
T.~S. Kuhn.
\newblock {\em The Structure of Scientific Revolutions}.
\newblock University of Chicago Press, Chicago, 1962.

\bibitem{Burke2015}
P.~Burke.
\newblock {\em What Is the History of Knowledge?}
\newblock Polity Press, Cambridge, UK, 2015.

\bibitem{Daston2017}
L.~Daston.
\newblock The history of science and the history of knowledge.
\newblock {\em KNOW: A Journal on the Formation of Knowledge}, 1, 2017.

\bibitem{OestlingLarsson2020}
J.~Östling and D.~L. Heidenblad.
\newblock Fulfilling the promise of the history of knowledge: Key approaches for the 2020s.
\newblock {\em Journal for the History of Knowledge}, 1, 2020.

\bibitem{RN2956}
S.~Cole.
\newblock The hierarchy of the sciences?
\newblock {\em American Journal of Sociology}, 89(1):111–139, 1983.
\newblock Model Review 3.

\bibitem{LundBensVin2000}
A.~Lundgren and B.~Bensaude-Vincent, editors.
\newblock {\em Communicating Chem-istry. Textbooks and Their Audiences, 1789-1939}.
\newblock Science History Publications, Canton, 2000.

\bibitem{Vicedo2012}
M.~Vicedo.
\newblock Introduction: The secret lives of textbooks.
\newblock {\em Isis}, 103(1), 2012.

\bibitem{SphaeraAuthors}
M.~Valleriani, editor.
\newblock {\em De sphaera of Johannes de Sacrobosco in the Early Modern Period: The Authors of the Commentaries}.
\newblock Springer, 2020.

\bibitem{VallerianiOttone2022}
M.~Valleriani and A.~Ottone, editors.
\newblock {\em Publishing Sacrobosco's \guillemotleft De sphaera\guillemotright, in Early Modern Europe. Modes of Material and Scientific Exchange}.
\newblock Springer Nature, Cham, 2022.

\bibitem{mva19}
M.~Valleriani, F.~Kr{\"a}utli, M.~Zamani, A.~Tejedor, C.~Sander, M.~Vogl, S.~Bertram, G.~Funke, and H.~Kantz.
\newblock The emergence of epistemic communities in the sphaera corpus: Mechanisms of knowledge evolution.
\newblock {\em Journal of Historical Network Research}, 3:50--91, 2019.

\bibitem{SRN2020}
M.~Zamani, A.~Tejedor, M.~Vogl, F.~Kräutli, M.~Valleriani, and H.~Kantz.
\newblock Evolution and transformation of early modern cosmological knowledge: A network study.
\newblock {\em Scientific Reports}, 10:19822, 2020.

\bibitem{Zamani2023}
M.~Zamani, H.~El-Hajj, M.~Vogl, H.~Kantz, and M.~Valleriani.
\newblock A mathematical model for the process of accumulation of scientific knowledge in the early modern period.
\newblock {\em Humanities and Social Sciences Communications}, 10(533), 2023.

\bibitem{Gingerich1988}
O.~Gingerich.
\newblock Sacrobosco as a textbook.
\newblock {\em Journal for the History of Astronomy}, 19(4):269--273, 1988.

\bibitem{Gingerich1990}
O.~Gingerich.
\newblock Five centuries of astronomical textbooks and their role in teaching.
\newblock In J.~M. Pasachoff and J.~R. Percy, editors, {\em The Teaching of Astronomy}, pages 189--211. Cambridge University Press, Cambridge, 1990.

\bibitem{Gingerich2004}
O.~Gingerich.
\newblock {\em The Book Nobody Read: Chasing the Revolutions of Nicolaus Copernicus}.
\newblock Walker Books, London, 2004.

\bibitem{ChabasGoldstein2003}
J.~Chábas and B.~R. Goldstein.
\newblock {\em The Alfonsine Tables of Toledo}.
\newblock Springer, Dordrecht, 2003.

\bibitem{ChabasGoldstein2012}
J.~Chabás and B.~R. Goldstein.
\newblock {\em A Survey of European Astronomical Tables in the Late Middle Ages}.
\newblock Brill, Leiden, 2012.

\bibitem{lecun2015deep}
Y.~LeCun, Y.~Bengio, and G.~Hinton.
\newblock Deep learning.
\newblock {\em Nature}, 521(7553):436--444, 2015.

\bibitem{hochreiter1997long}
S.~Hochreiter and J.~Schmidhuber.
\newblock Long short-term memory.
\newblock {\em Neural computation}, 9(8):1735--1780, 1997.

\bibitem{schmidhuber2015deep}
J.~Schmidhuber.
\newblock Deep learning in neural networks: An overview.
\newblock {\em Neural networks}, 61:85--117, 2015.

\bibitem{radford2019language}
A.~Radford, J.~Wu, R.~Child, D.~Luan, D.~Amodei, and I.~Sutskever.
\newblock Language models are unsupervised multitask learners.
\newblock Technical report, OpenAI, 2019.

\bibitem{hinton2012deep}
G.~Hinton, L.~Deng, D.~Yu, G.~E. Dahl, A.-r. Mohamed, N.~Jaitly, A.~Senior, V.~Vanhoucke, P.~Nguyen, T.~N. Sainath, et~al.
\newblock Deep neural networks for acoustic modeling in speech recognition: The shared views of four research groups.
\newblock {\em IEEE Signal Processing Magazine}, 29(6):82--97, 2012.

\bibitem{Graves2013SpeechRW}
A.~Graves, A.~rahman Mohamed, and G.~E. Hinton.
\newblock Speech recognition with deep recurrent neural networks.
\newblock {\em 2013 IEEE International Conference on Acoustics, Speech and Signal Processing}, pages 6645--6649, 2013.

\bibitem{speechrecog_chiu_2018}
C.-C. Chiu, T.~N. Sainath, Y.~Wu, R.~Prabhavalkar, P.~Nguyen, Z.~Chen, A.~Kannan, R.~J. Weiss, K.~Rao, E.~Gonina, N.~Jaitly, B.~Li, J.~Chorowski, and M.~Bacchiani.
\newblock State-of-the-art speech recognition with sequence-to-sequence models.
\newblock In {\em 2018 IEEE International Conference on Acoustics, Speech and Signal Processing (ICASSP)}, pages 4774--4778, 2018.

\bibitem{ott2019fairseq}
M.~Ott, S.~Edunov, A.~Baevski, A.~Fan, S.~Gross, N.~Ng, D.~Grangier, and M.~Auli.
\newblock fairseq: A fast, extensible toolkit for sequence modeling.
\newblock In {\em Proceedings of NAACL-HLT 2019: Demonstrations}, 2019.

\bibitem{vaswani2017attention}
A.~Vaswani, N.~Shazeer, N.~Parmar, J.~Uszkoreit, L.~Jones, A.~N. Gomez, {\L}.~Kaiser, and I.~Polosukhin.
\newblock Attention is all you need.
\newblock In {\em Advances in Neural Information Processing Systems}, pages 5998--6008, 2017.

\bibitem{devlin-etal-2019-bert}
J.~Devlin, M.-W. Chang, K.~Lee, and K.~Toutanova.
\newblock {BERT}: Pre-training of deep bidirectional transformers for language understanding.
\newblock In {\em Proceedings of the 2019 Conference of the North {A}merican Chapter of the Association for Computational Linguistics: Human Language Technologies, Volume 1 (Long and Short Papers)}, pages 4171--4186, Minneapolis, Minnesota, June 2019. Association for Computational Linguistics.

\bibitem{brown2020language}
T.~B. Brown, B.~Mann, N.~Ryder, M.~Subbiah, J.~Kaplan, P.~Dhariwal, A.~Neelakantan, P.~Shyam, G.~Sastry, A.~Askell, et~al.
\newblock Language models are few-shot learners.
\newblock {\em arXiv preprint arXiv:2005.14165}, 2020.

\bibitem{lambda2022}
A.~D. Cohen, A.~Roberts, A.~Molina, A.~Butryna, A.~Jin, A.~Kulshreshtha, B.~Hutchinson, B.~Zevenbergen, B.~H. Aguera-Arcas, C.~ching Chang, C.~Cui, C.~Du, D.~D.~F. Adiwardana, D.~Chen, D.~D. Lepikhin, E.~H. Chi, E.~Hoffman-John, H.-T. Cheng, H.~Lee, I.~Krivokon, J.~Qin, J.~Hall, J.~Fenton, J.~Soraker, K.~Meier-Hellstern, K.~Olson, L.~M. Aroyo, M.~P. Bosma, M.~J. Pickett, M.~A. Menegali, M.~Croak, M.~Díaz, M.~Lamm, M.~Krikun, M.~R. Morris, N.~Shazeer, Q.~V. Le, R.~Bernstein, R.~Rajakumar, R.~Kurzweil, R.~Thoppilan, S.~Zheng, T.~Bos, T.~Duke, T.~Doshi, V.~Y. Zhao, V.~Prabhakaran, W.~Rusch, Y.~Li, Y.~Huang, Y.~Zhou, Y.~Xu, and Z.~Chen.
\newblock Lamda: Language models for dialog applications.
\newblock In {\em arXiv}. 2022.

\bibitem{mnih-atari-2013}
V.~Mnih, K.~Kavukcuoglu, D.~Silver, A.~Graves, I.~Antonoglou, D.~Wierstra, and M.~Riedmiller.
\newblock Playing atari with deep reinforcement learning.
\newblock In {\em NIPS Deep Learning Workshop}. 2013.

\bibitem{mnih2015humanlevel}
V.~Mnih, K.~Kavukcuoglu, D.~Silver, A.~A. Rusu, J.~Veness, M.~G. Bellemare, A.~Graves, M.~Riedmiller, A.~K. Fidjeland, G.~Ostrovski, S.~Petersen, C.~Beattie, A.~Sadik, I.~Antonoglou, H.~King, D.~Kumaran, D.~Wierstra, S.~Legg, and D.~Hassabis.
\newblock Human-level control through deep reinforcement learning.
\newblock {\em Nature}, 518(7540):529--533, February 2015.

\bibitem{alphago_2016}
D.~Silver, A.~Huang, C.~J. Maddison, A.~Guez, L.~Sifre, G.~van~den Driessche, J.~Schrittwieser, I.~Antonoglou, V.~Panneershelvam, M.~Lanctot, S.~Dieleman, D.~Grewe, J.~Nham, N.~Kalchbrenner, I.~Sutskever, T.~Lillicrap, M.~Leach, K.~Kavukcuoglu, T.~Graepel, and D.~Hassabis.
\newblock Mastering the game of {Go} with deep neural networks and tree search.
\newblock {\em Nature}, 529(7587):484--489, 01 2016.

\bibitem{DBLP:journals/corr/LillicrapHPHETS15}
T.~P. Lillicrap, J.~J. Hunt, A.~Pritzel, N.~Heess, T.~Erez, Y.~Tassa, D.~Silver, and D.~Wierstra.
\newblock Continuous control with deep reinforcement learning.
\newblock In Y.~Bengio and Y.~LeCun, editors, {\em 4th International Conference on Learning Representations, {ICLR} 2016, San Juan, Puerto Rico, May 2-4, 2016, Conference Track Proceedings}, 2016.

\bibitem{won2020adaptive}
D.-O. Won, K.-R. M{\"u}ller, and S.-W. Lee.
\newblock An adaptive deep reinforcement learning framework enables curling robots with human-like performance in real-world conditions.
\newblock {\em Science Robotics}, 5(46), 2020.

\bibitem{lecun98}
Y.~Lecun, L.~Bottou, Y.~Bengio, and P.~Haffner.
\newblock Gradient-based learning applied to document recognition.
\newblock {\em Proceedings of the IEEE}, 86(11):2278--2324, 1998.

\bibitem{yolo2016}
J.~Redmon, S.~Divvala, R.~Girshick, and A.~Farhadi.
\newblock You only look once: Unified, real-time object detection.
\newblock In {\em 2016 IEEE Conference on Computer Vision and Pattern Recognition (CVPR)}, pages 779--788, 2016.

\bibitem{resnet2016}
K.~He, X.~Zhang, S.~Ren, and J.~Sun.
\newblock Deep residual learning for image recognition.
\newblock In {\em Proceedings of the IEEE Conference on Computer Vision and Pattern Recognition (CVPR)}, June 2016.

\bibitem{Baldi2014}
P.~Baldi, P.~Sadowski, and D.~Whiteson.
\newblock Searching for exotic particles in high-energy physics with deep learning.
\newblock {\em Nature Communications}, 5(4308), 2014.

\bibitem{Schuett2017}
K.~T. Sch\"{u}tt, F.~Arbabzadah, S.~Chmiela, K.~R. M\"{u}ller, and A.~Tkatchenko.
\newblock Quantum-chemical insights from deep tensor neural networks.
\newblock {\em Nature Communications}, 8:13890, 2017.

\bibitem{doi:10.1021/acs.chemrev.1c00107}
J.~A. Keith, V.~Vassilev-Galindo, B.~Cheng, S.~Chmiela, M.~Gastegger, K.-R. Müller, and A.~Tkatchenko.
\newblock Combining machine learning and computational chemistry for predictive insights into chemical systems.
\newblock {\em Chemical Reviews}, 121(16):9816--9872, 2021.
\newblock PMID: 34232033.

\bibitem{Reichstein2019DeepLA}
M.~Reichstein, G.~Camps-Valls, B.~Stevens, M.~Jung, J.~Denzler, N.~Carvalhais, and Prabhat.
\newblock Deep learning and process understanding for data-driven earth system science.
\newblock {\em Nature}, 566:195 -- 204, 2019.

\bibitem{samek2021explaining}
W.~Samek, G.~Montavon, S.~Lapuschkin, C.~J. Anders, and K.-R. M{\"u}ller.
\newblock Explaining deep neural networks and beyond: A review of methods and applications.
\newblock {\em Proceedings of the IEEE}, 109(3):247--278, 2021.

\bibitem{binder2021morphological}
A.~Binder, M.~Bockmayr, M.~H{\"a}gele, S.~Wienert, D.~Heim, K.~Hellweg, M.~Ishii, A.~Stenzinger, A.~Hocke, C.~Denkert, et~al.
\newblock Morphological and molecular breast cancer profiling through explainable machine learning.
\newblock {\em Nature Machine Intelligence}, 3:355–366, 2021.

\bibitem{Jumper2021HighlyAP}
J.~M. Jumper, R.~Evans, A.~Pritzel, T.~Green, M.~Figurnov, O.~Ronneberger, K.~Tunyasuvunakool, R.~Bates, A.~Z{\'i}dek, A.~Potapenko, A.~Bridgland, C.~Meyer, S.~A.~A. Kohl, A.~Ballard, A.~Cowie, B.~Romera-Paredes, S.~Nikolov, R.~Jain, J.~Adler, T.~Back, S.~Petersen, D.~A. Reiman, E.~Clancy, M.~Zielinski, M.~Steinegger, M.~Pacholska, T.~Berghammer, S.~Bodenstein, D.~Silver, O.~Vinyals, A.~W. Senior, K.~Kavukcuoglu, P.~Kohli, and D.~Hassabis.
\newblock Highly accurate protein structure prediction with alphafold.
\newblock {\em Nature}, 596:583 -- 589, 2021.

\bibitem{Papadopoulos2013}
C.~Papadopoulos, S.~Pletschacher, C.~Clausner, and A.~Antonacopoulos.
\newblock The impact dataset of historical document images.
\newblock In {\em Proceedings of the 2nd International Workshop on Historical Document Imaging and Processing}, HIP '13, page 123–130, New York, NY, USA, 2013. Association for Computing Machinery.

\bibitem{Fischer2020}
A.~Fischer.
\newblock {\em IAM-HistDB: A Dataset of Handwritten Historical Documents}, pages 11 -- 23.
\newblock World Scientific, 2020.

\bibitem{Nikolaidou2022}
K.~Nikolaidou, M.~Sueret, H.~Mojayet, and M.~Liwicki.
\newblock A survey of historical document image datasets.
\newblock {\em International Journal on Document Analysis and Recognition (IJDAR)}, 25:305--338, 2022.

\bibitem{Buttner2022}
J.~Büttner, J.~Martinetz, H.~El-Hajj, and M.~Valleriani.
\newblock Cordeep and the sacrobosco dataset: Detection of visual elements in historical documents.
\newblock {\em Journal of Imaging}, 8(285), 2022.

\bibitem{grasshoff2021kepler}
G.~Gra{\ss}hoff and M.~Y. Abkenar.
\newblock Kepler’s astronomia nova--a challenge for computational history and the philosophy of science.
\newblock In {\em Applied and Computational Historical Astronomy. Angewandte und computergest{\"u}tzte historische Astronomie.: Proceedings of the Splinter Meeting in the Astronomische Gesellschaft, Sept. 25, 2020. Nuncius Hamburgensis-Beitr{\"a}ge zur Geschichte der Naturwissenschaften; Vol. 55}, volume~55. tredition, 2021.

\bibitem{journals/corr/RonnebergerFB15}
O.~Ronneberger, P.~Fischer, and T.~Brox.
\newblock {\em U-Net: Convolutional Networks for Biomedical Image Segmentation}, pages 234--241.
\newblock Springer International Publishing, 2015.

\bibitem{FasterRCNN2015}
S.~Ren, K.~He, R.~Girshick, and J.~Sun.
\newblock Faster r-cnn: Towards real-time object detection with region proposal networks.
\newblock In C.~Cortes, N.~Lawrence, D.~Lee, M.~Sugiyama, and R.~Garnett, editors, {\em Advances in Neural Information Processing Systems}, volume~28. Curran Associates, Inc., 2015.

\bibitem{monnier2020docExtractor}
T.~Monnier and M.~Aubry.
\newblock {docExtractor: An off-the-shelf historical document element extraction}.
\newblock In {\em ICFHR}, 2020.

\bibitem{Abhishek2021}
A.~Dutta, G.~Bergel, and A.~Zisserman.
\newblock Visual analysis of chapbooks printed in scotland.
\newblock In {\em The 6th International Workshop on Historical Document Imaging and Processing}, HIP '21, page 67–72, New York, NY, USA, 2021. Association for Computing Machinery.

\bibitem{Tsochatzidis2021}
L.~Tsochatzidis, S.~Symeonidis, A.~Papazoglou, and I.~Pratikakis.
\newblock Htr for greek historical handwritten documents.
\newblock {\em Journal of Imaging}, 7(12), 2021.

\bibitem{Wick2021}
C.~Wick, J.~Z{\"o}llner, and T.~Gr{\"u}ning.
\newblock Transformer for handwritten text recognition using bidirectional post-decoding.
\newblock In J.~Llad{\'o}s, D.~Lopresti, and S.~Uchida, editors, {\em Document Analysis and Recognition -- ICDAR 2021}, pages 112--126, Cham, 2021. Springer International Publishing.

\bibitem{Li2021TrOCR}
M.~Li, T.~Lv, L.~Cui, Y.~Lu, D.~A.~F. Flor{\^{e}}ncio, C.~Zhang, Z.~Li, and F.~Wei.
\newblock Trocr: Transformer-based optical character recognition with pre-trained models.
\newblock {\em CoRR}, abs/2109.10282, 2021.

\bibitem{strobel2022transformerbased}
P.~Ströbel, S.~Clematide, T.~Hodel, and M.~Volk.
\newblock Transformer-based htr for historical documents.
\newblock In {\em Workshop on Computational Methods in the Humanities 2022}, 2022.

\bibitem{Smits2023}
T.~Smits and M.~Wevers.
\newblock {A multimodal turn in Digital Humanities. Using contrastive machine learning models to explore, enrich, and analyze digital visual historical collections}.
\newblock {\em Digital Scholarship in the Humanities}, 03 2023.
\newblock fqad008.

\bibitem{Assael2019}
Y.~Assael, T.~Sommerschield, and J.~Prag.
\newblock Restoring ancient text using deep learning: a case study on {G}reek epigraphy.
\newblock In {\em Proceedings of the 2019 Conference on Empirical Methods in Natural Language Processing and the 9th International Joint Conference on Natural Language Processing (EMNLP-IJCNLP)}, pages 6368--6375, Hong Kong, China, November 2019. Association for Computational Linguistics.

\bibitem{AssaelEtAlNature2022}
Y.~Assael, T.~Sommerschield, B.~Schillingford, M.~Bodbar, J.~Pavlopoulos, M.~Chatzipanangiotou, I.~Androutsopulos, J.~Prag, and N.~de~Freitas.
\newblock Restoring and attributing ancient texts using deep neural networks.
\newblock {\em Nature}, 603:280--283, 2022.

\bibitem{Bamman2020}
D.~Bamman and P.~J. Burns.
\newblock Latin {BERT:} {A} contextual language model for classical philology.
\newblock {\em CoRR}, abs/2009.10053, 2020.

\bibitem{Fetaya2020}
E.~Fetaya, Y.~Lifshitz, E.~Aaron, and S.~Gordin.
\newblock Restoration of fragmentary babylonian texts using recurrent neural networks.
\newblock {\em Proceedings of the National Academy of Sciences}, 117(37):22743--22751, sep 2020.

\bibitem{Barucci2021}
A.~Barucci, C.~Cucci, M.~Franci, M.~Loschiavo, and F.~Argenti.
\newblock A deep learning approach to ancient egyptian hieroglyphs classification.
\newblock {\em IEEE Access}, 9:123438--123447, 2021.

\bibitem{Pawlowicz2021}
L.~M. Pawlowicz and C.~E. Downum.
\newblock Applications of deep learning to decorated ceramic typology and classification: A case study using tusayan white ware from northeast arizona.
\newblock {\em Journal of Archaeological Science}, 130:105375, 2021.

\bibitem{Bell2021}
P.~Bell and F.~Offert.
\newblock Reflections on connoisseurship and computer vision.
\newblock {\em Journal of Art Historiography}, 24, 2021.

\bibitem{ElHajjEberle_xaidh_2023}
H.~El-Hajj, O.~Eberle, A.~Merklein, A.~Siebold, N.~Shlomi, J.~Büttner, J.~Martinetz, K.-R. Müller, G.~Montavon, and M.~Valleriani.
\newblock Explainability and transparency in the realm of digital humanities: Toward a historian xai.
\newblock {\em International Journal of Digital Humanities}, 2023.

\bibitem{sugiyama2012machine}
M.~Sugiyama and M.~Kawanabe.
\newblock {\em Machine learning in non-stationary environments: Introduction to covariate shift adaptation}.
\newblock MIT press, 2012.

\bibitem{Gilbert1995}
B.~Gilbert.
\newblock {\em The Art of the Woodcut in the Italian Renaissance Book}.
\newblock The Grolier Club, New York, 1995.

\bibitem{Eisenstein1996}
E.~L. Eisenstein.
\newblock {\em The Printing Revolution in Early Modern Europe}.
\newblock Cambridge University Press, Cambridge, 1996.

\bibitem{MacLean2009}
I.~Maclean.
\newblock {\em Learning and the Market Place: Essays in the History of the Early Modern Book}.
\newblock Brill, Leiden, 2009.

\bibitem{Nuovo2013}
A.~Nuovo.
\newblock {\em The Book Trade in the Italian Renaissance}.
\newblock Brill, Leiden, 2013.

\bibitem{eberle2020}
O.~Eberle, J.~Büttner, F.~Kräutli, K.-R. Müller, M.~Valleriani, and G.~Montavon.
\newblock Building and interpreting deep similarity models.
\newblock {\em IEEE Transactions on Pattern Analysis and Machine Intelligence}, 44(3):1149--1161, 2022.

\bibitem{Grendler2022}
P.~F. Grendler.
\newblock The «sphaera» in the jesuit education.
\newblock In M.~Valleriani and A.~Ottone, editors, {\em Publishing Sacrobosco’s «De sphaera» in Early Modern Europe. Modes of Material and Scientific Exchange}, pages 369--406. Springer Nature, Cham, 2022.

\bibitem{VallerianiOttoneChap12022}
M.~Valleriani and A.~Ottone.
\newblock Printers, publishers, and sellers: Actors in the process of consolidation of epistemic communities in the early modern academic world.
\newblock In M.~Valleriani and A.~Ottone, editors, {\em Publishing Sacrobosco’s «De sphaera» in Early Modern Europe. Modes of Material and Scientific Exchange}, pages 1--24. Springer Nature, Cham, 2022.

\bibitem{Maclean2022}
I.~Maclean.
\newblock Sacrobosco at the book fairs, 1576–1624: The pedagogical marketplace.
\newblock In M.~Valleriani and A.~Ottone, editors, {\em Publishing Sacrobosco’s «De sphaera» in Early Modern Europe. Modes of Material and Scientific Exchange}, pages 195--232. Springer Nature, Cham, 2022.

\bibitem{RN2923}
M.~Valleriani, M.~Vogl, H.~El-Hajj, and K.~Pham.
\newblock The network of early modern printers and its impact on the evolution of scientific knowledge: Automatic detection of awareness relations.
\newblock {\em Histories}, 2(4):466–503, 2022.

\bibitem{Axworthy2020}
A.~Axworthy.
\newblock Oronce fine and sacrobosco: From the edition of the \textit{Tractatus de sphaera} (1516) to the cosmographia (1532).
\newblock In M.~Valleriani, editor, {\em \textit{De sphaera} of Johannes de Sacrobosco in the Early Modern Period: The Authors of the Commentaries}, pages 185--264. Springer Nature, 2020.

\bibitem{LimbachForth}
S.~Limbach.
\newblock Scholars, printers, the sphere: New evidence for the challenging production of academic books in wittenberg, 1531–1550.
\newblock In M.~Valleriani and A.~Ottone, editors, {\em Publishing Sacrobosco's De sphaera in Early Modern Europe. Modes of Material and Scientific Exchange}, pages 147--185. Springer, 2022.

\bibitem{RN2847}
C.~Domtera-Schleichardt.
\newblock {\em Die Wittenberger »Scripta publice proposita« (1540–1569). Universitätsbekanntmachungen im Umfeld des späten Melanchthon}.
\newblock Evangelische Verlagsanstalt, Leipzig, 2021.

\bibitem{Jackson2013}
C.~D. Jackson.
\newblock Educational reforms of wittenberg and their faithfulness to martin luther’s thought.
\newblock {\em Christian Education Journal: Research on Educational Ministry}, 10:71--87, 2013.

\bibitem{RN1938}
R.~L. Kremer.
\newblock {\em Incunable Almanacs and Practica as Practical Knowledge Produced in Trading Zones}, page 333–369.
\newblock Springer, Dordrecht, 2017.

\bibitem{RN820}
H.~Leitão.
\newblock {\em Um Mundo Novo e uma Nova Ciência}, page 16–39.
\newblock Fundação Calouste Gulbenkian, Lisboa, 2013.

\bibitem{Burke1969}
P.~Burke.
\newblock {\em The Renaissance Sense of the Past}.
\newblock Edward Arnold, London, 1969.

\bibitem{Tanaka2019}
S.~Tanaka.
\newblock {\em History without Chronology}.
\newblock Lever Press, Ann Arbor, MI, 2019.

\bibitem{RN1480}
K.~Reich and E.~Knobloch.
\newblock Melanchthons vorreden zu sacroboscos «spahera» (1531) und zum «computus ecclesiasticus» (1538).
\newblock {\em Beiträge zur Astronomiegeschichte}, 7:13–44, 2004.

\bibitem{MVBFON2022}
M.~Valleriani, B.~Federau, and O.~Nicolaeva.
\newblock The hidden \textit{Praeceptor}: How georg rheticus taught geocentric cosmology to europe.
\newblock {\em Perspectives on Science}, 30(3):1--46, 2022.

\bibitem{Pantin2020}
I.~Pantin.
\newblock Borrowers and innovators in the printing history of sacrobosco: The case of the “in-octavo” tradition.
\newblock In M.~Valleriani, editor, {\em De sphaera of Johannes de Sacrobosco in the Early Modern Period: The Authors of the Commentaries}, pages 265--312. Springer Nature, Cham, 2020.

\bibitem{RN2670}
F.~Faleiro.
\newblock {\em Tratado del Esphera y del arte del marear: con el regimiento de las alturas: con algunas reglas nuevamente escritas muy necessarias}.
\newblock Juan Cromberger, Seville, 1535.

\bibitem{CorpusTracer18}
F.~Kr{\"{a}}utli and M.~Valleriani.
\newblock Corpus{T}racer: {A} {CIDOC} database for tracing knowledge networks.
\newblock {\em Digital Scholarship in the Humanities}, 33:336--346, 2018.

\bibitem{mva20}
M.~Valleriani.
\newblock Prolegomena to the study of early modern commentators on {J}ohannes de {S}acrobosco's tractatus de sphaera.
\newblock In M.~Valleriani, editor, {\em De sphaera of Johannes de Sacrobosco in the Early Modern Period: The Authors of the Commentaries}, pages 1--23. Springer, 2019.

\bibitem{Simonyan15}
K.~Simonyan and A.~Zisserman.
\newblock Very deep convolutional networks for large-scale image recognition.
\newblock In {\em International Conference on Learning Representations}, 2015.

\bibitem{e2cnn}
M.~Weiler and G.~Cesa.
\newblock {General E(2)-Equivariant Steerable CNNs}.
\newblock In {\em Conference on Neural Information Processing Systems (NeurIPS)}, 2019.

\bibitem{Kauffmann20}
J.~Kauffmann, K.-R. M{\"u}ller, and G.~Montavon.
\newblock Towards explaining anomalies: A deep {T}aylor decomposition of one-class models.
\newblock {\em Pattern Recognition}, 101:107198, 2020.

\bibitem{bach2015pixel}
S.~Bach, A.~Binder, G.~Montavon, F.~Klauschen, K.-R. M{\"u}ller, and W.~Samek.
\newblock On pixel-wise explanations for non-linear classifier decisions by layer-wise relevance propagation.
\newblock {\em PloS one}, 10(7):e0130140, 2015.

\bibitem{lrpoverview}
G.~Montavon, A.~Binder, S.~Lapuschkin, W.~Samek, and K.-R. M{\"{u}}ller.
\newblock Layer-wise relevance propagation: An overview.
\newblock In {\em Explainable {AI}}, volume 11700 of {\em Lecture Notes in Computer Science}, pages 193--209. Springer, 2019.

\bibitem{macqueen1967}
J.~MacQueen.
\newblock Some methods for classification and analysis of multivariate observations.
\newblock In {\em Proceedings of the Fifth Berkeley Symposium on Mathematical Statistics and Probability, Volume 1: Statistics}, pages 281--297, Berkeley, Calif., 1967. University of California Press.

\bibitem{RN2520}
M.~Valleriani and C.~Sander.
\newblock Paratexts, printers, and publishers: Book production in social context.
\newblock In M.~Valleriani and A.~Ottone, editors, {\em Publishing Sacrobosco’s «De sphaera» in Early Modern Europe. Modes of Material and Scientific Exchange}, page 337–367. Springer, Cham, 2022.

\bibitem{Kikuchi2022}
C.~Rideau-Kikuchi.
\newblock Erhard ratdolt’s edition of sacrobosco’s «tractatus de sphaera:» a new editorial model in venice?
\newblock In M.~Valleriani and A.~Ottone, editors, {\em Publishing Sacrobosco’s «De sphaera» in Early Modern Europe. Modes of Material and Scientific Exchange}, pages 61--98. Springer Nature, Cham, 2022.

\bibitem{RN2949}
M.~Valleriani, F.~Kräutli, D.~Lockhorst, and N.~Shlomi.
\newblock {\em Vision on Vision: Defining Similarities Among Early Modern Illustrations on Cosmology}, page 99–137.
\newblock Springer, Cham, 2023.

\bibitem{RN2852}
M.~Valleriani and F.~Kräutli.
\newblock {\em The Necessity of Linked Data alias Thinking Big in Computational History}, page 171–191.
\newblock vdf Hochschulverlag AG, Zürich, 2022.

\bibitem{RN2922}
J.~Büttner, J.~Martinetz, H.~El-Hajj, and M.~Valleriani.
\newblock Cordeep and the sacrobosco dataset: Detection of visual elements in historical documents.
\newblock {\em Journal of Imaging}, 8(10):285, 2022.

\bibitem{el-HajjetalRevPap2022}
H.~El-Hajj, M.~Zamani, J.~Büttner, J.~Martinetz, O.~Eberle, N.~Shlomi, A.~Siebold, G.~Montavon, K.-R. Müller, H.~Kantz, and M.~Valleriani.
\newblock An ever-expanding humanities knowledge graph: The sphaera corpus at the intersection of humanities, data management, and machine learning.
\newblock {\em Datenbank-Spektrum: Zeitschrift für Datenbanktechnologien und Information Retrieval}, 2022.

\bibitem{Kraeutli2020}
F.~Kr{\"a}utli, D.~Lockhorst, and M.~Valleriani.
\newblock Calculating sameness: Identifying early-modern image reuse outside the black box.
\newblock {\em Digital Scholarship in the Humanities}, 36(2):165--174, 12 2020.

\bibitem{Limbach2022}
S.~Limbach.
\newblock Scholars, printers, and the sphere: New evidence for the challenging production of academic books in wittenberg, 1531–1550.
\newblock In M.~Valleriani and A.~Ottone, editors, {\em Publishing Sacrobosco’s «De sphaera» in Early Modern Europe. Modes of Material and Scientific Exchange}, pages 155--194. Springer Nature, Cham, 2022.

\bibitem{Sacroboscoetal1490}
J.~de~Sacrobosco, J.~Regiomontanus, and G.~von Peuerbach.
\newblock {\em Spaerae mundi compendium foeliciter inchoat. Noviciis adolescentibus: ad astronomicam rem publicam capessendam aditum impetrantibus: pro brevi rectoque tramite a vulgari vestigio semoto: Ioannis de Sacro busto sphaericum opusculum una cum additionibus nonnullis littera A sparsim ubi intersertae sint signatis: Contraque cremonensia in planetarum theoricas delyramenta Ioannis de monte regio disputationes tam acuratiss. atque utills. Nec non Georgii purbachii in erundem motus planetarum acuratiss. theoricae: dicatum opus: utili serie contextum: fausto sidere inchoat}.
\newblock Ottaviano Scoto I, Venice, 1490.

\bibitem{Buettner2017}
J.~B{\"u}ttner.
\newblock Shooting with ink.
\newblock In M.~Valleriani, editor, {\em The Structures of Practical Knowledge}, pages 115--166. Springer Nature, Cham, 2017.

\bibitem{VallerianiQuad2022}
M.~Valleriani.
\newblock From the quadrivium to modern science.
\newblock {\em HoST - Journal of History of Science and Technology}, 16(1):34--45, 2022.

\bibitem{Cooper2004}
G.~M. Cooper.
\newblock Numbers, prognosis, and healing: Galen on medical theory.
\newblock {\em Journal of the Washington Academy of Sciences}, 98(2):45--60, 2004.

\bibitem{Clavius1585}
J.~de~Sacrobosco and C.~Clavius.
\newblock {\em Christophori Clavii Bambergensis ex Societate Iesu in Sphaeram Ioannis de Sacro Bosco commentarius Nunc tertio ab ipso Auctore recognitus, \& plerisque in locis locupletatus. Permissu superiorem}.
\newblock Domenico Basa, Rome, 1585.

\bibitem{Clavius1591}
J.~de~Sacrobosco and C.~Clavius.
\newblock {\em Christophori Clavii Bambergensis ex Societate Iesu In Sphaeram Ioannis de Sacro Bosco commentarius, Nunc tertio ab ipso Auctore recognitus, \& plerisque in locis locupletatus. Permissu Superiorum}.
\newblock Giovanni Battista Ciotti, Venice, 1591.

\bibitem{Fine1532}
O.~Finé.
\newblock {\em Orontii Finei Delphinatis, liberalium disciplinarum professoris regii, protomathesis: Opus varium, ac scitu non minus utile quàm iucundum, nunc primùm in lucem foeliciter emissum. Cuius index universalis, in versa pagina continetur}.
\newblock Jean Pierre de Tour for Gérard Morrhy, Paris, 1532.

\bibitem{Qiu2016ASO}
J.~Qiu, Q.~hui Wu, G.~Ding, Y.~Xu, and S.~Feng.
\newblock A survey of machine learning for big data processing.
\newblock {\em EURASIP Journal on Advances in Signal Processing}, 2016:1--16, 2016.

\bibitem{Adibuzzaman2017}
M.~Adibuzzaman, P.~DeLaurentis, J.~Hill, and B.~Benneyworth.
\newblock Big data in healthcare - the promises, challenges and opportunities from a research perspective: A case study with a model database.
\newblock {\em AMIA ... Annual Symposium proceedings. AMIA Symposium}, 2017:384--392, 04 2018.

\bibitem{Kelly2019}
C.~Kelly, A.~Karthikesalingam, M.~Suleyman, G.~Corrado, and D.~King.
\newblock Key challenges for delivering clinical impact with artificial intelligence.
\newblock {\em BMC Medicine}, 17, 12 2019.

\bibitem{Fine1542}
O.~Finé.
\newblock {\em De Mundi sphaera, sive Cosmographia, primáve Astronomiae parte, Lib. V}.
\newblock Simon de Colines, Paris, 1542.

\bibitem{Fine1587}
O.~Finé.
\newblock {\em Opere...Divise in cinque parti; arimetica, geometria, cosmografia, et orivoli}.
\newblock Francesco de Franceschi, Venice, 1587.

\bibitem{Giuntini1578}
J.~d. Sacrobosco and F.~Giuntini.
\newblock {\em Commentaria in Sphæram Ioannis de Sacro Bosco accuratissima}.
\newblock Philippe Tinghi, Lyon, 1578.

\bibitem{SacroWitt1550}
J.~d. Sacrobosco and P.~Melanchthon.
\newblock {\em Ioannis de Sacrobusto libellus de sphaera}.
\newblock Johann Krafft, Wittenberg, 1550.

\bibitem{matterport_maskrcnn_2017}
W.~Abdulla.
\newblock Mask r-cnn for object detection and instance segmentation on keras and tensorflow.
\newblock \url{https://github.com/matterport/Mask_RCNN}, 2017.

\bibitem{10.5555/3016100.3016186}
B.~Sun, J.~Feng, and K.~Saenko.
\newblock Return of frustratingly easy domain adaptation.
\newblock In {\em Proceedings of the Thirtieth AAAI Conference on Artificial Intelligence}, AAAI'16, page 2058–2065. AAAI Press, 2016.

\bibitem{pmlr-v97-zhao19a}
H.~Zhao, R.~T.~D. Combes, K.~Zhang, and G.~Gordon.
\newblock On learning invariant representations for domain adaptation.
\newblock In K.~Chaudhuri and R.~Salakhutdinov, editors, {\em Proceedings of the 36th International Conference on Machine Learning}, volume~97 of {\em Proceedings of Machine Learning Research}, pages 7523--7532. PMLR, 09--15 Jun 2019.

\bibitem{DBLP:journals/corr/abs-2103-03097}
J.~Wang, C.~Lan, C.~Liu, Y.~Ouyang, and T.~Qin.
\newblock Generalizing to unseen domains: {A} survey on domain generalization.
\newblock {\em CoRR}, abs/2103.03097, 2021.

\bibitem{DBLP:journals/corr/abs-2106-04923}
L.~And{\'{e}}ol, Y.~Kawakami, Y.~Wada, T.~Kanamori, K.-R. M{\"{u}}ller, and G.~Montavon.
\newblock Learning domain invariant representations by joint wasserstein distance minimization.
\newblock {\em CoRR}, abs/2106.04923, 2021.

\bibitem{10.5555/3385337}
M.~Y. Yang, B.~Rosenhahn, and V.~Murino.
\newblock {\em Multimodal Scene Understanding: Algorithms, Applications and Deep Learning}.
\newblock Academic Press, Inc., USA, 1st edition, 2019.

\bibitem{NEURIPS2019_9015}
A.~Paszke, S.~Gross, F.~Massa, A.~Lerer, J.~Bradbury, G.~Chanan, T.~Killeen, Z.~Lin, N.~Gimelshein, L.~Antiga, A.~Desmaison, A.~Kopf, E.~Yang, Z.~DeVito, M.~Raison, A.~Tejani, S.~Chilamkurthy, B.~Steiner, L.~Fang, J.~Bai, and S.~Chintala.
\newblock Pytorch: An imperative style, high-performance deep learning library.
\newblock In H.~Wallach, H.~Larochelle, A.~Beygelzimer, F.~d\textquotesingle Alch\'{e}-Buc, E.~Fox, and R.~Garnett, editors, {\em Advances in Neural Information Processing Systems 32}, pages 8024--8035. Curran Associates, Inc., 2019.

\bibitem{537667}
E.~Simoncelli and W.~Freeman.
\newblock The steerable pyramid: a flexible architecture for multi-scale derivative computation.
\newblock In {\em Proceedings., International Conference on Image Processing}, volume~3, pages 444--447 vol.3, 1995.

\bibitem{Zheng2004}
Y.~Zheng, H.~Li, and D.~Doermann.
\newblock Machine printed text and handwriting identification in noisy document images.
\newblock {\em IEEE Transactions on Pattern Analysis and Machine Intelligence}, 26(3):337--353, 2004.

\bibitem{Martinek2020}
J.~Mart\'{\i}nek, L.~Lenc, and P.~Kr\'{a}l.
\newblock Building an efficient ocr system for historical documents with little training data.
\newblock {\em Neural Comput. Appl.}, 32(23):17209–17227, dec 2020.

\bibitem{Lijun2021}
L.~Lyu, M.~Koutraki, M.~Krickl, and B.~Fetahu.
\newblock {Neural OCR Post-Hoc Correction of Historical Corpora}.
\newblock {\em Transactions of the Association for Computational Linguistics}, 9:479--493, 05 2021.

\bibitem{Diaz2021}
D.~H. Diaz, S.~Qin, R.~R. Ingle, Y.~Fujii, and A.~Bissacco.
\newblock Rethinking text line recognition models.
\newblock {\em CoRR}, abs/2104.07787, 2021.

\bibitem{Alaa2019}
A.~Sulaiman, K.~Omar, and M.~F. Nasrudin.
\newblock Degraded historical document binarization: A review on issues, challenges, techniques, and future directions.
\newblock {\em Journal of Imaging}, 5(4), 2019.

\bibitem{Bamman2012}
D.~Bamman and D.~Smith.
\newblock Extracting two thousand years of latin from a million book library.
\newblock {\em J. Comput. Cult. Herit.}, 5(1), apr 2012.

\bibitem{lapuschkin-ncomm19}
S.~Lapuschkin, S.~W{\"a}ldchen, A.~Binder, G.~Montavon, W.~Samek, and K.-R. M{\"u}ller.
\newblock Unmasking {C}lever {H}ans predictors and assessing what machines really learn.
\newblock {\em Nature Communications}, 10:1096, 2019.

\bibitem{DBLP:journals/dsp/MontavonSM18}
G.~Montavon, W.~Samek, and K.-R. M{\"{u}}ller.
\newblock Methods for interpreting and understanding deep neural networks.
\newblock {\em Digital Signal Processing}, 73:1--15, 2018.

\bibitem{DBLP:series/lncs/11700}
W.~Samek, G.~Montavon, A.~Vedaldi, L.~K. Hansen, and K.-R. M{\"{u}}ller, editors.
\newblock {\em Explainable {AI:} Interpreting, Explaining and Visualizing Deep Learning}, volume 11700 of {\em Lecture Notes in Computer Science}.
\newblock Springer, 2019.

\bibitem{PMID:33079674}
E.~Tjoa and C.~Guan.
\newblock A survey on explainable artificial intelligence (xai): Toward medical xai.
\newblock {\em IEEE transactions on neural networks and learning systems}, 32(11):4793—4813, November 2021.

\bibitem{Rupp2012}
M.~Rupp, A.~Tkatchenko, K.-R. M\"{u}ller, and O.~A. von Lilienfeld.
\newblock Fast and accurate modeling of molecular atomization energies with machine learning.
\newblock {\em Physical review letters}, 108:058301, January 2012.

\bibitem{SchNet}
K.~T. Sch\"{u}tt, H.~E. Sauceda, P.-J. Kindermans, A.~Tkatchenko, and K.-R. M\"{u}ller.
\newblock {SchNet} {\textendash} a deep learning architecture for molecules and materials.
\newblock {\em The Journal of Chemical Physics}, 148(24):241722, 2018.

\bibitem{doi:10.1021/acs.chemrev.0c01111}
O.~T. Unke, S.~Chmiela, H.~E. Sauceda, M.~Gastegger, I.~Poltavsky, K.~T. Schütt, A.~Tkatchenko, and K.-R. Müller.
\newblock Machine learning force fields.
\newblock {\em Chemical Reviews}, 121(16):10142--10186, 2021.
\newblock PMID: 33705118.

\bibitem{Sumbul2019BigearthnetAL}
G.~Sumbul, M.~Charfuelan, B.~Demir, and V.~Markl.
\newblock Bigearthnet: A large-scale benchmark archive for remote sensing image understanding.
\newblock {\em IGARSS 2019 - 2019 IEEE International Geoscience and Remote Sensing Symposium}, pages 5901--5904, 2019.

\bibitem{doi:10.1126/sciadv.aau4996}
J.~Runge, P.~Nowack, M.~Kretschmer, S.~Flaxman, and D.~Sejdinovic.
\newblock Detecting and quantifying causal associations in large nonlinear time series datasets.
\newblock {\em Science Advances}, 5(11):eaau4996, 2019.

\bibitem{https://doi.org/10.1029/2019MS002002}
B.~A. Toms, E.~A. Barnes, and I.~Ebert-Uphoff.
\newblock Physically interpretable neural networks for the geosciences: Applications to earth system variability.
\newblock {\em Journal of Advances in Modeling Earth Systems}, 12(9):e2019MS002002, 2020.
\newblock e2019MS002002 10.1029/2019MS002002.

\bibitem{Shallue2017IdentifyingEW}
C.~J. Shallue and A.~M. Vanderburg.
\newblock Identifying exoplanets with deep learning: A five planet resonant chain around kepler-80 and an eighth planet around kepler-90.
\newblock {\em arXiv: Earth and Planetary Astrophysics}, 2017.

\bibitem{Valizadegan2021Exominer}
H.~Valizadegan, M.~Martinho, L.~Wilkens, J.~Jenkins, J.~Smith, D.~Caldwell, P.~Gerum, N.~Walia, K.~Hausknecht, N.~Lubin, J.~Twicken, and N.~Oza.
\newblock Exominer: A highly accurate and explainable deep learning classifier that validates 200+ new exoplanets.
\newblock {\em Bulletin of the AAS}, 53(6), 6 2021.
\newblock https://baas.aas.org/pub/2021n6i108p06.

\bibitem{Klauschen2018}
F.~Klauschen, K.-R. Müller, A.~Binder, M.~Bockmayr, M.~Hägele, P.~Seegerer, S.~Wienert, G.~Pruneri, S.~Maria, S.~Badve, S.~Michiels, T.~Nielsen, S.~Adams, P.~Savas, F.~Symmans, S.~Willis, T.~Gruosso, M.~Park, B.~Haibe-Kains, and C.~Denkert.
\newblock Scoring of tumor-infiltrating lymphocytes: From visual estimation to machine learning.
\newblock {\em Seminars in Cancer Biology}, 52, 07 2018.

\bibitem{breastcancer2021}
A.~Binder, M.~Bockmayr, M.~Hägele, S.~Wienert, D.~Heim, K.~Hellweg, M.~Ishii, A.~Stenzinger, A.~Hocke, C.~Denkert, K.-R. Müller, and F.~Klauschen.
\newblock Morphological and molecular breast cancer profiling through explainable machine learning.
\newblock {\em Nature Machine Intelligence}, 3:1--12, 04 2021.

\bibitem{guclu2015}
U.~G{\"u}{\c c}l{\"u} and M.~A.~J. van Gerven.
\newblock Deep neural networks reveal a gradient in the complexity of neural representations across the ventral stream.
\newblock {\em Journal of Neuroscience}, 35(27):10005--10014, 2015.

\bibitem{Cadena2019}
S.~A. Cadena, G.~H. Denfield, E.~Y. Walker, L.~A. Gatys, A.~S. Tolias, M.~Bethge, and A.~S. Ecker.
\newblock Deep convolutional models improve predictions of macaque v1 responses to natural images.
\newblock {\em PLoS Computational Biology}, 2019.

\bibitem{Neumann2019}
W.~J. Neumann, R.~S. Turner, B.~Blankertz, T.~Mitchell, A.~A. Kühn, and R.~M. Richardson.
\newblock Toward electrophysiology-based intelligent adaptive deep brain stimulation for movement disorders.
\newblock {\em Neurotherapeutics : the journal of the American Society for Experimental NeuroTherapeutics}, 16:105--118, 1 2019.

\bibitem{MATHIS20201}
M.~W. Mathis and A.~Mathis.
\newblock Deep learning tools for the measurement of animal behavior in neuroscience.
\newblock {\em Current Opinion in Neurobiology}, 60:1--11, 2020.
\newblock Neurobiology of Behavior.

\bibitem{Roscher_2020_insights}
R.~Roscher, B.~Bohn, M.~F. Duarte, and J.~Garcke.
\newblock Explainable machine learning for scientific insights and discoveries.
\newblock {\em IEEE Access}, 8:42200--42216, 2020.

\bibitem{10.1145/3567592}
S.~Ranathunga, E.-S.~A. Lee, M.~Prifti~Skenduli, R.~Shekhar, M.~Alam, and R.~Kaur.
\newblock Neural machine translation for low-resource languages: A survey.
\newblock {\em ACM Comput. Surv.}, 55(11), feb 2023.

\bibitem{pine-etal-2022-requirements}
A.~Pine, D.~Wells, N.~Brinklow, P.~Littell, and K.~Richmond.
\newblock Requirements and motivations of low-resource speech synthesis for language revitalization.
\newblock In {\em Proceedings of the 60th Annual Meeting of the Association for Computational Linguistics (Volume 1: Long Papers)}, pages 7346--7359, Dublin, Ireland, May 2022. Association for Computational Linguistics.

\bibitem{7780634}
L.~A. Gatys, A.~S. Ecker, and M.~Bethge.
\newblock Image style transfer using convolutional neural networks.
\newblock In {\em 2016 IEEE Conference on Computer Vision and Pattern Recognition (CVPR)}, pages 2414--2423, 2016.

\bibitem{10.1186/s13673-016-0063-4}
S.-G. Lee and E.-Y. Cha.
\newblock Style classification and visualization of art painting's genre using self-organizing maps.
\newblock {\em Hum.-Centric Comput. Inf. Sci.}, 6(1), dec 2016.

\bibitem{10.1007/978-3-319-46604-0_52}
B.~Seguin, C.~Striolo, I.~diLenardo, and F.~Kaplan.
\newblock Visual link retrieval in a database of paintings.
\newblock In G.~Hua and H.~J{\'e}gou, editors, {\em Computer Vision -- ECCV 2016 Workshops}, pages 753--767, Cham, 2016. Springer International Publishing.

\bibitem{Lang_art_similartiy_2018}
S.~Lang and B.~Ommer.
\newblock Attesting similarity: Supporting the organization and study of art image collections with computer vision.
\newblock {\em Digital Scholarship in the Humanities, Oxford University Press}, 33:845--856, 2018.

\bibitem{4586391}
M.~Panagopoulos, C.~Papaodysseus, P.~Rousopoulos, D.~Dafi, and S.~Tracy.
\newblock Automatic writer identification of ancient greek inscriptions.
\newblock {\em IEEE Transactions on Pattern Analysis and Machine Intelligence}, 31(8):1404--1414, 2009.

\bibitem{vane2016}
O.~Vane.
\newblock Using data visualisation to tell stories about cultural collections.
\newblock In {\em Proceedings of the 2017 CHI Conference Extended Abstracts on Human Factors in Computing Systems}, CHI EA '17, page 335–339, New York, NY, USA, 2017. Association for Computing Machinery.

\bibitem{schlag_coins_2017}
I.~Schlag and O.~Arandjelovic.
\newblock Ancient roman coin recognition in the wild using deep learning based recognition of artistically depicted face profiles.
\newblock In {\em 2017 IEEE International Conference on Computer Vision Workshops (ICCVW)}, pages 2898--2906, 2017.

\bibitem{shen2019discovery}
X.~Shen, A.~A. Efros, and M.~Aubry.
\newblock Discovering visual patterns in art collections with spatially-consistent feature learning.
\newblock In {\em Proceedings IEEE Conf. on Computer Vision and Pattern Recognition (CVPR)}, 2019.

\bibitem{monnier_doc_extractor_2020}
T.~Monnier and M.~Aubry.
\newblock docextractor: An off-the-shelf historical document element extraction.
\newblock In {\em 2020 17th International Conference on Frontiers in Handwriting Recognition (ICFHR)}, pages 91--96, 2020.

\bibitem{Tangherlini2013}
T.~R. Tangherlini and P.~Leonard.
\newblock Trawling in the sea of the great unread: Sub-corpus topic modeling and humanities research.
\newblock {\em Poetics}, 41(6):725--749, 2013.

\bibitem{jockers2013significant}
M.~L. Jockers and D.~Mimno.
\newblock Significant themes in 19th-century literature.
\newblock {\em Poetics}, 41(6):750--769, 2013.

\bibitem{schoch2021topic}
C.~Sch{\"o}ch.
\newblock Topic modeling genre: an exploration of french classical and enlightenment drama.
\newblock {\em arXiv preprint arXiv:2103.13019}, 2021.

\bibitem{koppel-etal-2016-reconstructing}
M.~Koppel, M.~Michaely, and A.~Tal.
\newblock Reconstructing ancient literary texts from noisy manuscripts.
\newblock In {\em Proceedings of the Fifth Workshop on Computational Linguistics for Literature}, pages 40--46, San Diego, California, USA, 06 2016. Association for Computational Linguistics.

\bibitem{YadavNisha2010Saot}
N.~Yadav, H.~Joglekar, R.~P.~N. Rao, M.~N. Vahia, R.~Adhikari, and I.~Mahadevan.
\newblock Statistical analysis of the indus script using n-grams.
\newblock {\em PloS one}, 5(3):e9506--e9506, 2010.

\bibitem{luo-etal-2019-neural}
J.~Luo, Y.~Cao, and R.~Barzilay.
\newblock Neural decipherment via minimum-cost flow: From {U}garitic to {L}inear {B}.
\newblock In {\em Proceedings of the 57th Annual Meeting of the Association for Computational Linguistics}, pages 3146--3155, Florence, Italy, July 2019. Association for Computational Linguistics.

\bibitem{10.1162/tacl_a_00354}
J.~Luo, F.~Hartmann, E.~Santus, R.~Barzilay, and Y.~Cao.
\newblock {Deciphering Undersegmented Ancient Scripts Using Phonetic Prior}.
\newblock {\em Transactions of the Association for Computational Linguistics}, 9:69--81, 02 2021.

\bibitem{asssome2022restoring}
Y.~Assael*, T.~Sommerschield*, B.~Shillingford, M.~Bordbar, J.~Pavlopoulos, M.~Chatzipanagiotou, I.~Androutsopoulos, J.~Prag, and N.~de~Freitas.
\newblock Restoring and attributing ancient texts using deep neural networks.
\newblock {\em Nature}, 2022.

\bibitem{Bekiari2021}
C.~Bekiari, G.~Bruseke, M.~Doerr, C.-E. Ore, S.~Stead, and A.~Velios.
\newblock Definition of the cidoc conceptual reference model v7.1.1.
\newblock {\em The CIDOC Conceptual Reference Model Special Interest Group}, 2021.

\bibitem{Bekiari2015}
C.~Bekiari, M.~Doerr, P.~L. Boeuf, and P.~Riva.
\newblock Definition of frbroo: A conceptual model for bibliographic information in object-oriented formalism, 2015.

\bibitem{Meghini2018}
C.~Meghini and M.~Doerr.
\newblock A first-order logic expression of the cidoc conceptual reference model.
\newblock {\em International Journal of Metadata, Semantics and Ontologies}, 13(2):131--149, 2018.

\bibitem{ElHajj2022}
H.~El-Hajj, M.~Zamani, J.~Büttner, J.~Martinetz, O.~Eberle, N.~Shlomi, A.~Siebold, G.~Montavon, K.-R. Müller, H.~Kantz, and M.~Valleriani.
\newblock An ever-expanding humanities knowledge graph: The sphaera corpus at the intersection of humanities, data management, and machine learning.
\newblock {\em Datenbank-Spektrum: Zeitschrift für Datenbanktechnologien und Information Retrieval}, 2022.

\bibitem{Kraeutli2021}
F.~Kr{\"a}utli, E.~Chen, and M.~Valleriani.
\newblock {\em Information and Knowledge Organisation in Digital Humanities}, chapter Linked data strategies for conserving digital research outputs, pages 206 -- 224.
\newblock Routledge, London, 2021.

\bibitem{brausch2023machine}
S.~Brausch and G.~Gra{\ss}hoff.
\newblock Machine learning for the history of ideas.
\newblock {\em Future Humanities}, 2023.

\bibitem{Pantin2013}
I.~Pantin.
\newblock Oronce finé mathématiien et homme du livre: la pratique éditoriale comme moteur d’évolution.
\newblock In I.~Pantin and G.~Péoux, editors, {\em Mise en forme des savoirs à la Renaissance. À la croisée des idées, des techniques et des public}, pages 19--40. Armand Colin, Paris, 2013.

\bibitem{FineLat1551}
O.~Finé.
\newblock {\em Sphaera mundi, sive cosmographia quinque recèns auctis \& emendatis absoluta}.
\newblock Michel Vascosan, Paris, 1551.

\bibitem{FineLat1552}
O.~Finé.
\newblock {\em Sphaera mundi, sive cosmographia quinque libris recèns auctis \& emendatis absoluta}.
\newblock Michel Vascosan, Paris, 1552.

\bibitem{FineLat1555}
O.~Finé.
\newblock {\em Orontii Finaei Delphinatis, regii mathematicarum Lutetiae professoris, de mundi sphaera, sive cosmographia, libri V}.
\newblock Michel Vascosan, Paris, 1555.

\bibitem{FineFrench1551}
O.~Finé.
\newblock {\em Le sphere du monde, proprement ditte cosmographie, composee nouvellement en francois, \& divisee en cinq livres}.
\newblock Michel Vascosan, Paris, 1551.

\bibitem{FineFrench1552}
O.~Finé.
\newblock {\em Le sphere du monde, proprement ditte cosmographie, composee nouvellement en francois, \& divisee en cinq livres}.
\newblock Michel Vascosan, Paris, 1552.

\bibitem{Hennen2022}
I.~C. Hennen.
\newblock Printers, booksellers, and bookbinders in wittenberg in the sixteenth century: Real estate, vicinity, political, and cultural activities.
\newblock In M.~Valleriani and A.~Ottone, editors, {\em Publishing Sacrobosco's De sphaera in Early Modern Europe. Modes of Material and Scientific Exchange}, pages 99--154. Springer, 2022.

\bibitem{Melanchthon1531}
J.~de~Sacrobosco and P.~Melanchthon.
\newblock {\em Liber Iohannis de Sacro Busto, de Sphaera. Addita est praefatio in eundem librum Philippi Melanchthonis ad Simonem Gryneum}.
\newblock Joseph Klug, Wittenberg, 1531.

\bibitem{Shcheglov:2004ul}
D.~Shcheglov.
\newblock Ptolemy's system 0f seven climata and eratosthenes geography.
\newblock {\em Geographia antiqua}, 13:21--37, 2004.

\bibitem{grasshoff2017living}
G.~Gra{\ss}hoff.
\newblock Living according to the seasons.
\newblock {\em Knowledge, Text and Practice in Ancient Technical Writing}, page 200, 2017.

\bibitem{Honigmann:1992tt}
E.~Honigmann and F.~Sezgin.
\newblock {\em Die sieben Klimata und die Poleis Episemoi : eine Untersuchung zur Geschichte der Geographie und Astrologie im Altertum und Mittelalter}.
\newblock Institute for the History of Arabic-Islamic Science, Frankfurt am Main, 1992.

\bibitem{dicks_1955}
D.~R. Dicks.
\newblock The ΚΛΙΜΑΤΑ in greek geography.
\newblock {\em The Classical Quarterly}, 5(3-4):248–255, 1955.

\bibitem{2014BlgAJ..20...68N}
M.~G. {Nickiforov}.
\newblock {Analysis of the calendar C. Ptolemy ``Phases of the fixed stars''}.
\newblock {\em Bulgarian Astronomical Journal}, 20:68, January 2014.

\bibitem{Peucer1558}
K.~Peucer.
\newblock {\em Elementa doctrinae de circulis coelestibus, et primo motu, recognita et correcta, autore Casparo Peucero}.
\newblock Johann Krafft the Elder, Wittenberg, 1558.

\bibitem{Cortes1556}
M.~Cortés.
\newblock {\em Breve compendio de la sphera y de la arte de navegar, con nueuos instrumentos y reglas, exemplificado con muy subtiles demonstraciones: compuesto por Martin Cortes natural de burjalaros en el reyno de Aragon y de presente vezino de la ciudad de Cadiz: dirigido al invictissimo Monarcha Carlo Quinto Rey de las Hespanas etc. Senor Nuestro}.
\newblock António Alvares, Seville, 1556.

\bibitem{Beyer1560}
J.~d. Sacrobosco and H.~Beyer.
\newblock {\em Quaestiones in libellum de Sphaera Ioannis de Sacro busto, in gratiam studiosae iuventutis olim in Academia, Vuitebergensi collectae, per Hartmannum Beyer, nunc emendatae \& auctae}.
\newblock Peter Braubach, Frankfurt am Main, 1560.

\bibitem{Witekind1590}
H.~Witekind.
\newblock {\em De sphaera mundi: Et Témporis ratione apud Christianos. Hermanni VVitekindi}.
\newblock Matthäus Harnisch, Neustadt an der Weinstraße, 1590.

\bibitem{Dietrich1591}
S.~Dietrich.
\newblock {\em Novae quaestiones sphaericae, hoc est, de circulis coelestibus \& primo mobili, in gratiam studiosae iuventutis scriptae, a M. Sebastiano Theodorico Vuinshemio. Mathematum Professore}.
\newblock Matthaeus Welack, Wittenberg, 1591.

\bibitem{Sacrobosco1495}
J.~d. Sacrobosco.
\newblock {\em Opusculum Johannis de sacro busto spericum cum notabili commento atque figuris textum declarantibus utilissimis}.
\newblock Martin Landsberg, Leipzig, 1495.

\bibitem{Barozzi1598}
F.~Barozzi.
\newblock {\em Cosmographia in quatuor libros distributa, summo ordine, miraque facilitate, ac brevitate ad Magnam Ptolemaei Mathematicam Constructionem, ad universamque Astrologiam instituens: Francisco Barocio, Iacobi Filio, Patritio Veneto autore. Cum Prefatione eiusdem Authoris, in qua perfecta quidem Astrologiae Divisio, \& enarratio Aurorum illustrium, \& voluminum ab eis conscriptorum in singulis Astrologiae partibus habetur: Ioannis de Sacrobosco verò 84 errores, \& alij permulti suorum expositorum, \& sectatorum ostenduntur, rationibusque redarguuntur. Precesserunt etiam quaedem Communia Mathematica, necnon Arithmetica \& Geometrica principia, nonnulleque Propositiones, de quibus in toto opere saepe sit mentio: Ac demum locupletissimus Index eorum, que ipsa Cosmographia continentur. Omnia nuper in hac secunda editione ab ipso Autore diligenter recognita, multisque in locis aucta}.
\newblock Grazioso Percacino, Venice, 1598.

\bibitem{Pantin2021OP}
I.~Pantin.
\newblock Lire le ciel dans les poèmes anciens. le \textit{De ortu poetico} et la pédagogie de melanchthon.
\newblock In M.~C. de~Ribes, S.~Dembruk, D.~Fliege, and V.~Oberliessen, editors, {\em ‘Une honnête curiosité de s’enquérir de toutes choses’. Mélanges en l'honneur d'Olivier Millet}, pages 373--384. Droz, Genève, 2021.

\bibitem{Blebel1582}
T.~Blebel.
\newblock {\em De sphaera et primis astronomiae rudimentis libellus ad usum Scholarum maximè accomodatus: accurata methodo \& brevitate conscriptus, ac denuo editus. A M. Thoma Blebelio Budissino}.
\newblock Johann Krafft, Wittenberg, 1582.

\bibitem{10.2307/20488785}
M.~H. Close.
\newblock Hipparchus and the precession of the equinoxes.
\newblock {\em Proceedings of the Royal Irish Academy (1889-1901)}, 6:450--456, 1900.

\end{thebibliography}
\normalsize

\section*{Data and Code Availability}
The  Sacrobosco Tables dataset can be accessed via \protect\href{https://zenodo.org/record/5767440}{https://zenodo.org/record/5767440}. All further data including annotated ground truth data for training and evaluation are available upon request. Full code for reproduction of our results as well as pre-trained models and demonstrations are available upon request. 

\section*{Acknowledgements}
This work was partly funded by the German Ministry for Education and Research (under refs 01IS14013A-E, 01GQ1115, 01GQ0850, 01IS18056A, 01IS18025A and 01IS18037A) and BBDC/BZML and BIFOLD.
Furthermore KRM was partly supported by the Institute of Information \& Communications Technology Planning \& Evaluation (IITP) grants funded by the Korea Government (MSIT) (No. 2019-0-00079, Artificial Intelligence Graduate School Program, Korea University and No. 2022-0-00984, Development of Artificial Intelligence Technology for Personalized Plug-and-Play Explanation and Verification of Explanation). Finally, the Sphere project is also supported by the Max Planck Institute for the History of Science.
We would like to thank Olya Nicolaeva, Tilman Kemeny and Stephan Tietz for their support in the early phase of this research especially concerning the organization of the training set. We also would like to thank Nana Citron, Beate Federau, and Victoria Beyer for their help in cleaning the data. Finally, our gratitude goes to Lindy Divarci for her support in managing the publication process.

\section*{Author Contributions}
O.\ E., J.\ B. and G.\ M. developed the ML software. O.\ E., J.\ B, M.\ V. and H.\ H. performed the data analysis. M.\ V. and J.\ B. performed the historical interpretation. H.\ H. managed the data repository and performed data curation; O.\ E., G.\ M., J.\ B., M.\ V., K.-R.\ M. conceptualized the experiments and discussed the results. O.\ E., G.\ M., J.\ B.,  H.\ H., M.\ V. and K.-R.\ M. wrote the manuscript. All authors read and approved the manuscript. Co-corresponding authors are M.\ V. and K.-R.M.
\end{multicols}

\newpage

\setcounter{section}{-1}\stepcounter{section}
\setcounter{table}{0}
\setcounter{figure}{0}
\renewcommand{\thetable}{S\arabic{table}}
\renewcommand\thefigure{S\arabic{figure}}
\renewcommand{\theHtable}{Supplement.\thetable}
\renewcommand{\theHfigure}{Supplement.\thefigure}

\appendix
\renewcommand{\thesection}{\Alph{section}}

\section*{Supplementary materials}

\section{Materials and Methods}
\subsection{The Sacrobosco Collection from the \textit{Sphaera} Corpus}\label{text:suppl_sacro_collection}

\begin{figure}[h!]
\centering
\includegraphics[width = 0.9\linewidth]{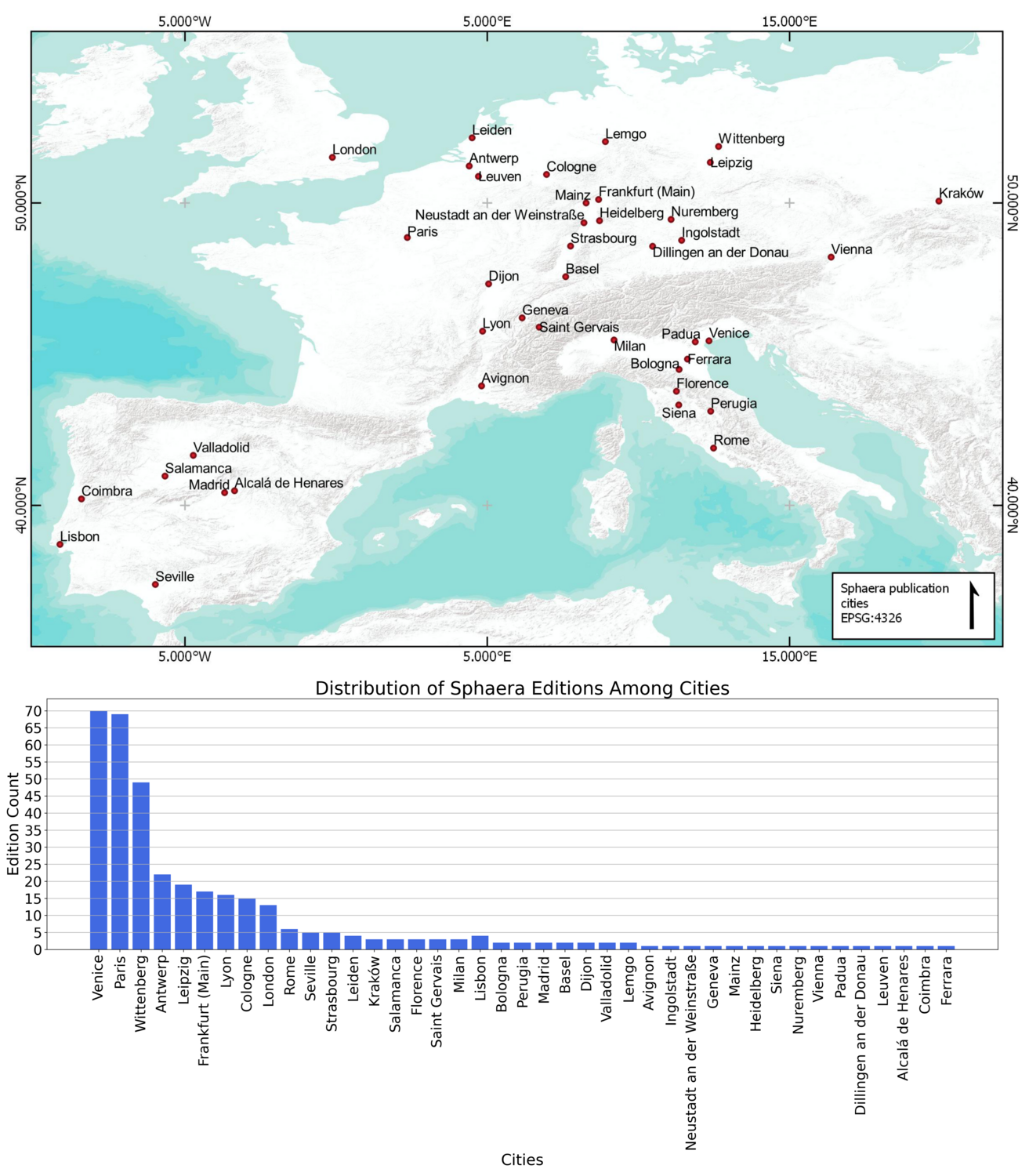}
\caption{Geographical distribution of the \textit{Sphaera} editions.}
\label{fig:sphaera_geo}
\end{figure}

\begin{figure}[h!]
    \centering
    \includegraphics[width = 1\linewidth]{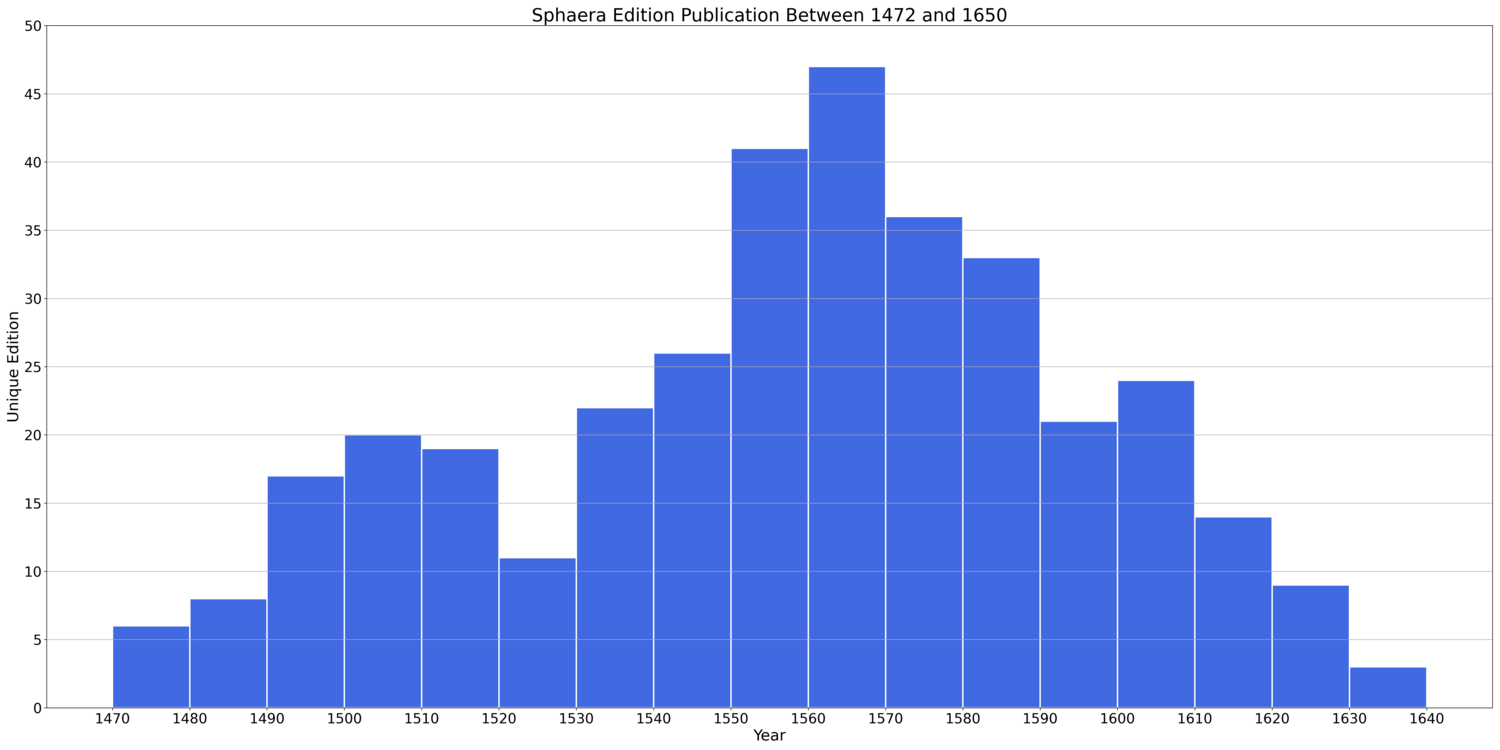}
    \caption{Publication rate of \textit{Sphaera} textbooks between 1475 and 1650.}
    \label{fig:publication_rate}
\end{figure}

The \textit{Sphaera} corpus contains four collections. One of them is called ``Sacrobosco''. It  is composed of 359 different editions of printed textbooks used across European universities to teach the introductory class on geocentric cosmology and astronomy during the early modern period. These 359 editions were published between 1472, the year of the first print (and of the first print of a scientific, mathematical text ever), and 1650, which marks the decline of geocentric astronomy, almost a 100 years after the publication of Nikolaus Copernicus's \textit{De revolutionibus orbium coelestium} in 1543, which introduced a mathematical system based on a heliocentric worldview to early modern academia. The Sacrobosco Collection is composed of $\approx 76,000$ pages. If we consider a realistic print-run of about 1000 exemplars for each edition \cite{Gingerich1988, Gingerich1990}, the collection under examination here represents ca. 350.000 textbooks that were circulating during the time period of at least 180 years and were used by students and lecturers in a geographic area that extends from Krakow to Lisbon and from London to Rome (Figs. \ref{fig:sphaera_geo},\ref{fig:publication_rate}).\par

All the collected editions are related to one specific text: the \textit{Tractatus de sphaera} by Johannes de Sacrobosco (-- 1256). This text was originally compiled and published in Paris in the first half of the thirteenth century. As an elementary text on geocentric cosmology, the tract was used in astronomy classes of almost all European universities during the first year of the curriculum. These classes were mandatory for all students, regardless of their ultimate field of study, because astronomy as a discipline belonged to the quadrivium. The quadrivium represented the curriculum of studies  that any student had to accomplish during the first years at the universities in order to be allowed to gain access to further curricula, such as medicine or jurisprudence or theology. Despite the relative simplicity of the treatise's content, its importance to understanding the evolution of knowledge stems from the fact that it was used from the thirteenth to the seventeenth century and was subject to continuous transformations and modifications, by means of commentaries and further texts that were placed or printed together and which deepened more specific, related subjects. This motivates to use  this particular collection to investigate the broader mechanisms of knowledge evolution during this period. We rely solely on printed editions of textbooks that contain the \textit{Tractatus de sphaera} in order to construct a structured and systematic dataset for the computational analyses discussed here. \par

Focusing the research on university textbooks means that the present work examines processes of scientific transformation on a large scale concerning the dominant knowledge of the educated society of early modern Europe. In other words, the corpus under examination reveals the knowledge possessed by those who became the readers of seminal works such as those of Copernicus and Galileo. It reveals their background knowledge and how this changed over time.\par

\begin{figure}[h!]
    \centering
    \includegraphics[width = 1\linewidth]{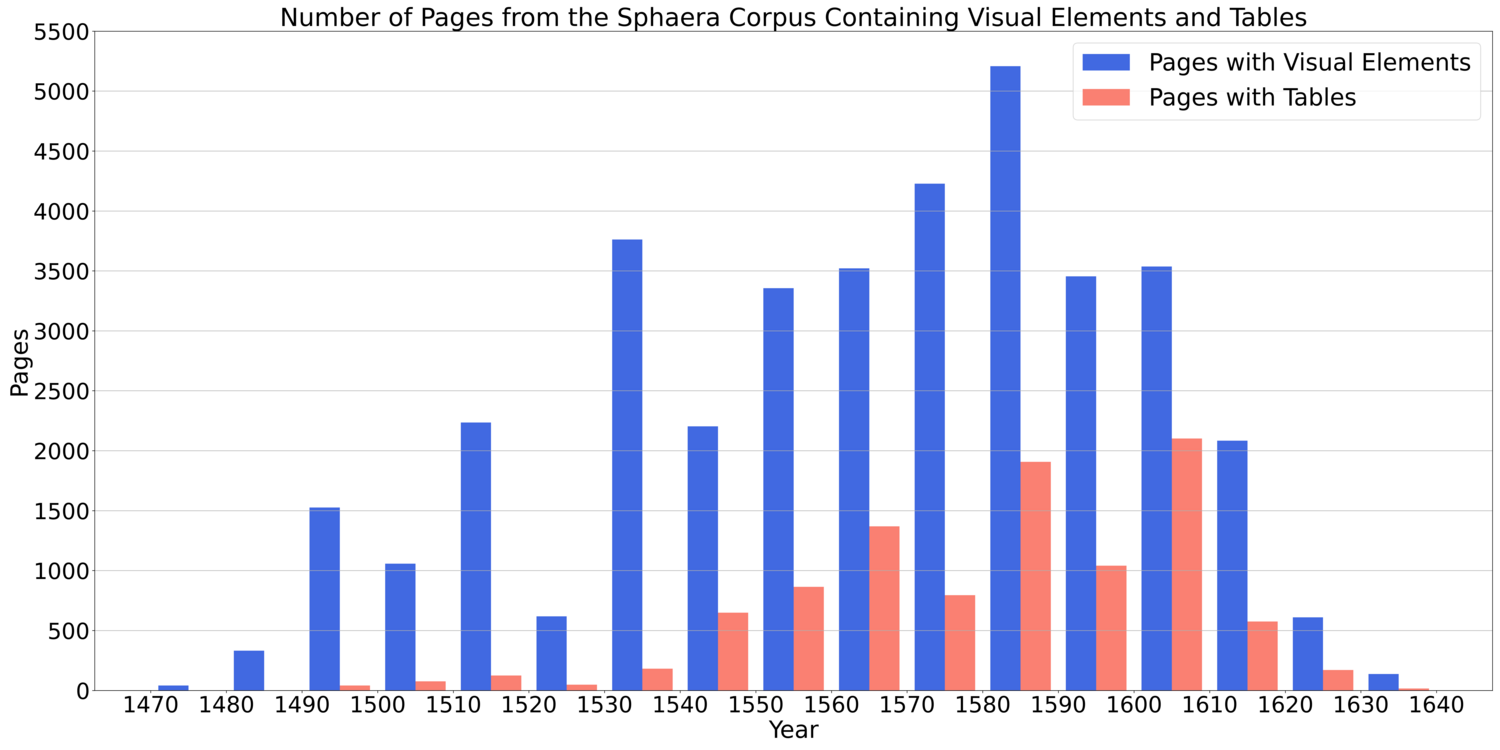}
    \caption{Histogram showing the variation of the number of pages containing visual elements and computational tables in the Sacrobosco Collection.}
    \label{fig:hist_ill_table}
\end{figure}

In general, we suggest a corpus analysis that follows three different axes, which can be re-aggregated at the end. The three axes emerge as based on three different types of data, into which we de-compose and dissect the historical sources. We call these different kinds of data ``knowledge atoms''. These are ``text-parts'', ``visual elements'' such as scientific diagrams and illustrations, initials, and printers' devices, and ``computational tables'' represented by numerical and alpha numerical tables, most of which resulting from calculations following astronomic computational workflows. In the case of our collection, the \textit{Alfonsine tables} were the basis for many of these calculations \cite{ChabasGoldstein2003}. The collection page statistics shown in Fig. \ref{fig:hist_ill_table} highlight how book production varies over time and, more specifically, how table pages have increasingly been included as part of standard text books. The present work focuses on the investigation of the last of these knowledge atoms, namely on the tables, a kind of document that, because of its complexity could not hitherto be analyzed in great quantities either by humans or by machines. As it will be shown, focusing on the computational astronomic tables means investigating the process of mathematization of astronomy as taught at European universities during the early modern period.\par

The great variety of computational astronomic tables in the  collection considered here informs our modeling approach in the present work and enables us to analyze and reconstruct scientific knowledge as disclosed and externalized by such tables. Before moving to this main subject, however, we briefly sketch the historical results already achieved on the basis of the other knowledge atoms while the data infrastructure needed to execute such research is described in the section \ref{text:infrastructure}. This overview concerning the results of previous researches is necessary to understand the implications of the results presented in this work. The dataset is retrieved from the research project `The Sphere. Knowledge System Evolution and the Shared Scientific Identity of Europe'\footnote{\href{https://sphaera.mpiwg-berlin.mpg.de}{https://sphaera.mpiwg-berlin.mpg.de}}.\par

\subsubsection{Studying Knowledge Systems}\label{text:knowledge_systems}
The dataset described in this section is the backbone of what we consider a knowledge system. Such a system results from the re-integration of the identified knowledge atoms into diachronic and synchronic graphs. We first describe the taxonomy used to categorize the 359 editions and then those graphs constituted by the knowledge atom `text-part', which describes self-contained text sections in a book.\par

\begin{figure}[h!]
    \centering
    \includegraphics[width = 0.7\linewidth]{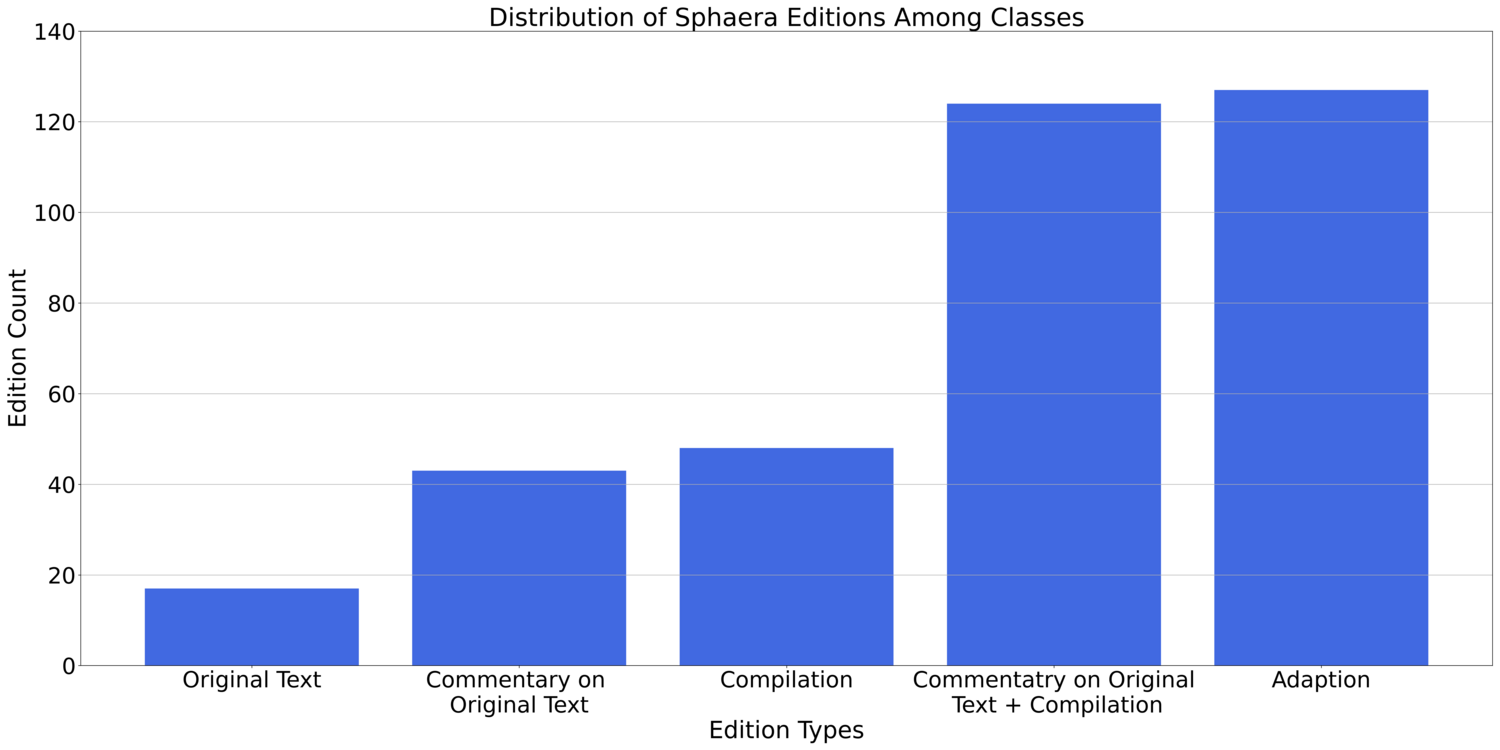}
    \caption{Taxonomy for the editions constituting the Sacrobosco Collection: editions that contain the original medieval tract only; those that contain the original treatise with at least one commentary; those that contain the original treatise and other treatises (compilations); those that contain the original treatise, at least one commentary, and other texts; adaptions.}
    \label{fig:Taxonomy_Editions}
\end{figure}

The rigorous historical analyses that form the foundation of the research resulted in the identification of five different edition classes within the collection, clearly differentiated by the form of their content in such a way to allow the identification of the modes of knowledge production in the period examined here (Figure \ref{fig:Taxonomy_Editions}).\par

The ``original treatises'' class represents a total of 17 editions, which exclusively contain the original text of the \textit{Tractatus de sphaera} without added contemporary commentaries. The 48  editions, classified as ``annotated original treatises,'' contain the original work of Johannes de Sacrobosco, with additional commentaries by various authors. As ``Compilation of texts'',  we define a class of 43 editions, which include the original \textit{Tractatus de sphaera} along with other original treatises by various authors, while the class ``compilation of texts and annotated originals'' contains 124 editions which include a commented or annotated \textit{Tractatus de sphaera} along with other treatises. The final and largest class is constituted by editions defined as ``adaptions'', which numbers 127 and displays texts that are strongly influenced by the content and structure of the \textit{Tractatus de sphaera}, but do not include the original treatise itself. \par

Each of these editions is dissected into text-parts. Each text-part represents a textual component that is both larger than a single paragraph and also conveys a coherent body of information. These text-parts are then classified into two main categories, 322 ``content'' and 261 ``paratext'' text-parts, the former referring to text-parts containing scientific treatises, while the latter refers to short texts that are often added to original content, containing poetry, letters to the reader or prefaces, dedication letters, or other literary compositions useful to understand the social, institutional, and political context in which the editions were conceived and produced \cite{RN2520}. We built graphs (both diachronic and synchronic) among the editions on the basis of semantic relations among the text-parts that they contain.\par

To build a synchronic graph on the basis of the text-parts, we performed a content-related analysis in order to assess their mutual semantic relations: We related the text-parts to each other using the relationships ``commentary of'', ``translation of,'' and ``fragment of''. The diachronic graph is instead represented by the re-occurrences of text-parts over time. The integration of both graphs creates a high-dimensional matrix that, by adding the available historical metadata, allowed to establish the multiplex networks by means of which we investigated the emergence of epistemic communities within the corpus \cite{mva19, SRN2020}\footnote{To interactively explore the dynamic of re-occurrence of the text-parts also according to their mutual semantic relationships, see \href{https://sphaera.mpiwg-berlin.mpg.de/adoption}{https://sphaera.mpiwg-berlin.mpg.de/adoption}}. \par

The first and most fundamental result of our previous network analyses concerns the process of homogenization of knowledge and, specifically, the underlying mechanism, which we now can best describe as a mechanism of imitation \cite{RN2923, VallerianiOttone2022, Kikuchi2022}. We were able to identify families of treatises characterized by their inherent text-parts similarity, while at the same time they executed a strong influence -- their content was imitated -- on the content of other treatises produced elsewhere. By matching this analysis with the metadata, we were finally able to find out that the dominant family of treatises that gave birth to such a process was produced in the reformed Wittenberg during the 30s of the sixteenth century. Assessing all the reasons that brought the scientific production of Wittenberg into the sights of European scholars of the period remains a complex task and, as it will be shown, the present work represents a fundamental step forward in the understanding of this complex process. At this stage, however, it can be stated that while the Protestant Reformation created a confessional, institutional, and political division in Europe, it also created the backdrop against which scientists made their first step toward the formation of a community that begins to show some of the traits characteristic for the modern international scientific society. Other editions that could be identified and that we defined as ``Enduring innovations'' and ``Great transmitters'', show the relevance of Wittenberg, especially around the middle of the sixteenth century \cite{SRN2020}. At this point Wittenberg changed its strategy, moving from a more radically innovative position toward integrating innovations and tradition in a way that would have supported the primacy of Wittenberg's scientific literary output in Europe for many decades, furnishing therefore the fuel for a long-term process of homogenization. In conclusion, we were already able to show that, at the end of the sixteenth century, based on a mix of imitation and a center-emanating output of innovations, students across Europe were all learning the same astronomy and cosmology, at least for what concerns the scientific knowledge conveyed through the textual apparatus of the textbooks under investigation.\par

But the textual apparatus is not the only means used to convey knowledge in the textbooks. During the early modern period, written text was considered highly authoritative. Science was produced mostly by commenting on older texts, be these medieval as in the case of the Sacrobosco collection or from classical Greek or Roman antiquity. The texts of reference, which were commented on,  were usually not changed or updated but they were illustrated. With regard to  the visual apparatus the situation was different. Since the late middle ages, the use of visualization started becoming increasingly prominent in Western science, a trend that continues to the present day. While medieval manuscripts of Sacrobosco's \textit{De sphaera} rarely display more than five illustrations, early modern editions developed a visual apparatus that was constituted of 40 to 50 illustrations, in certain extreme cases even more than 70. While our research focused on the visual apparatus is ongoing \cite{RN2949,RN2852,RN2922,el-HajjetalRevPap2022,Kraeutli2020}, traditional analyses seem to indicate that the Wittenberg production of textbooks was able to take the lead in the process of homogenization of knowledge, also concerning the scientific visual apparatus \cite{Pantin2020, Limbach2022}.\par

Finally, the third sort of knowledge atom, the numerical table, is the one the present work is pivoted around and, therefore, will be introduced in a separate section.\par

\subsubsection{Numerical Tables and Their Role}\label{text:table_meaning}

\begin{figure}[h!]
    \centering
    \includegraphics[width = 0.5\linewidth]{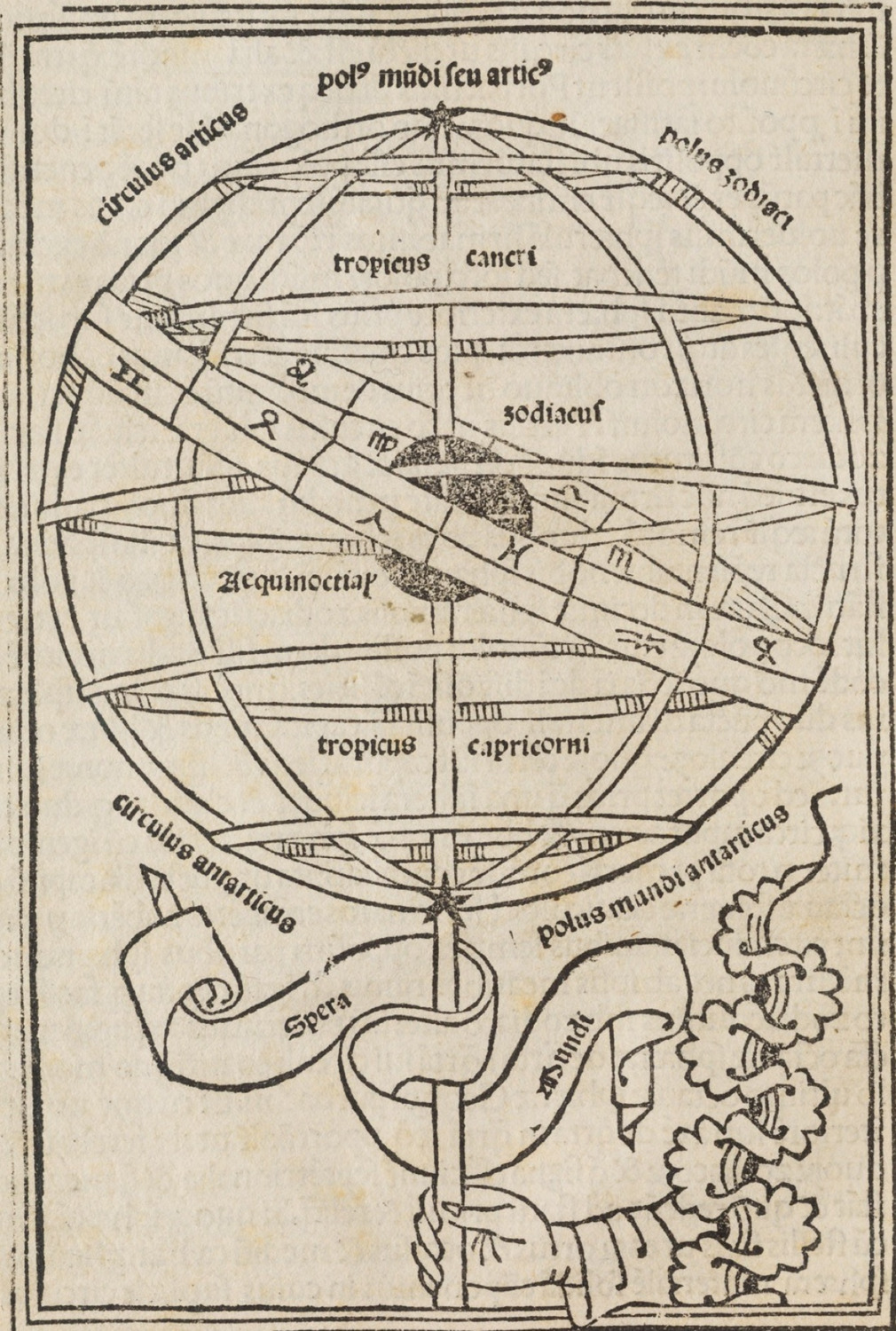}
    \caption{Typical graphic representation of an armillary sphere in a \textit{De sphaera} textbook. An armillary sphere is a mechanical representation of the geocentric cosmos and, at the same time, a scientific instrument. From \cite[sign. a-III-2]{Sacroboscoetal1490}. Courtesy of the Library of the Max Planck Institute for the History of Science.}
    \label{fig:Armillary_Sphere_1490_Scoto}
\end{figure}

The specific treatise around which the Sacrobosco Collection is centered (Sacrobosco's \textit{De sphaera}) is a \textit{qualitative} introduction to geocentric astronomy. Qualitative means that students could learn the composition and the elements of the cosmos, in certain cases also by working with the corresponding mechanical device, the armillary sphere (Figure \ref{fig:Armillary_Sphere_1490_Scoto}). Finally they apprehended fundamental notions concerning the movements of the celestial bodies: for instance that the outer sphere, the sphere of the fixed stars (firmament) moves from east to west on a daily basis and from west to east by about one degree every 100 years. What they could not learn by any means from this text was for instance to calculate in advance the position of a celestial body, for example a planet. This fundamental treatise, which remained in use at nearly all European universities for about 400 years, was \textit{not} an introduction to mathematical astronomy. During the thirteenth and fourteenth centuries, the period before the one considered here, only very few scholars had the chance and the skills to enter the realm of mathematical astronomy through the study of extremely difficult and rare works such as Ptolemy's \textit{Almagest}. Outside this expert culture astronomy was fundamentally non-mathematical; it was part of natural philosophy, which was essentially the result of a speculative  search for causes of natural phenomena. Astronomy, like the other disciplines of the quadrivium (geometry, arithmetic, and music) was considered as mathematical discipline but, in the general cultural context of the Middle Ages, apart from the fact that only few scholars really accessed such mathematical knowledge, the mathematical apparatus of astronomy was considered only as an instrument for calculations and not a method to describe the real world, only its appearance. Mathematical astronomy was not natural science.\par

The path toward modern science can be interpreted therefore also as a process of \textit{mathematization}. Practical knowledge, for instance, such as the knowledge accumulated by specialized artisans and engineers in the frame of mechanics and machine building, was integrated to mathematics and gave rise to theoretical mechanics starting from the sixteenth century. It was thus for instance from the integration of the practical knowledge of the artillerists and geometry that the new science of ballistics emerged during the sixteenth century \cite{Buettner2017}. In the case of the so-called mathematical disciplines, the process of mathematization was realized following two different directions simultaneously \cite{VallerianiQuad2022}.\par

On one side, the disciplines themselves evolved. Contrary to what is commonly believed, the above mentioned studies have demonstrated that the geocentric worldview was not a stagnant scientific theory but rather a subject of lively debate. A  myriad of observational data collected since antiquity still needed to find an appropriate theoretical framework. This dynamic led to the identification of specific sub-areas of study -- for instance nautical astronomy -- , which in turn resulted in the creation of new textbooks. These texts were designed to be more accessible and focused on teaching not an all-encompassing mathematical system for the cosmos, but rather the individual aspects of it,  such as the movements of each single planet or of only the outer sphere of the stars. These new texts -- most famously among them those entitled \textit{Theoricae planetarum} by means of which students could  learn a mathematical treatment of the orbits of each planet but disjointed from the general view of the cosmos -- actually were new text-parts added to the original tract of Sacrobosco. They lowered the threshold of access to mathematical knowledge in astronomy and kept the traditional texts as relevant introductory text valid for centuries.\par

The lowered threshold complemented the second direction. The latter is due to the emergence and increasing relevance of the universities, a genuine late medieval innovation in the framework of educational institutions. The late medieval and early modern universities linked disciplines that were not connected in such a systematic way in the previous centuries. Particularly relevant for astronomy was, for instance, the increasing integration with medicine, which was to a good part the result of the reception of Islamicate science. Largely due to a revival of Galen’s theory of critical days \cite{Cooper2004}, astrological medicine became a fundamental scientific and cultural component of European society. As soon as sickness occurred, physicians were required to know the positions of the planets on the day of appearance of the sickness in order to be able to deliver a suitable prognosis. They were therefore very accomplished in using the \textit{Theoricae} and its volvelles, paper instruments to determine positions of celestial bodies, to make precise calculations backwards in time.\par

Cultural trends like the one just described increased the demand for  a mathematical approach to astronomy. This trends resulted in a process of mathematization of astronomy phenomenologically characterized by the fact that an increased number of aspects of mathematical astronomy was taught to an increased number of people. This process is inherently connected to the homogenization of scientific knowledge, as is clearly demonstrated if, for instance, two pairs of editions of textbooks on astronomy, one from the fourteenth and one from the seventeenth century, are compared. What however remains unclear  is how exactly such process of mathematization worked, which kind of mathematics was really involved, what came first and how was it developed, whether all the attempts to introduce mathematical astronomy in a standard curriculum were successful, who promoted such process, when and where. This process has never been reconstructed in its details concerning history of astronomy --that is, leaving aside history of arithmetic and geometry -- and the reason for that is that the historical sources that can mainly disclose to us such a process could not be analyzed systematically until now. These sources consist of thousands of numerical tables, namely computational astronomic tables that were printed in the textbooks. In practice, we need to (1) identify recurring instances of particular tables across all printed editions, and (2) observe diachronic and synchronic trends in the inclusion of tables in the editions, averaged over the entire collection.\par

\subsubsubsection[]{Example of a Computational Astronomic Table}
\label{sec:right_asc_comp}

\begin{figure*}[h!]
    \centering
    \includegraphics[width=0.8\textwidth]{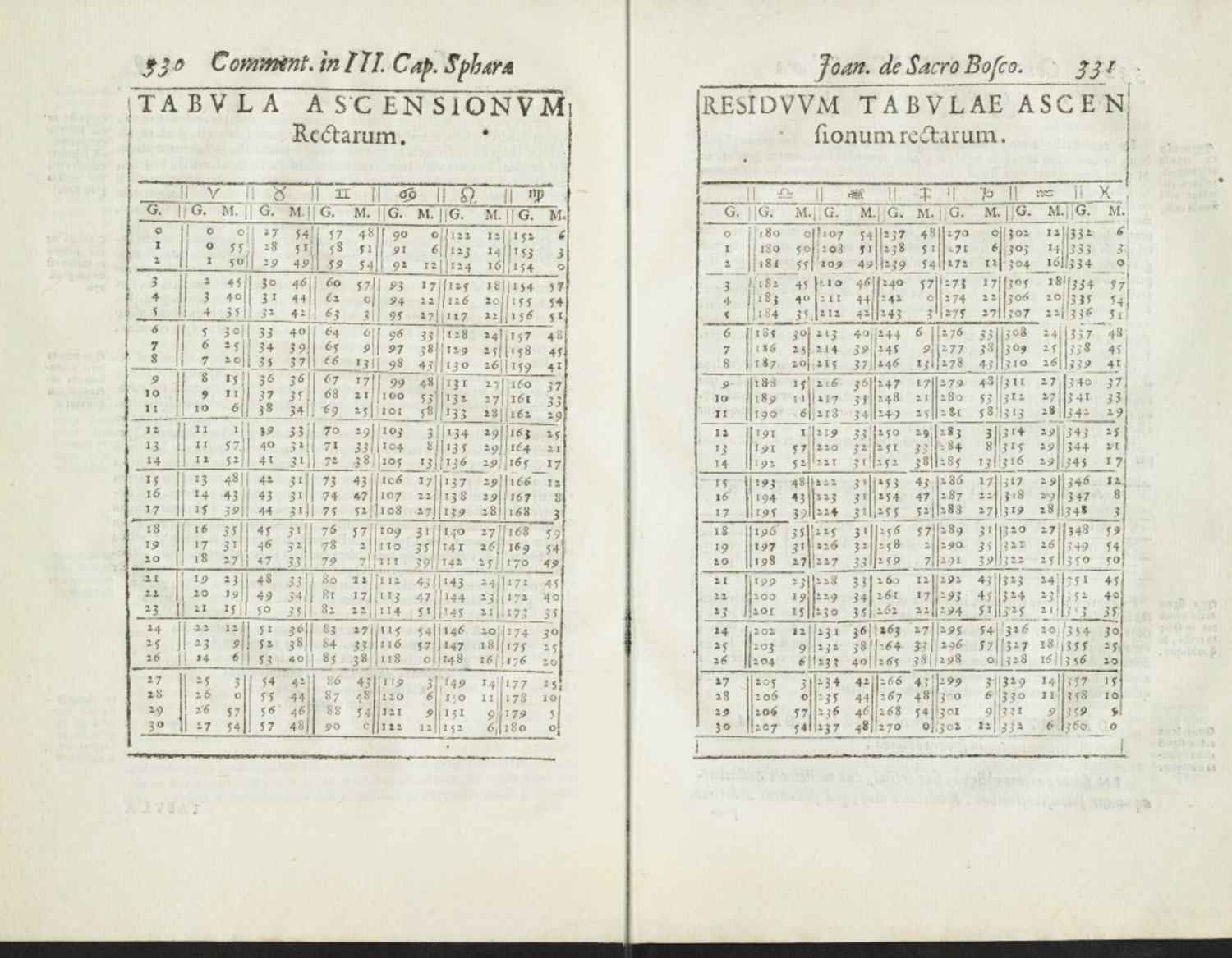}
    \caption{\textbf{Table of `right ascensions' as an example of a computational astronomical table.} This table is taken from \cite[530]{Clavius1585} published in 1585. Many exemplars of this table are contained in the collection. Courtesy of the Library of the Max Planck Institute for the History of Science.
    }
    \label{fig:example_computational_table}
\end{figure*}

In Figure \ref{fig:example_computational_table}, we present an example of a computational astronomic table which is frequently encountered in the collection. This table of the `right ascension' gives the degree of the celestial equator measured from the vernal equinox eastward that rises together with each degree of the ecliptic in the `right sphere', i.e. for an observer at the equator of the earth \cite[24–28]{ChabasGoldstein2012}. Positions on the ecliptic are specified by degrees into the signs, with each sign listed in separate column of the table. Counting begins with the beginning of Aries. Thus for instance, 10 degrees into Taurus would correspond 40 degrees along the ecliptic from the beginning of Aries.\par

The relation between the equatorial latitude and the celestial latitude of a point on the ecliptic was  derived by means of spherical geometry. The computational workflow in the table's background can be expressed in modern notation as: 
\begin{align*}
\alpha = \arctan(\cos(\epsilon)*\sin(\lambda)/\cos(\lambda)),    
\end{align*}
\noindent where  $\epsilon$  denotes the angle of the ecliptic, $\lambda$ is the angle along the ecliptic and the right ascension, i.e. the angle along the equator is given by $\alpha$.

It is relevant to note that the vernal equinox coincided with this first point of Aries in antiquity. Hipparchus defined this point, also known as the Cusp of Aries, as the reference point for specifying celestial equatorial longitude (even though the vernal equinox entered Aries only approx. 100 years after Hipparchus' death). Due to the procession of the equinoxes the vernal equinox wanders about 1 degree along the ecliptic in 72 years. Thus in the sixteenth century the vernal equinox point would have been about half way  into Pisces and, strictly speaking, the tables in Figure \ref{fig:example_computational_table} give the right ascension for an ancient observer in the first century BCE and are presented in Table \ref{table:right_asc}.

\begin{table}[h!]
\centering
\small
\begin{tabular}{l|lll}
   & \aries & \taurus & \gemini \\
   \midrule
1  & 0  55                 & 28  51                 & 58  51                 \\
2  & 1  50                 & 29  49                 & 59  54                 \\
3  & 2  45                 & 30  47                 & 60  57                 \\
4  & 3  40                 & 31  44                 & 61  60                 \\
5  & 4  35                 & 32  42                 & 63  3                  \\
6  & 5  30                 & 33  40                 & 64  6                  \\
7  & 6  25                 & 34  39                 & 65  10                 \\
8  & 7  21                 & 35  37                 & 66  13                 \\
9  & 8  16                 & 36  36                 & 67  17                 \\
10 & 9  11                 & 37  35                 & 68  21                 \\
11 & 10  6                 & 38  34                 & 69  25                 \\
12 & 11  2                 & 39  33                 & 70  29                 \\
13 & 11  57                & 40  32                 & 71  34                 \\
14 & 12  53                & 41  32                 & 72  38                 \\
15 & 13  48                & 42  31                 & 73  43                 \\
16 & 14  44                & 43  31                 & 74  47                 \\
17 & 15  40                & 44  31                 & 75  52                 \\
18 & 16  36                & 45  32                 & 76  57                 \\
19 & 17  31                & 46  32                 & 78  2                  \\
20 & 18  27                & 47  33                 & 79  7                  \\
21 & 19  24                & 48  33                 & 80  12                 \\
22 & 20  20                & 49  34                 & 81  17                 \\
23 & 21  16                & 50  35                 & 82  22                 \\
24 & 22  13                & 51  37                 & 83  28                 \\
25 & 23  9                 & 52  38                 & 84  33                 \\
26 & 24  6                 & 53  40                 & 85  38                 \\
27 & 25  3                 & 54  42                 & 86  44                 \\
28 & 25  60                & 55  44                 & 87  49                 \\
29 & 26  57                & 56  46                 & 88  55                 \\
30 & 27  54                & 57  48                 & 90  -0
\\
\end{tabular}
\caption{\textbf{Rendition of the first three columns of the table of the right ascension, calculated according to the modern formula.} The angle used for the obliquity of the ecliptic is 23.5 degrees. There is an excellent correspondence to the values in the table given in Figure \ref{fig:example_computational_table}.}
\label{table:right_asc}
\end{table}

\subsubsection{From Individual Tables to Corpus-Level Analysis: Assessing Similarity}\label{supplement:sec:scaling_tables}

\begin{figure*}[h!]
    \centering
    \includegraphics[width=1.\textwidth]{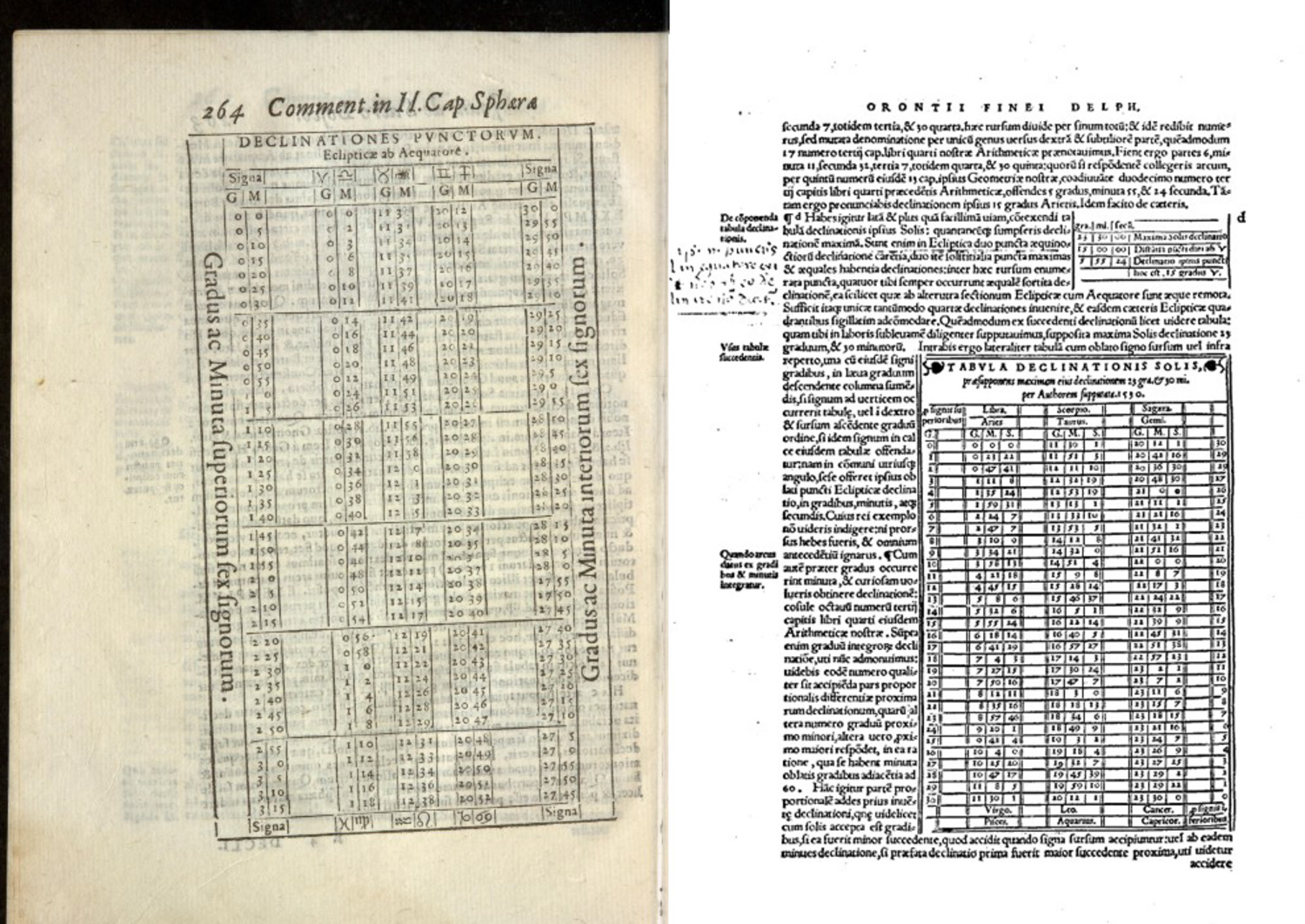}
    \caption{\textbf{Table of the declination of the Sun.} Left: Page taken from \cite[264]{Clavius1591} published in 1591 (table continues over the next 8 pages). Right: Page from \cite[Fo. III-v]{Fine1532} published in 1532. Left: Courtesy of the Library of the Max Planck Institute for the History of Science. Right: Public Domain, Google-digitized.
    }
    \label{fig:example_two_tables}
\end{figure*}

Judging whether two tables are similar in the sense that they express basically the same information is a complicated and time consuming process which can only be accomplished by experts. As an example in Figure \ref{fig:example_two_tables}, we provide two different  versions of a table of the declination of the Sun with respect to the celestial equator. The relation expressed in this table is the angular distance of points on the ecliptic to the celestial equator. As can be read off the first row, the table, like in the case discussed in \ref{sec:right_asc_comp} is completed under the assumption that the vernal equinox coincided with this first point of Aries based on comparable mathematical relation derived from spherical trigonometry.\par

While expressing the same astronomical relation, there are some substantial differences between the two tables in Figure \ref{fig:example_two_tables} expressing this same relation. While the table on the right, taken from an edition of Oronce Finé covers merely one page, the table from which we show one page on the left and which was taken from an edition of Christophorus Clavius stretches over altogether nine pages. The reason for this is that Clavius lists the declination for corresponding points on the ecliptic for steps of 5 arc minutes along the celestial equator while the step size in Fine's table is of one full degree. Thus only every 12th value in Clavius table corresponds to a value in Finé's table explaining why the former used up so much more space than the latter. Somewhat anachronistically speaking both tables list arguments and function values for the same function but the step-size in which the argument progresses is much smaller in one case.\par

This is, however, not the only difference between the tables. While Clavius specifies the declination in degrees and minutes, Finé, in addition to this, also adds arc seconds. Clavius thus for example gives a declination of 0 degrees 24 minutes for the point one degree into Aries, where Fine gives 0 degrees 23 minutes and 22 seconds. It is thereby somewhat surprising that Clavius, who obviously aims for higher precision using the smaller step-size, provides the more coarsely rounded results for the declinations. Moreover, Clavius value is obviously not attained by rounding the value to be found in Finé, and we can infer that both values and thus in essence both tables resulted from separate, independent calculations.\par

This example has highlighted the analytical effort and level of expertise that can be required to asses if and in which sense two tables are similar and made clear that such effort is  indeed unattainable in a collection like ours with thousands of tables implying a myriad of comparisons. An expert would then need to carefully inspect each of the individual digits composing the table. But even before this step the tables would first have to be identified via a manual lookup of the $\approx 76,000$ pages of the Sacrobosco Collection. The required analysis can now for the first time  be to a large extent  automated or facilitated by the use of machine learning. By means of a page classifier described below in section \ref{supplement:subsec:page_classifier}, we first were able to identify $\approx 10,000$ pages containing tables, which we also refer to as the \textit{Sacrobosco} Table corpus. This implies that a manual assessment of table similarity would require a meticulous examination of each table content from which similarity scores can subsequently be computed, or up to $10,000 \times 10,000$ manual pairwise table comparisons for an optimal result. This aspect ultimately clarifies why this material has remained inaccessible until now. However, this situation has changed due to the machine learning model we propose, as described below.\par

Building on the collection of automatically detected tables and using our model, we can now predict the similarity between every pair of tables, so that groups of similar tables (clusters) can be extracted, or alternatively, a list of most relevant tables can be retrieved from queries. However, for such machine learning approach to deliver accurate results (and to understand the reasons as to how we have developed the model), one needs to make sure that it applies reliably and systematically to the high heterogeneity of historical data, in particular the heterogeneity present in tabular data.\par

\subsection{Data Heterogeneity} \label{supplement:sec:heterogeneity}

The challenge of heterogeneous data emerges across many domains and is one of the key limiting factors to automate data analysis processes. It is characterized by a lack of uniform character and composition across samples in a dataset and makes up more than 90\% of big data \cite{Qiu2016ASO}. Typical examples are unstructured collections of texts and images, i.e.\ from different online sources, biological, geographical or medical sensor data as well as climate records. The field of  information fusion offers methods that combine data from different sources in order to improve information content via integration.\par

In real-world applications this poses a challenge, even in scenarios in which sensors are comparable in function and measurement quality as well as standardized data acquisition protocols are in place. For example, the heterogeneity of medical data is a key challenge to achieve robust models across hospitals and populations. Sources of heterogeneity can be divided into the following main categories: (i) technological heterogeneity due to different sensor manufacturer, recording protocols and data management, (ii) expert or institutional heterogeneity caused by individual experts inferring different information from comparable material and (iii) underlying differences in the observed population and their environmental conditions. Each of these categories adds to the complexity of data and makes it difficult for ML models to generalize to unseen data and infer robust predictions \cite{Adibuzzaman2017, Kelly2019}. Similar sources of heterogeneity are typical for historical corpora which have emerged throughout centuries and were only recently digitized. \par

\subsubsection{Heterogeneity in the \textit{Sacrobosco} Table corpus}\label{sec:sphaera_heterogeneity}
\begin{figure*}[h!]
    \centering
    \includegraphics[width=0.8\textwidth]{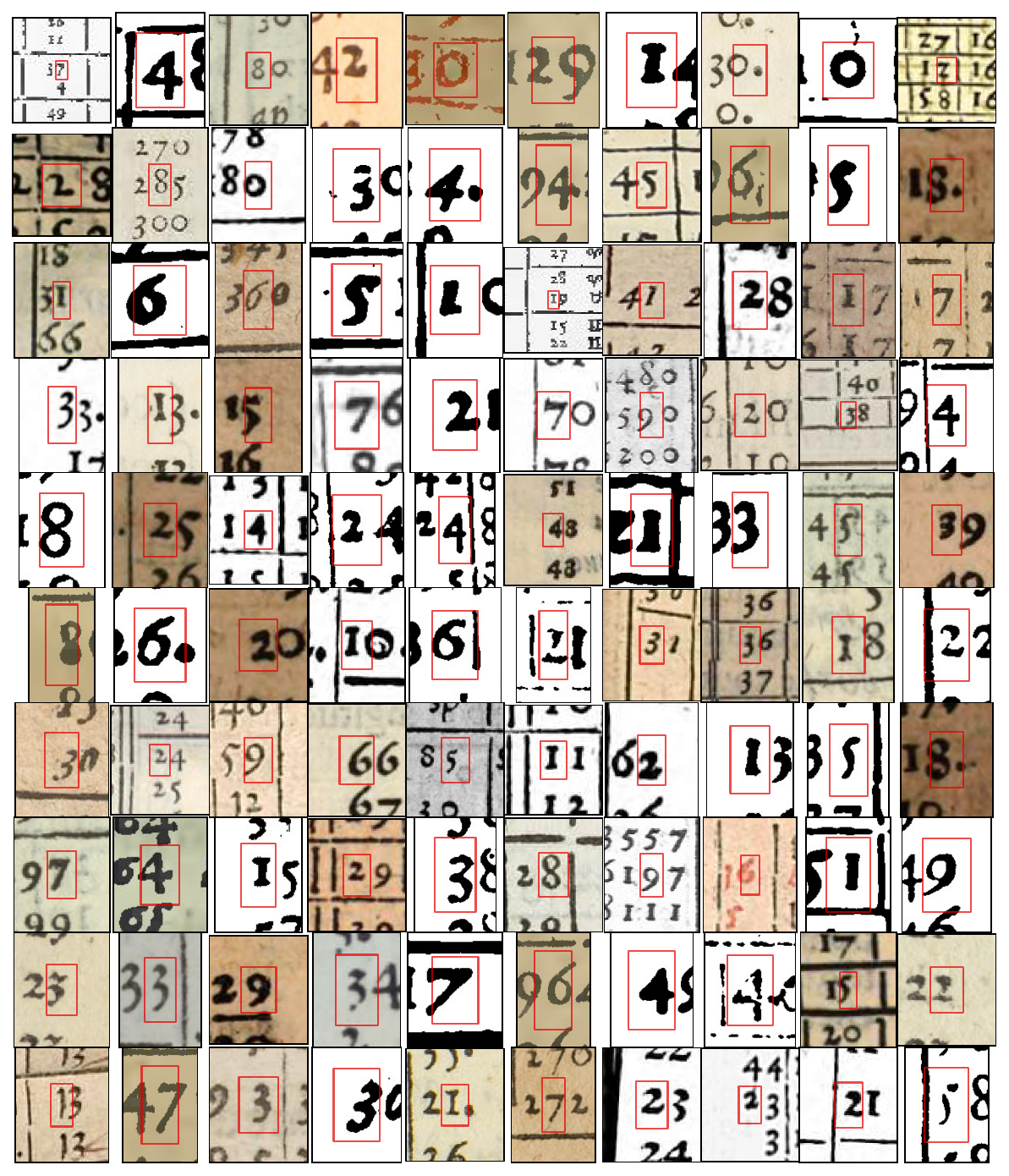}
    \caption{\textbf{Digit patches.} A hundred examples of the great heterogeneity in historical printing. The patches displayed are directly extracted from the scanned material before any pre-processing was applied. They are randomly selected digit patch examples used for the training of the digit recognition network.
    }
    \label{fig:example_patches}
\end{figure*}

\begin{figure*}[h!]
    \centering
    \includegraphics[width=0.8\textwidth]{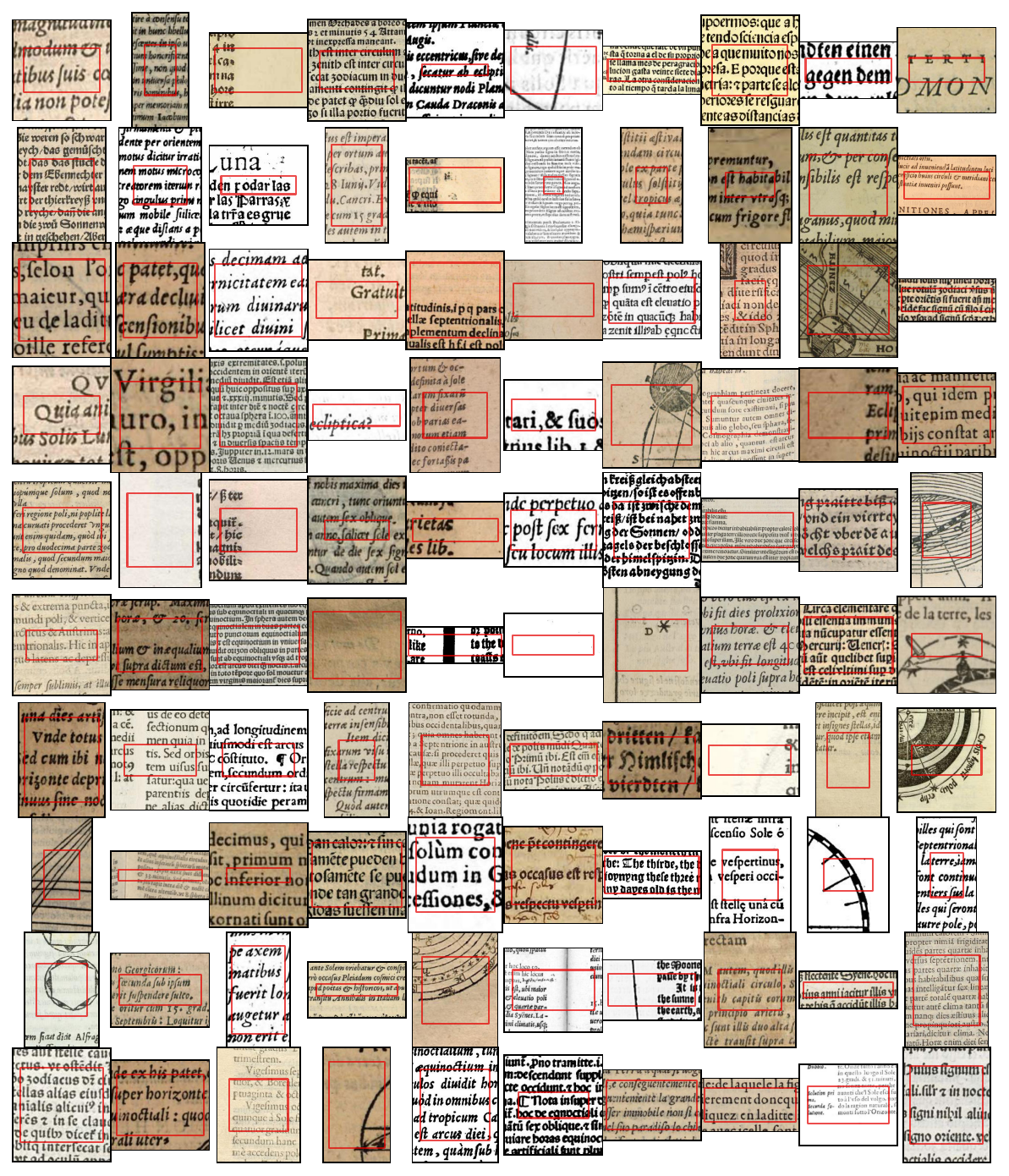}
    \caption{\textbf{Contrast patches.} Examples of non-table patches used as contrastive learning signal. Patches are extracted via randomly sampling regions from non-table book pages in the collection.
    }
    \label{fig:contrast_patches}
\end{figure*}

We illustrate some examples of heterogeneity in the \textit{Sacrobosco} Table corpus using digit and non-digit patches in Figure \ref{fig:example_patches} and Figure \ref{fig:contrast_patches} respectively. We further analyse the various reasons that result in the high heterogeneity of historical corpora and focus on the \textit{Sacrobosco} Table corpus specifically: \par

\begin{figure*}[t!]
    \centering
    \includegraphics[width=\textwidth]{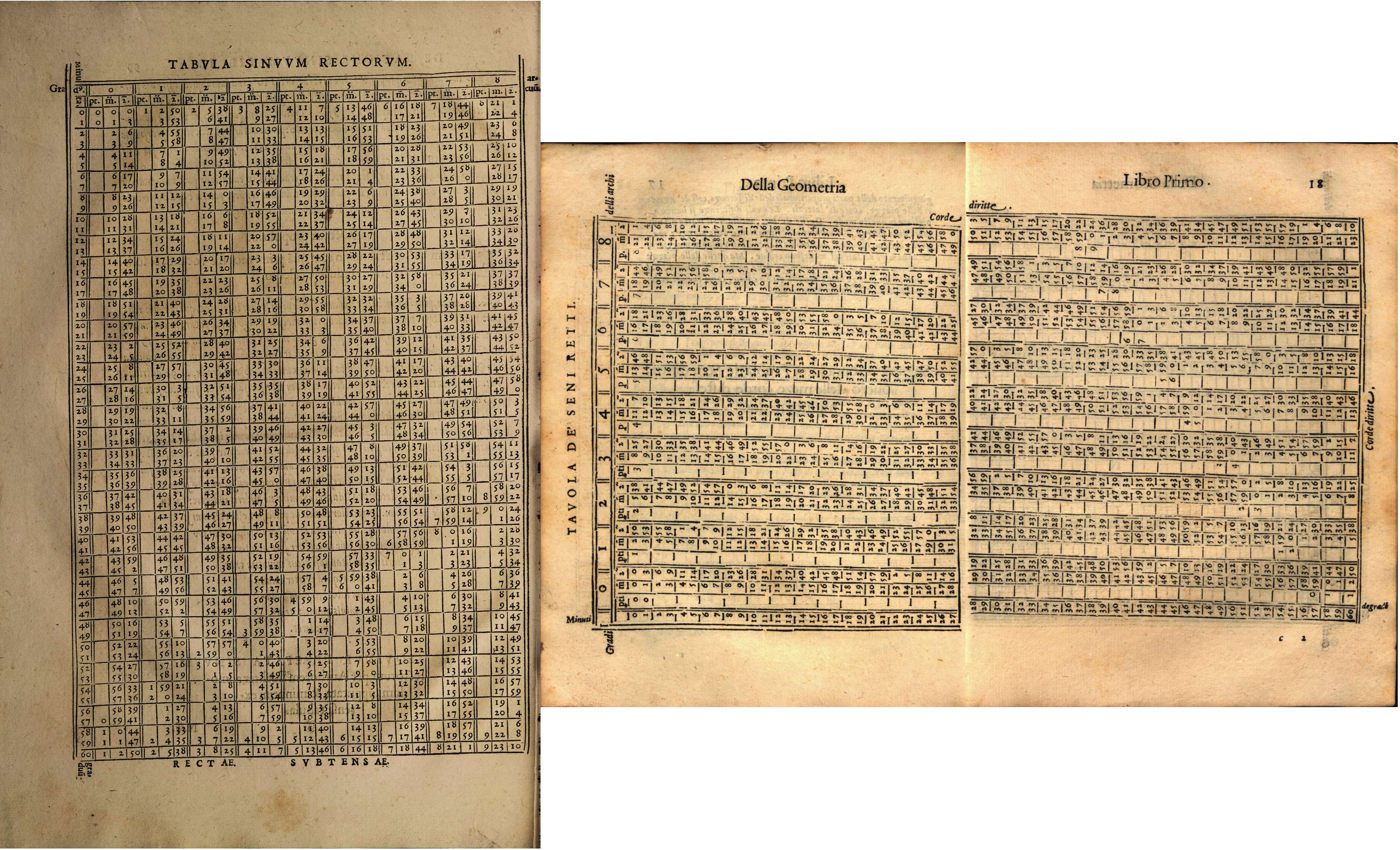}
    \caption{\textbf{Heterogeneity in layout of table content.} The same table of sinus values as published in two different works in 1542 and 1587. Typeface, layout, orientation and number of pages on which the table is set are different. Left: \cite[~99v]{Fine1542}, Right: \cite[~Libro primo della Geometria, 17v–18r]{Fine1587}. Courtesy of the Library of the Max Planck Institute for the History of Science.}
    \label{fig:sinus_tables}
\end{figure*}

\begin{figure*}[t!]
    \centering
    \includegraphics[width=\textwidth]{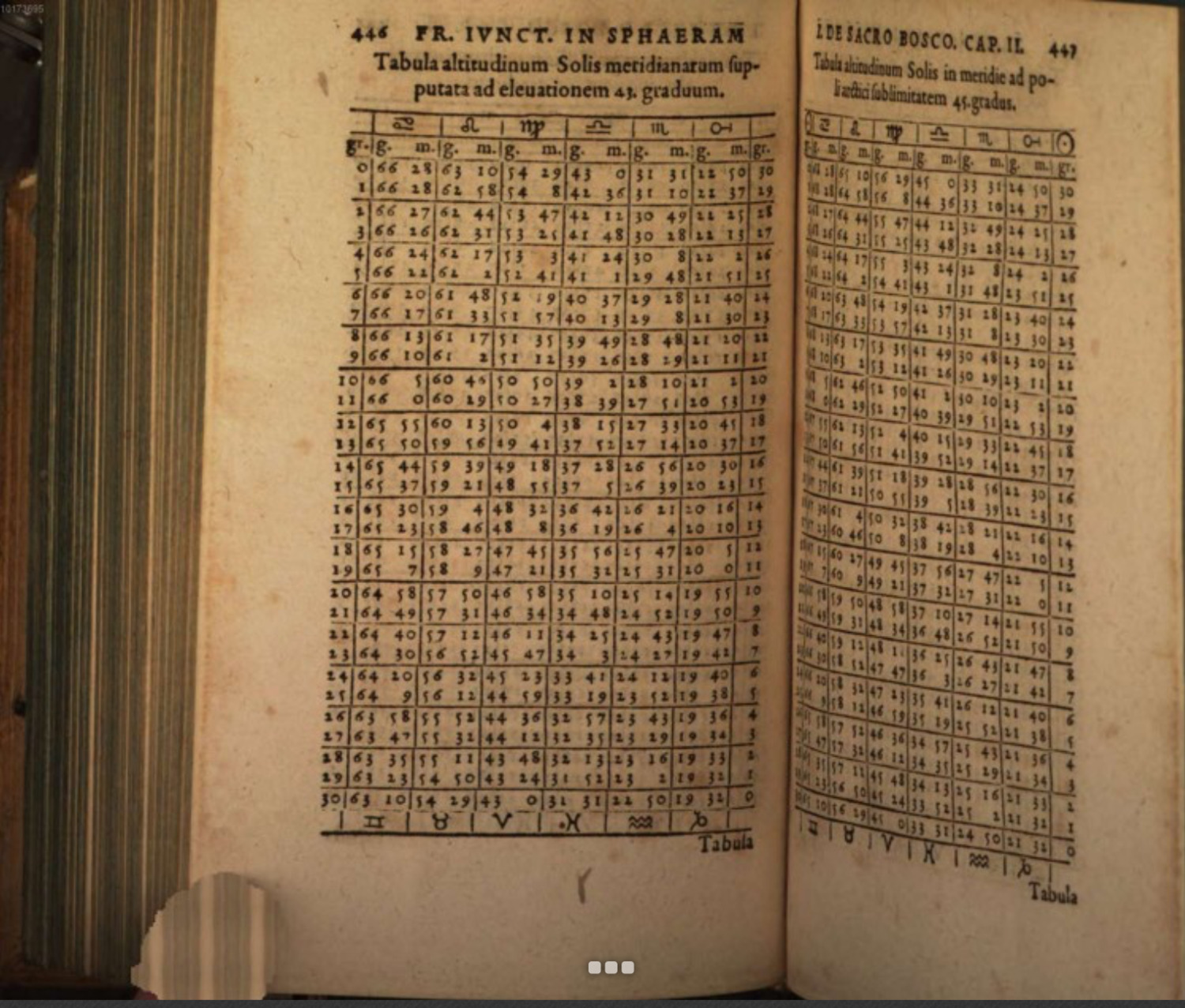}
    \caption{\textbf{Scanned page heterogeneity.} Due to the frequent impossibility to completely open rare ancient books in order to avoid damage of the binding, electronic reproductions include the page aside captured with a different angle to the lens of the camera. In the bottom left corner, we can additionally see an example of how devices used to fixate the page during scanning are digitally covered during post-processing. From \cite[446--447]{Giuntini1578}. München, Bayerische Staatsbibliothek, urn:nbn:de:bvb:12-bsb10173695-4.}
    \label{fig:Bad_Scan_01}
\end{figure*}

\begin{figure*}[t!]
    \centering
    \includegraphics[width=0.5\textwidth]{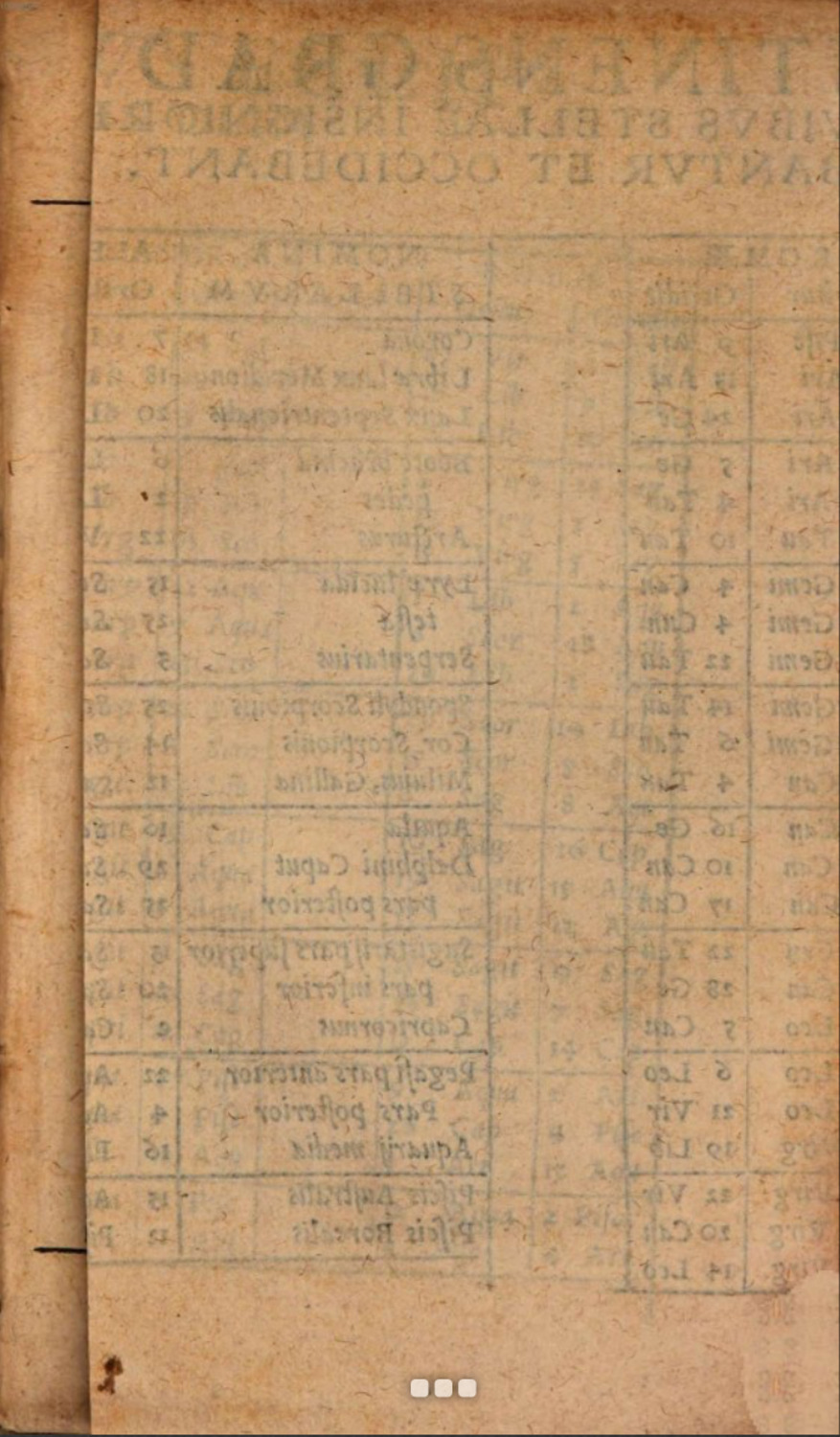}
    \caption[]{\textbf{Folded table page.} The hitherto impossibility for historians to access tables on a large scale brought, as a consequence, that numerical tables and even computational astronomic tables, have often been considered as historical sources of secondary relevance. This wrong assumption is confirmed by the practice of some archives and libraries not to unfold large-size numerical tables bound in the ancient books while scanning them. From  \cite[Unnumbered foldout]{SacroWitt1550}. München, Bayerische Staatsbibliothek, urn:nbn:de:bvb:12-bsb10998883-7.}
    \label{fig:Bad_Scan_02}
\end{figure*}

\begin{itemize}[label={}]
   \setlength\itemsep{0.8em}
    \item \textit{Technological heterogeneity} is a result of both the historical printing process which has caused irregularities during typesetting as well as the more recent and non-standardized digitization process across libraries and research projects. Typical cases include: (1) the uncontrolled digitization history by archives and libraries over the last decades which has resulted in electronic copies that are extremely heterogeneous with regard to resolution, colors, size, and both production and post-production procedures, also due to different hard- and software set-ups. In addition, (2) the fragility of the historical material may not permit a standard digitization set-up, which extends to the fact that the section of the scanned page can vary greatly, as in Fig. \ref{fig:Bad_Scan_01} and the page orientation is not standardized.
    
    \item \textit{Institutional heterogeneity} concerns the question of what \textit{similarity} between pages is based upon, i.e. (3) whether layout and decorative elements are considered when judging table similarity (stylistic overlap) or whether similarity is based purely on semantic overlap. 

    \item \textit{Population differences} reflect varying print traditions and printing quality as well as the preservation practice and status of the material. In the case of the \textit{Sacrobosco} Collection, the original material treatises are in (4) very different states of preservation which is a result of their individual histories in the last 500 years. Moreover, (5) tables that are printed in very different layouts, that is, the same table can ``look'' very different across books, as for example in Fig. \ref{fig:sinus_tables}; (6) depending on layout and format of the book, the same table can be found on one single page or stretched out over many successive pages in different books; (7) many of the tables are alpha-numerical, where the fractions of the `alpha' and the `numerical' components greatly vary; (8) each early modern printer had his/her own type-font and (9) numerical tables with many numbers were tedious to typeset resulting in a rather high level of noise of the actual with respect to the `correct' numbers. Finally, (10) pages can in part also be damaged, folded (Fig. \ref{fig:Bad_Scan_02}), wrinkled, stained or de-saturated.

\end{itemize}
This high heterogeneity is here further highlighted by the electronic copies of historical sources used in the entire Supplementary Material. We have consciously not post-processed these images but left in the exact same way they can be found in the repositories of libraries and archives. As mentioned, such heterogeneity precludes using standard ML solutions and we will next describe different directions to deal with heterogeneous material before introducing our \textit{atomization-recomposition} approach.\par

\subsubsection{Standard Approaches to Heterogeneous Data}\label{supplement:sec:standard_approaches_hetereogeneity}

Before model optimization, standardizing heterogeneous material through pre-processing is usually advantageous. This allows the ML model to focus on the extraction of task-related features rather than identifying and filtering various types of noise. This includes standard centering of data using corpus statistics, thresholding and binarization of inputs, or transformation of input features, e.g. using whitening to de-correlate the data. This can be a powerful step to alleviate heterogeneity that can be attributed to factors that are distinguishable from the relevant signal via a statistical analysis of the raw input data, e.g. variations in color distributions across images, sensor noise or varying signal strength.\par

Data heterogeneity that arises as a result of more complex variations usually has to be handled as part of an end-to-end training pipeline. This assumes that sufficient amounts of training data from sufficiently variable sources are available, and that these can be used to extract representations that are invariant towards various types of heterogeneity. Then, one can attempt to infer structured information by transfer-learning from pre-trained models requiring that data distributions lie on the same or very similar data-manifolds as the training set. Especially end-to-end deep-learning approaches have been a driver to bring annotations to unstructured data. Prominent examples are segmentation models \cite{journals/corr/RonnebergerFB15, matterport_maskrcnn_2017}  that are trained to extract object boundaries on images and have shown very promising transfer to domain-similar material. These can serve as the basis for subsequent object classification and knowledge discovery in heterogeneous material. Again, the main limiting factor is the availability of either ground truth bounding boxes or object masks which require human or even expert annotations. While community efforts have resulted in the availability of such data in some domains, a transfer to novel applications remains extremely challenging, i.e. microscopy data in the biomedical sciences or historical material in the digital humanities.\par

Rather than collecting additional annotated data from various domains, the field of domain generalization aims to enhance the model's ability to handle semantically similar data from out-of-training  distributions. This approach enables to bring structure to unseen domains and improves invariance and robustness properties across data from different sources \cite{10.5555/3016100.3016186, pmlr-v97-zhao19a, DBLP:journals/corr/abs-2103-03097,DBLP:journals/corr/abs-2106-04923}. Achieving this goal requires good knowledge of the data domain, as well as comprehensive labels that are sufficiently similar to enable successful generalization. In our case concerned with table similarities, however, there is no possibility to be provided with such labels in advance, which makes our development particularly innovative.\par

However, when dealing with historical material, we are limited to intermediate labels, e.g. character-level labels of digits. Nevertheless, we can leverage this data to build more complex features by employing our proposed \textit{atomization-recomposition} approach.\par

\subsection{Atomization-Recomposition Approach to Represent Historical Material} \label{sec:atomization_recomposition}

In order to deal with the different types of variability in the Sacrobosco Collection, we will next give a detailed description of our modeling steps. Our proposed approach involves an initial atomization step, which entails breaking down the intricate composition of numerical features into its basic components. In our setting, this refers to identifying single digits as the basic building block to compose more complex numerical strings. This approach offers the possibility to handle heterogeneity at a much lower data complexity, as previously suggested in the remote-sensing literature \cite{10.5555/3385337}. This further allows the use of  simpler and in total less annotations, while still being able to handle challenges related to robustness and invariance at a lower data complexity. In addition, this offers the possibility to build-in expert knowledge at the subsequent recomposition step.
The following sections will provide a more comprehensive description of how we have implemented the atomization-recomposition approach and conclude with a detailed demonstration on a pair of historical table pages. \par

\subsubsection{Pre-processing}
As a first step, we apply binarization to the full corpus. This involves normalizing each image using min-max normalization, applying a percentile filter at 0.8 and use the 10\% and 90\% quantiles of the pixel value distribution as the high and low cutoff values, which produces the binarized image. This process addresses  heterogeneity in color, different page background texture, as well as variations in contrast and brightness. We define a reference page height of 1200 pixels to which all pages are scaled in proportion to their original dimensions using bilinear interpolation. This allows to capture the statistics of the page features in sufficiently high resolution while still enabling a processing of full pages on standard GPU-hardware. We used Tesla P100 and V100 GPUs with 16GB/32GB storage.\par

\subsubsection{Atomization}
The backbone of our approach lies in a robust recognition of the basic atoms. In order to achieve this, our model has to be able to detect the correct digit with high accuracy, while avoiding to produce activity for non-digit context such as text, symbols or illustrations. To achieve this, we first introduce the recognition architecture which consists of two main encoder modules, namely, (i) the \textit{encoder} and (ii) the \textit{convolutional\_encoder} that together form our 7-layer neural network. The digit recognition model  was implemented in the \texttt{PyTorch 1.8.1} \cite{NEURIPS2019_9015} framework and its architecture is summarized in Figure \ref{code:digit_architecture}. The encoder consists of a 4-layer block of equivariant convolution layers as proposed in the framework of Equivariant Steerable Pyramids \cite{e2cnn}.
After all layers but the last, we use ReLU activation functions. 
The subsequent convolutional encoder processes extracted features of the first block further to build the digit detectors which output the single-digit activation  maps. This block consists of three standard convolutional layers of kernel sizes \{5$\times$5, 1$\times$1, 1$\times$1\}, strides of 1$\times$1 and padding of \{2$\times$2, 0$\times$0, 0$\times$0\}. \par

\begin{figure*}
\centering
\small
\begin{minipage}{0.8\textwidth}
\footnotesize\begin{spverbatim}
DigitModel(
  (encoder): Sequential(
    (0): R2Conv([8-Rotations], kernel_size=3, stride=1, padding=1, bias=False)
    (1): ReLU(inplace=True)
    (2): R2Conv([8-Rotations], kernel_size=3, stride=1, padding=1, bias=False)
    (3): ReLU(inplace=True])
    (4): R2Conv([8-Rotations], kernel_size=5, stride=1, padding=2, bias=False)
    (5): ReLU(inplace=True)
    (6): R2Conv([8-Rotations], kernel_size=5, stride=1, padding=2, bias=False)
    (7): GroupPooling([8-Rotations])
  )
  (convolutional_encoder): Sequential(
    (0): Conv2d(64, 64, kernel_size=(5, 5), stride=(1, 1), padding=(2, 2), bias=False)
    (1): ReLU(inplace=True)
    (2): Conv2d(64, 32, kernel_size=(1, 1), stride=(1, 1), bias=False)
    (3): ReLU(inplace=True)
    (4): Conv2d(32, 10, kernel_size=(1, 1), stride=(1, 1), bias=False)
  )
)
\end{spverbatim}
\caption{\textbf{Atom recognition architecture.} An initial encoder block extracts invariant feature representations that are then combined into single digit representations in a second convolutional encoder block.}
\label{code:digit_architecture}
\end{minipage}
\end{figure*}

\begin{figure*}[h!]
    \centering
    \includegraphics[width=1.\textwidth, trim={0 0 0 0},clip]{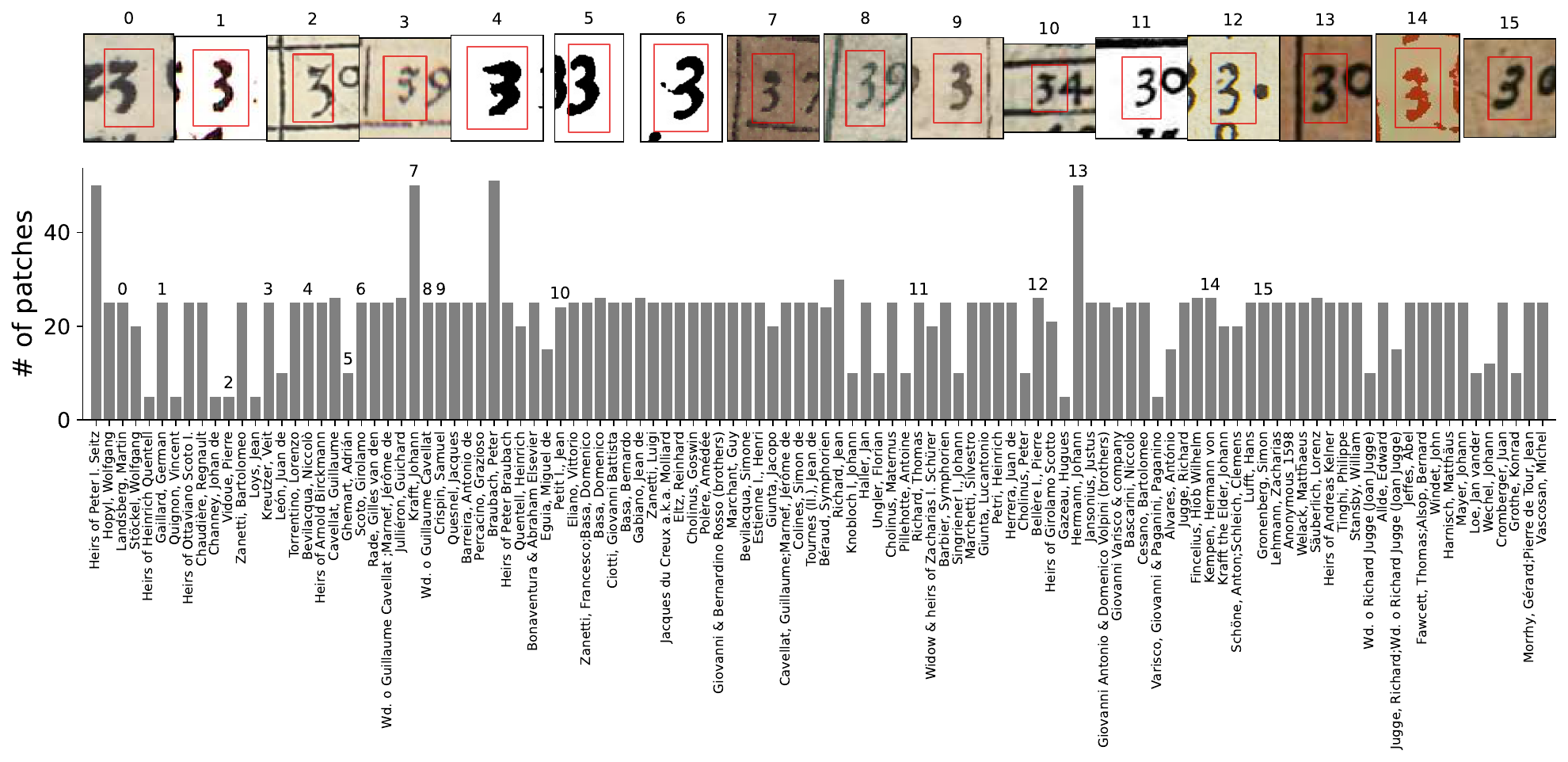}
    \caption{\textbf{Distribution of digit patches.} Histogram of number of annotated patches for each printer. For a randomly selected subset of printers, we show examples of the digit `3' that was produced by them.
    }
    \label{fig:printer_dist}
\end{figure*}

\subsubsubsection[]{Stylistic Invariance}
To capture the significant differences in historic fonts throughout the corpus, we have carefully designed the dataset to cover a representative set of fonts by sampling patches from different printers. The distribution of annotated digit patches over printers is shown in Figure \ref{fig:printer_dist}. This results in a total of 2,494 annotated full number patches from which 4,687 single digit patches are extracted.\par

\subsubsubsection[]{Local Scale and Rotation Invariance}
We further robustify the learned representations against style and scale heterogeneity by augmenting the training data patches using the following transformations: (i) We apply rotations of $\pm 10^{\circ}$, (ii) translations of the patch by (0.025$\times\text{img\_width/height}$ in x- and y-direction, (iii) proportional scaling of the full patch by a factor in the range ($0.8-1.2\times$) using bi-linear interpolation and (iv) shearing transformation of ($\pm 5^{\circ}$) along both spatial directions. For each possible augmentation, a random value from the specified range is sampled and added to the training dataset. In total, we sample as many augmented datapoints as there are annotated patches.\par

\subsubsubsection[]{Background Invariance Through Contrastive Learning}
At a semantic level, each page can consist of a combination of many distinct elements, including illustrations, text, mathematical equations, and tables. Each of them can be further broken down into sub-categories, e.g. illustrations can be geometric diagrams, star maps, depictions of scenes, etc. and similarly tables can contain mostly text, mostly numerical values or -- as is often the case -- a combination of both. This poses an additional challenge during processing since the recognition network has to be able to not only detect our desired features but in parallel has to learn to ignore the entire non-digit content. Considering for example that the letter 'O' is visually very similar to the digit '0', we aim to prevent page similarity to be based on such effects. To achieve this, we use all pages that do not contain any tabular structure in the Sacrobosco Collection and subsample pages from a diverse set of printers and books, similar to the selection of digit patches for annotation. A subset of these contrast patches is shown in Figure \ref{fig:contrast_patches} and illustrates the diverse elements that can occur in the collection.\par

\subsubsubsection*{Training} 
For model optimization, we use 80-20 train/test splits of the dataset and the digit model parameters are then trained using same amounts of single-digit and non-table patches. We find that including context improves digit recognition, and thus include a border of 10px  surrounding the digit bounding box. We minimize the mean squared error between true activation maps and model outputs using the loss term $\ell= \ell_{bbox} + 0.3 \cdot \ell_{context}$ with the Adam optimizer.

\begin{figure*}[h!]
    \centering
    \includegraphics[width=0.95\textwidth]{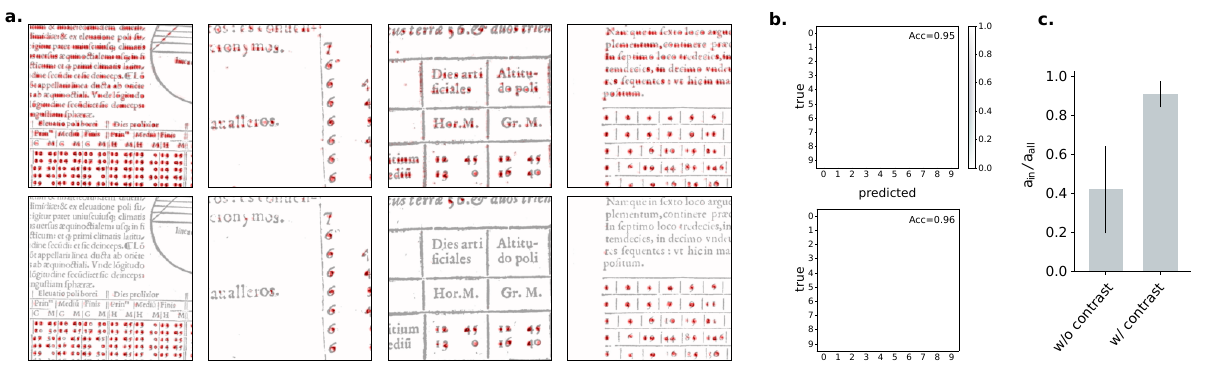}
    \caption{\textbf{Effect of contrast patches.} \textbf{a.} Pooled single-digit activation of the model either after training on single-digit patches (top row) or after adding additional contrastive non-digit patches during training (bottom row). \textbf{b.} Confusion matrices for the two different training scenarios.  \textbf{c.} Fraction of summed activation that falls inside annotated digit bounding boxes compared to the total page activation.
    }
    \label{fig:effect_contrast_patches}
\end{figure*}

The effect of training with or without contrast patches is further investigated in Fig. \ref{fig:effect_contrast_patches}. For a random subset of fully annotated pages we show patches as processed by the single digit model trained on digit patches only (top row) and a model trained using same amounts of digit and contrast patches (bottom row). We clearly observe that both approaches attribute activity successfully to the single-digits that occur in the various tables and both achieve comparable classification accuracies of 95-96\%. But, naive training using digit patches only produces considerable activity over text, letters and geometric elements as visible in \ref{fig:effect_contrast_patches}.a (top row). We use the fully annotated subset of Sacrobosco pages to compute the ratio of all activity that falls inside the digit bounding boxes as compared to all page activity (\ref{fig:effect_contrast_patches}.c) and find that without contrastive training  almost 60\% of the activation occurs on non-digit locations whereas we can reduce this number to 9\% when including contrast patches. \par

\subsubsubsection[]{Global Scale Invariance}
Global Scale differences in the collection can be caused by either (i) different sizes of the movable types used for  printing, i.e. larger or smaller typesetting, but also (ii) from the resolution differences that can result in several orders of pixel height and width spans in the data. In order to jointly model both of these sources, we chose to implement a multi-scale feature pyramid approach similar to the framework of steerable pyramids \cite{537667}. This has the advantage of parameter-efficacy since no additional trainable parameters are introduced and of model transparency since the multi-scale approach is based on the linear decomposition of the image at different scales from which the most activating feature scale is chosen and thus, remains fully explainable. \\
For this, we re-scale the image to a reference  height or width of $1200$px at reference scale $s=1.0$ (depending on portrait or landscape orientation) using bilinear interpolation. 
Resulting input images are collected for every scale $s \in S =\{s_1,...,1.0, ...,s_K\}$ and fed through the atom-recognition network. The scale  $s^{*} = \max_{s\in S } \sum_j {\ba_j(\bx; s)}$, which maximizes the spatially pooled activity over all features $j$  is then chosen for further processing.\par

\subsubsubsection[]{Global Rotation Invariance}
Likewise, variations in page orientation may arise due to either
(i) the printing process, where a table or illustration was considered more legible in a landscape layout, or (ii) the more recent digitization process itself. We model both of these as in the previous section concerning scale by including page input rotations $\theta \in \Theta =\{-90, 0, 90\}^{\circ}$ and select the rotation that maximizes activity:
$\theta^{*} = \max_{\theta\in \Theta } \sum_j {\ba_j(\bx; \theta)}$.\par

With the different sources of heterogeneity addressed, we are now able to robustly extract single-digit activation maps. These representations will serve as the necessary building blocks to recompose more complex and task-relevant features, i.e. bigrams.

\subsubsection{Recomposition} \label{supplement:recomposition}

To efficiently achieve the recomposition of single-digit activation to bigram maps, we apply a hard-coded structure on top of the learned recognition model to compute  bigram maps via an element-wise `min' operation:

\begin{align*}
\ba_{jk}^{(\tau)}(\x; s, \theta) &= \min\big\{\ba_j(\bx; s, \theta) ,\tau(\ba_k(\bx; s, \theta))\big\},
\end{align*}
which signals the presence of bigrams $jk \in 00$--$99$ at scale $s$ and rotation $\theta$, and can be seen as a continuous `\textsc{and}' \cite{Kauffmann20} operation. In addition, we build features that detect isolated single digits  $j \in \{\square0\square,\dots,\square9\square\}$ with ``$\square$'' indicating that no digit activity is present in the neighborhood. For this, the single digit activation maps and two binarized neighborhood maps with shifts $\pm \delta$ that signal absence of a digit feature are computed, and another `min' operation over all three maps outputs the final digit map. \par

The function $\tau$ represents a translation operation shifting activation maps by $\delta$. We use multiple shifts as candidate alignments and identify digit compositions by applying a spatial max-pooling layer:
\begin{align*}
\ba_{jk}(\x) &= \max_\tau \big\{ \ba_{jk}^{(\tau)}(\x; s, \theta) \big\}.
\end{align*}
The `max' operation can be interpreted as a continuous `\textsc{or}', and determines at each location whether a bigram has been found for at least one of the candidate alignments. This results in total number of 110 feature maps.
In our experiments, we use $s \in \{0.5, 0.65, 0.8, 0.95,  1.0\}$, $\theta \in \{-90,0,90\}^{\circ} $ and $\delta \in \{8,10\}$ pixels.\par

\subsubsubsection[]{Activity Peak Detection}
Having solved the challenge of identifying task-relevant features, we next would like to arrive at a summary representation of page content. To accomplish this, we can directly perform spatial pooling of activity over feature maps $\ba_{jk}$. While this is a simple and viable approach that does produce meaningful similarity as we will see in Section \ref{supplement:subsubsec:fully_annoated}, it may not be clear how the pooled activity corresponds to feature presence on a page: a pooled activity of 100 can correspond to two very prototypical bigrams that activate the network very strongly or four weakly activated less prototypical examples. Besides thresholding before pooling, we propose to use peak detection to convert the raw activation maps into bigram count maps. We start from a set of 100 bigram maps $\ba_{jk}$ with $jk=\{00,\ldots,99\}$ which are added to 10 maps for isolated digits $\hat{\ba}_{i}$ with $i=\{\_0\_,\ldots,\_9\_\}$ resulting in $\bar{\mathbf{a}} = (\ba_{i}, \ba_{jk} )$. Since, the max-pooling used for the bigrams reduces the activity levels in comparison to the isolated digit maps, we introduce a scaling parameter $\alpha$ to the latter $\ba_{i} = \hat{\ba}_{i}/\alpha$. Next, we subtract a bias term $\beta \cdot \max_{(x,y)} \bar{\mathbf{a}}_{(x,y)}$ computed as the product of relative scaling parameter $\beta$ and the maximum pixel value in all maps. Resulting maps are rectified, which, similarly to the processing applied to the single digit activation maps, reduces weak background activity. Then, for each of the 110 feature maps, we extract the feature regions that occur at all non-zero locations and compute all peaks using the center of activity mass. We determine the linkage matrix using the distances between centers and perform a hierarchical clustering to group close-by activated pixels into groups of pixels that belong to a single bigram. To limit the size of clustered regions, we define a maximum distance parameter $d$. We select optimal parameters using histogram Pearson correlation scores on the training patches and set $\alpha=3$, $\beta=0.12$ and $d=15$. Using the center of mass as the digit location and its extracted feature label, we now can inspect a human-readable digit decoding as presented in Figure \ref{fig:page_ex1_ex2} (lower left overlay) that can serve as a useful verification and insight step during the historical analysis.\par

\subsubsection{Demonstration of the Recomposition Steps on a Pair of Tables}
\begin{figure*}[h!]
    \begin{minipage}{0.46\textwidth}
    \centering
    \includegraphics[width=\textwidth]{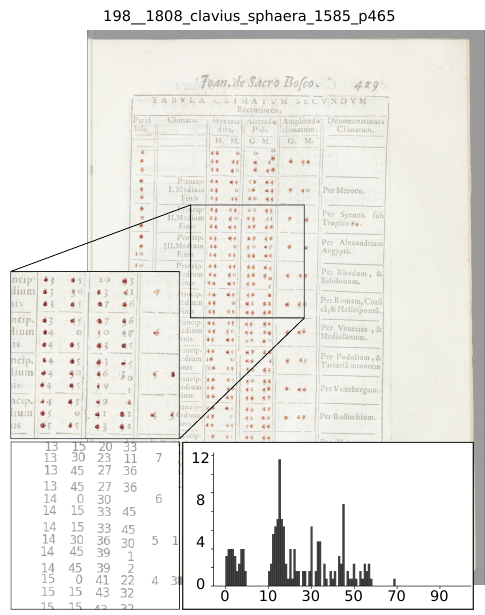}
    \end{minipage}\hfill
    \begin{minipage}{0.46\textwidth}
    \centering
    \includegraphics[width=\textwidth]{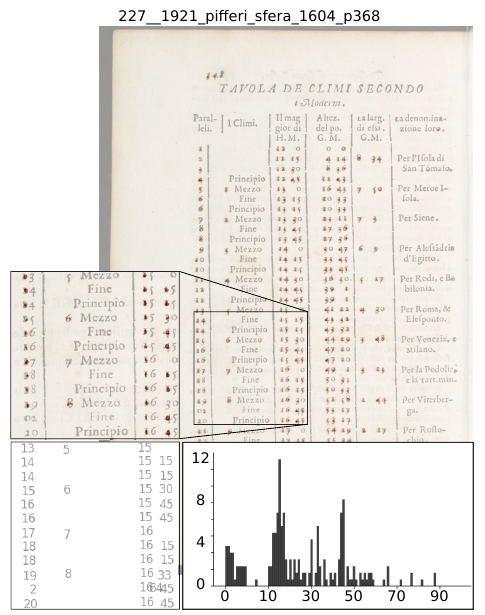}
    \end{minipage}
    \caption{\textbf{Visualization of different processing steps of our approach for two pages of same content from the Sacrobosco Tables dataset.} The background image shows the single digit activation maps pooled over digits 0--9. The zoomed in overlay contains the resulting bigram activations, below the extracted digits and the resulting histogram representation.}
    \label{fig:page_ex1_ex2}
\end{figure*}

In Figure \ref{fig:page_ex1_ex2}, we present two exemplary pages at different processing steps of the recomposition stage of our approach. The original page is displayed in full in the background and overlaid with the single digit activity. The inserts show bigram activity (top left) and the extracted digits (bottom left) after peak-detection was applied. Finally, we show the full histogram (bottom right).\par

\begin{figure*}[h!]
    \centering
    \includegraphics[width=0.9\textwidth]{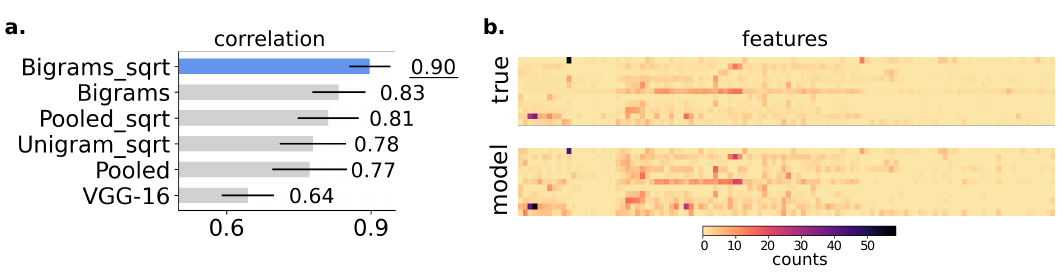}
    \caption{\textbf{Correlation on fully annotated pages.} \textbf{a.} Pearson correlation scores for different table representations. \textbf{b.} Ground truth and best model-based table histograms. In total there are 2261 bigram features in the source material.
    }
    \label{fig:annotated_page_corrs}
\end{figure*}

\subsubsubsection[]{Comparing Activity Pooling with Peak Detection}\label{supplement:subsubsec:fully_annoated}
We next want to quantify if, in addition to the above described advantages, the peak detection is also useful to provide a more accurate histogram representation. We have experimented with a non-linear mapping, i.e. the square root, of the histogram counts to take the scale differences between very frequent and rare number features into consideration. This allows to balance the vanishing contribution that less frequent occurrences have in presence of very frequent bigrams when computing distances or correlation scores. We use the fully annotated table pages and extract all occurrences of bigrams and isolated single-digits to compute ground-truth histograms for each page. The following approaches are used for comparison: (i) pooled bigram activity (Pooled),  (ii) square root transformed pooled bigram (Pooled\_sqrt), (iii) counts from the peak detection processing (Bigrams), (iv) square root transformed peak detection histograms (Bigrams\_sqrt), (v) square root transformed pooled unigram activity (Unigram\_sqrt) and (vi) spatially-pooled VGG-16 output feature maps after the last of five convolutional blocks (VGG-16). In Figure \ref{fig:annotated_page_corrs}, we see that peak detection based representations (Bigrams, Bigrams\_sqrt) indeed increase Pearson correlation scores over the pooled activations. In addition, applying the square root transformation further improves correlation in both the pooled and peak detection scenarios. This can be explained by the increased sensitivity towards less frequent bigram counts.\par 

\subsubsubsection[]{Evaluating cluster classification performance}\label{supplement:cluster_performance}
In addition to the validation of the accurate detection of digit distributions in the previous section, we additionally argue that the ML approach has to be evaluated on the task we are finally interested in. In our case this refers to the detection of groups of semantically similar tables. For this, we have used a subset of the Sacrobosco corpus that contains one and two-page instances of the sun-zodiac tables that are described in more detail in Section \ref{supplement:data} and have been annotated by a domain expert. The resulting 71 table pages contain more than 45,000 single digits, which we split into train-test (50/50) sets and a nearest-neighbor distance model was fitted on the training set. For all test data points, a class label according to the closest distance is assigned by the model and, finally, test set cluster purity is computed for ten random seeds. \par

\begin{figure*}[h!]
    \centering
    \includegraphics[width=0.4\textwidth]{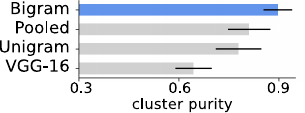}
    \caption{\textbf{Cluster purity of sun-zodiac tables.}
    }
    \label{fig:cluster_performance}
\end{figure*}

We have compared the following approaches to compute table page representations: (i) Bigrams\_sqrt: Bigram histogram counts were obtained using the bigram model with peak detection and square root mapping.  (ii) Pooled: Activity maps were obtained as in (i), but instead of peak detection, we directly applied spatial sum-pooling to the bigram maps.  (iii) Unigram: Instead of computing bigram maps, we built a ten-dimensional unigram count histogram using peak detection. (iv) VGG-16: We used the pretrained encoder of the deep image classification network VGG-16 \cite{Simonyan15} and extracted spatially-pooled output feature maps after the last of five convolutional blocks. Results in Figure \ref{fig:cluster_performance} show that the bigram-based approach outperforms simpler unigram-based or activity-based representations as well as VGG-16 representations with cluster purity at around 90\%.

The main steps of the atomization-recomposition approach can be summarized as follows:
\begin{minipage}[h!]{\textwidth}
\begin{itemize}[leftmargin=0.5cm]
    \item[] Atomization
    \begin{itemize}[leftmargin=0.5cm]
        \item[]  1.\quad Determine the basic building blocks (atoms) in the input data.
        \item[]  2.\quad Collect or extract atom  annotations.
        \item[]  3.\quad Train and Validate the atom recognition model.
    \end{itemize}
    \item[] Recomposition
        \begin{itemize}[leftmargin=0.5cm]
        \item[]  4.\quad Recompose the atoms to build task-relevant features.
        \item[]  5.\quad Verify the features using explainable AI.
        \item[]  6.\quad Evaluate the model on the final task of interest for which annotations are attainable.
    \end{itemize}
\end{itemize}
\end{minipage}

\subsection{Limitations of classical OCR approaches for digit recognition} \label{supplement:sec:limit_OCR}

While traditional OCR approaches rely on simple computer vision algorithms to segment and extract characters from pages \cite{Zheng2004}, the most recent approaches use more complex networks that segment the page and extract text regions, and a combination of convolutional and recurrent neural networks for character recognition and transcription \cite{Martinek2020, Lijun2021, Diaz2021}. However, historical corpora, including the Sacrobosco collection, present a major challenge for many OCR approaches due to their high degree of heterogeneity, characterized by diverse languages and fonts, complex page designs, as well as the myriad of issues that arise from bad scans, faded text, bleed-through, smears, and damage incurred over time \cite{Alaa2019} (cf. Section \ref{sec:sphaera_heterogeneity}). 
While impressive progress has recently been achieved to bring standard OCR approaches to historical data, the representation of digits and specifically tables has not been addressed so far.\par

\subsubsection{Evaluation on fully annotated pages} \label{supplement:sec:fully_annotated}
In order to investigate the effectiveness of our approach with respect to traditional OCR methods, we compare our results with the output obtained from Latin OCR, a model build by `The Duke Collaboratory for Classics Computing' and trained on a large collection of Latin texts covering almost two millennia \cite{Bamman2012}. Similar to our analysis in Section \ref{supplement:subsubsec:fully_annoated}, we use the fully annotated subset of the tables from the Sacrobosco collection and compute Pearson correlation coefficients between ground truth histograms and the extracted model histograms. On average we observe that our Peak Detection (PD) approach results in higher correlation scores that also vary less across different pages than the Latin OCR which shows the effectiveness of our approach to detect digits (Table \ref{table:Pearson_1}). In order to better understand the results, we report the pearson correlation on three different groups of pages (see Table \ref{table:Pearson_2}), low number density pages ($\leq$ 150 bigrams/page), dense pages (150 - 300 bigrams/page), and very dense pages (>300 bigrams/page). The Pearson correlation scores in Table \ref{table:Pearson_2} clearly show that while our approach outperforms OCR in all of these classes, the margin grows with the numerical density on a page.\par

\begin{table}[h!]
    \centering
    \begin{tabular}{lcccc}
    
       \toprule
                  & mean  & median & std \\
       \midrule

        Latin OCR & 0.747 & 0.849 & 0.272\\
        PD        & 0.871 & 0.938 & 0.166\\
        \bottomrule
    \end{tabular}
    \smallskip
    \caption{\textbf{Pearson correlation between ground truth annotations and our peak detection approach as compared to a state-of-the-art OCR system.}}
    \label{table:Pearson_1}
\end{table}

\begin{table}[h!]
\centering
\begin{tabular}{lllll}
\toprule
        density    & $\rho_{\text{OCR}}$  &  $\rho_{\text{PD}}$ & $N_\text{bigr.}$ & $N_\text{uni.}$ \\
\midrule
       low ($\leq$ 150)   &   0.76 &  0.84   &    493 &        916 \\
     dense (150-300)      &   0.86 &  0.88   &    786 &       1501 \\
very dense (>300)         &   0.49 &  0.93   &    982 &       1764 \\
\bottomrule
\end{tabular}
    \smallskip
    \caption{\textbf{Pearson correlation at different digit density levels for a state-of-the-art OCR system and our peak detection approach. }}
    \label{table:Pearson_2}
\end{table}

\subsection{Model Validation using Explainable AI} \label{supplement:sec:xai}
Making modern and typically complex machine-learning models more robust towards data distribution shifts and adversarial attacks is crucial for their application in science, society and industry. The traditional ML evaluation pipeline aims at validating the nominal accuracy of the model, but unfortunately, highly accurate ML models can ground their predictions in unexpected ways, i.e. via  overconfidence in certain data features, reliance on spurious correlations or classification sensitivity to noise. 
Thus, it is crucial to further validate the learned representations as well as the model's inner workings using additional techniques such as visualization and explainable AI \cite{lapuschkin-ncomm19}.\par

Visualization and projection techniques including clustering are useful to analyse full datasets by representing them in a lower-dimensional space that can be directly interpreted by humans. In the presence of labels, they can be used to measure how well a learned representation is able to separate datapoints from different classes, i.e. using  cluster purity, normalized mutual information or the Rand index. In the absence of any label information, formed clusters can be evaluated using distance scores as in the Silhouette Coefficient or Dunn's index. These unsupervised measures do not necessarily reflect user expectations since data points can be clustered perfectly but built on unexpected or unwanted data features. Thus, a manual validation of the projection or a subset thereof is crucial to move towards a conclusive evaluation.\par

In order to evaluate the ML model itself and the features that are used for a certain prediction, the field of explainable AI  \cite{DBLP:journals/dsp/MontavonSM18,DBLP:series/lncs/11700, PMID:33079674, samek2021explaining} has developed techniques to make models transparent and reveal their inner logic. This transparency enables the development of more trustworthy systems which are of crucial importance when we are interested in generating novel domain insights.
Historians for example need to be able to clearly understand which features in a document or collection thereof lead to a certain model prediction in order to arrive at well-grounded historical inferences.\par

A broad range of methods have been proposed for Explainable AI, and we briefly present here the `Layer-wise Relevance Propagation' (LRP) method \cite{bach2015pixel}, which applies to a broad range of complex classifiers, has advantageous computational and robustness properties, and an extension of which, called `BiLRP' has been developed to provide explanations for similarity models.

The LRP method considers a neural network composed of multiple layers, with input $\bx \in \mathbb{R}^d$ and output $f(\bx) \in \mathbb{R}$, e.g.\ the activation for a given class in the last layer. LRP seeks to attribute the prediction score to the input layer, specifically, producing scores $R_i$ for each input feature $x_i$ with $i=1\dots d$. To achieve this, LRP operates layer-wise, starting in the top layer and then redistributing the function output to the neurons one layer below. This redistribution proceeds layer after layer by means of propagation rules, until the input layer is reached, at which point the explanation can be collected.

For illustration, let $j$ and $k$ be indices of neurons in two consecutive layers, $a_j, a_k$ be the associated activations, and $w_{jk}$ the weight connecting the two neurons. In the forward pass, activations between these two layers are typically related via the equation $a_k = \max(0,\sum_{0,j} a_j w_{jk})$. For such layers, LRP redistributes using propagation rules of the type:
$$
R_j = \sum_k \frac{a_j (w_{jk} + \gamma w_{jk}^+)}{\sum_{j} a_j (w_{jk} + \gamma w_{jk}^+)} R_k,
$$
i.e.\ neurons that are active and to which the model responds strongest receive more relevance than their counterparts. The parameter $\gamma$ can be interpreted as a robustness parameter that needs to be tuned for explanation quality. When setting the parameter $\gamma$ to $0$, the procedure can be shown to reduce to simple methods such as Gradient$\,\times\,$Input. Other redistribution rules can be used for different layers. We refer to \cite{lrpoverview} for further examples of propagation rules.

In order to bring verifiability to our approach, in particular, our similarity model of table pages is of the type $y = \langle \phi(\bx),\phi(\bx') \rangle$
where the $\bx,\bx' \in \mathbb{R}^d$ are two input examples, where $\phi:\mathbb{R}^d \to \mathbb{R}^h$ is a feature map (typically a neural network) and where $y \in \mathbb{R}$ is the predicted similarity score. For such models, the LRP approach is not directly applicable and one needs to consider its extension BiLRP \cite{eberle2020}. BiLRP recognizes that models with dot product outputs are intrinsically locally bilinear (instead of locally linear as for LRP) and thus better explained in terms of \textit{joint} feature contributions.

BiLRP proceeds in a similar way as LRP, redistributing the relevance scores from layer to layer but this time using the propagation rule:
\begin{align}
R_{jj'}
&= \sum_{kk'}\frac{a_j a_{j'} (w_{jk}+\gamma w_{jk}^+) (w_{j'k'}+\gamma w_{j'k'}^+)}{\sum_{jj'} a_j a_{j'} (w_{jk}+\gamma w_{jk}^+) (w_{j'k'}+\gamma w_{j'k'}^+)}
 R_{kk'},
 \label{eq:lrpgamma2}
\end{align}
which bears resemblance to the standard LRP rule but many terms that are doubled. In this rule, $j$ and $k$ are neurons in two consecutive layers of the branch processing image $\bx$, and where $j'$ and $k'$ are neurons in two consecutive layers of the branch processing image $\bx'$. In other words, pairs of activations can only be relevant if they jointly activate and if the model responds to both of them. Like for the standard LRP, the parameter $\gamma$ controls robustness of the explanation. If $\gamma$ is set to zero, the explanation reduces to that of a simple second-order explanation called Hessian$\,\times\,$Product \cite{eberle2020}. In practice, due to the quadratic growth of elements of the sum, the BiLRP procedure can be applied more efficiently by computing standard LRP passes for each of the individual elements of the dot product, and recombining the produced explanations using a matrix product. Resulting scores are then only combined at the input into the full relevance matrix.

\begin{figure*}[t]
    \centering
    \includegraphics[width=0.8\textwidth]{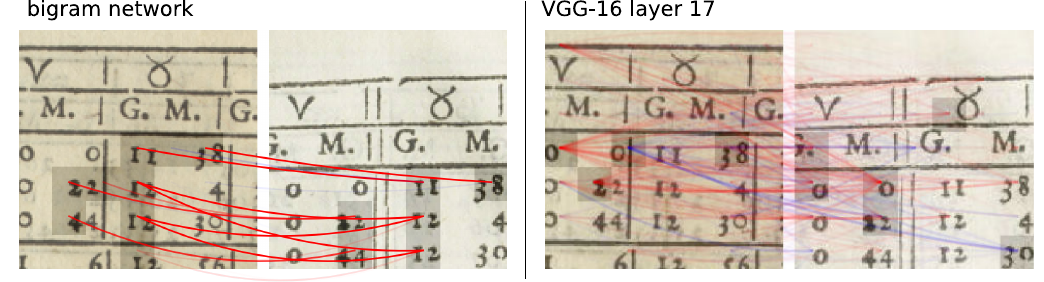}
    \caption{\textbf{Explaining similarity.} Left: Detailed BiLRP explanations highlighting the relevant feature interactions of predicted similarities between the two input tables for our bigram approach in red. Right: Resulting BiLRP explanations for the pretrained object recognition model VGG-16. Negatively relevant interactions are shown in blue.
    }
    \label{fig:bilrp_xai}
\end{figure*}

The information in this matrix can be visualized by plotting the scores as connections between pixel locations $i$ and $i'$. It can be beneficial to reduce pixel-level granularity of the explanation by grouping pixels into patches $(\mathcal{I}_1, \mathcal{I}_2, ...)$ and $(\mathcal{I'}_1, \mathcal{I'}_2,...) $. We compare the explanations computed by our bigram network to a standard VGG-16 represenation as shown in Figure \ref{fig:bilrp_xai}. Explanations for the high similarity in the bigram network are indeed based on numerical content shared among the two images. Since we explain the dot product of histograms computed by spatial pooling over the page, we observe that feature interactions of the same digit can appear at different locations as visible for the bigram `12'. While similarity between the VGG-16 embeddings is of comparable strength to the similarity score of the bigram representation, we find that it is predominantly based on task-irrelevant interactions like table borders and generally geometric shapes that interact across bigrams.  In comparison to the bigram network we observe overall that relevant interactions are less pronounced which indicates the lack of a meaningful similarity structure that matches related items and that some negatively relevant interactions contradict the similarity score. This highlights that model robustness and conformity with user expectations are not necessarily reflected by high model prediction scores and that in order to produce reliable insights from ML models we need to verify their inner workings. We conclude that without having to collect ground truth expert-annotations of the table similarity we are able to verify the proposed bigram approach from a single pair of tables.\par

\subsection{Generating Historical Insights}

The increasing use of machine learning across various scientific fields has not only enabled the large-scale analysis and automatic organization of big data but also started to be a valuable tool for the generation of novel domain insights, i.e. in quantum chemistry \cite{Rupp2012, SchNet, Schuett2017, doi:10.1021/acs.chemrev.0c01111}, the climate and earth sciences \cite{Sumbul2019BigearthnetAL,doi:10.1126/sciadv.aau4996,https://doi.org/10.1029/2019MS002002}, astronomy \cite{Shallue2017IdentifyingEW, Valizadegan2021Exominer}, biomedicine \cite{Klauschen2018,breastcancer2021} or neuroscience \cite{guclu2015, Cadena2019, Neumann2019, MATHIS20201}. This has been especially fruitful in domains in which computer-aided experimentation and mathematical tools are already integral part to the research process. The automatic storage and processing  of experimental data hereby serves as valuable training data for ML models. 

ML-assisted insight discovery has primarily found its application in the natural sciences \cite{Roscher_2020_insights}, but other fields have also begun to explore the potential of ML techniques to push existing boundaries of their respective domains. Examples include natural language processing for under-represented languages such as Sub-Saharan tongues or low-resource problems in the digital humanities and historical sciences \cite{10.1145/3567592, pine-etal-2022-requirements}.\par

Machine learning in the humanities has been used for broad sets of tasks mirroring the diverse  disciplines ranging from archaeology, history, literary studies, linguistics to philosophy. Most widely explored applications can be divided into the analysis of networks, images and texts.
Network studies construct a graph connecting items according to available metadata with the goal to explore and visualize large data, identify relational patterns or execute an analysis of the community structure \cite{mva19,SRN2020}.
The analysis of image material takes advantage of advances in computer vision and has been widely explored, for example, for the automated analysis of image style \cite{7780634, 10.1186/s13673-016-0063-4}, the extraction of similarity structure \cite{10.1007/978-3-319-46604-0_52, Lang_art_similartiy_2018}, for the image-based classification of visual material \cite{RN2922,4586391, vane2016, schlag_coins_2017, shen2019discovery, monnier_doc_extractor_2020}, and for image extraction from historical documents \cite{Buttner2022, monnier_doc_extractor_2020}.
Textual material has been analysed in the context of topic modeling \cite{Tangherlini2013, jockers2013significant, schoch2021topic},  ML-assisted annotation and text completion \cite{koppel-etal-2016-reconstructing, AssaelEtAlNature2022} as well as modeling ancient languages \cite{YadavNisha2010Saot,luo-etal-2019-neural, 10.1162/tacl_a_00354}.
In addition, hybrid approaches, e.g., for the task of reconstructing ancient text from images  \cite{asssome2022restoring}, have been
explored, too.\par

\begin{figure*}[t]
    \centering
    \includegraphics[width=0.9\textwidth]{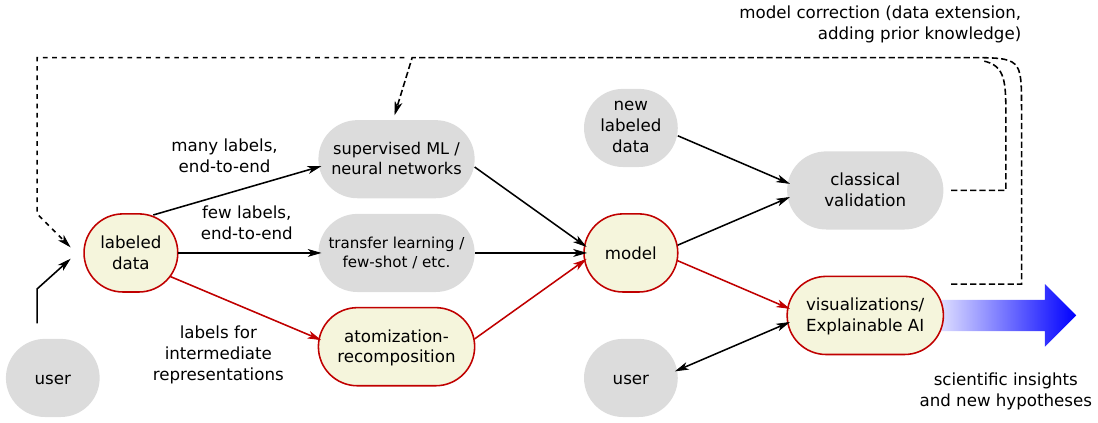}
    \caption{\textbf{Insights using machine learning.} Our proposed atomization-recomposition approach is embedded into the general machine learning and validation pipeline. The extraction of scientific insights relies on the analysis of interpretable model predictions in form of visualizations and explanations by the user.}
    \label{fig:flowchart}
\end{figure*}

In Figure \ref{fig:flowchart}, we summarize how ML can be integrated in the process of extracting scientific insights from data. If sufficient training data and appropriate, i.e. task-relevant, labels are available or can be easily collected, a model can be trained end-to-end. If only few labels are available, an and end-to-end approach can still be feasible, exploiting for instance methods of transfer or
few-shot learning.  Alternatively, we propose an atomization-recompostion approach in which a complex task is broken into easier steps, namely, the annotation of atoms for which a less complex model can be trained. These atoms then serve as a way to compose task-relevant features. As a fundamental next step, either model has to be verified and validated in order to ensure that the results are grounded in an accurate and consistent model behavior. Model transparency can be achieved either on a global scale for which the learned data representations can be visualized and clustered in order to be inspected by an expert, or on a local scale for which explainable AI can be used to attain fine-grained explanations that give detailed insight on which features were most relevant for the model prediction. In addition, the performance of the model should be validated on the original task of interest. While annotations can be costly to collect for this step, which in our setting requires the identification and labelling of same-class table pages, it is important to ensure that the original task of interest can be solved by the selected modeling approach.\par

\subsection{Data Availability}

\subsubsection{Data Infrastructure}\label{text:infrastructure}
Each of the 359 editions that form the Sacrobosco Collection, is represented by a single digital copy that is considered to be a representative sample of the entire edition print-run, resulting in a corpus that contains almost 76,000 pages. The result of this analysis is stored in a knowledge graph \cite{CorpusTracer18}. The knowledge graph is modeled according to the Conceptual Reference Model of the International Committee for Documentation (CIDOC-CRM) \cite{Bekiari2021}, as well as its extension for bibliographic records, FRBRoo \cite{Bekiari2015}. The CIDOC-CRM ontology and its extensions provide a useful and standardized framework for modelling and storing humanities and cultural heritage data; the framework also strives to create coherent and shareable datasets across research institutions. This ontology relies on a predefined set of classes and properties, as well as constraints, to ensure the consistent recording and storing of cultural heritage and humanities data \cite{Meghini2018}. Following the CIDOC-CRM standards, knowledge atoms were inserted into a knowledge graph, where entities (e.g., ``books'') are connected to each other through semantic relations validated by historians, effectively creating the \textit{Sphaera} Knowledge Graph \cite{ElHajj2022}.  \par

This knowledge graph forms the basis for all further investigation of the Sacrobosco Collection, and has expanded to be a number of times larger than its original size due to multiple consecutive historical and computational research cycles \cite{Kraeutli2021, ElHajj2022}. \par

The initial instance of the Sacrobosco knowledge graph stored metadata related to the physical version of the book, which included information that can be acquired by simply looking at each edition's digital copy. Such information included the individuals involved in the edition's production (e.g. author, publisher, printer, and/or translator) as well as the relevant information on the physical copy, such as number of pages, physical format and material, as well as the location of its printing and publishing. Further historical research gathered information on each person involved, such as their dates of birth and death, or alternatively their years of activity in cases when the former information is unknown, as well as mutual kinship relations. \par

\subsubsection{Data} \label{supplement:data}
In the following, we describe how we have obtained the Sacrobosco Tables corpus, provide details of the annotation process regarding ground truth for training the digit recognition model as well as the evaluation of model and historical analyses.
\subsubsubsection[]{Table pages}
\label{supplement:subsec:page_classifier}
From the approximately 76,0000 pages of the Sacrobosco Collection, we have selected 9793 pages bearing one or more numerical tables, which we submit to the table similarity workflow as the Sacrobosco Tables dataset. By numerical table  we refer to any tabular arrangement of data in our corpus which has at least one column with (predominantly) numerical content and specifically exclude tables of content and book indices. This selection was supported by an of-the-shelf CNN (VGG-16 \cite{Simonyan15}) trained to classify numerical table pages. The output of this CNN was checked down to a low probability of the assignment of a page as bearing a numerical table. Due to the human post-processing, the list of pages with numerical tables has virtually perfect precision and very high  recall. A list of all pages with numerical tables is provided as  \texttt{spharea\_tables\_meta.csv}, the trained model instrumental in establishing this list is provided as \texttt{sphaera\_tables\_classifier.h5}. The digital images of the pages, that we refer to as the Tables dataset of the Sacrobosco Collection can be obtained at \texttt{sphaera\_tables\_images.zip}.\par

\subsubsubsection[]{Preparation and acquisition of ground truth} Four different ground-truth datasets have been prepared to train and test our model, \emph{single digits} and \emph{non-digit content} to train the digit model, \emph{fully annotated numbers} to test the digit recognition and the bigram expansion and \emph{sun zodiac pages} to evaluate the table similarity model. These sets are provided as \texttt{numerical\_patches.csv}, \texttt{contrast\_patches.csv}, \texttt{digit\_page\_annotations.csv} and \texttt{sun\_zodiac.csv} in the code and data  repository.\par

\medskip
\paragraph*{Single-digits} In the period covered by our corpus print types where far less standardized than they are today. To capture the wide range of typological variations present in our corpus,  we have selected each printer that contributed at least one book to the collection. From the printed output for each of these printers we have selected (where possible) five pages bearing numbers and annotated on each of these pages five individual numbers by bounding boxes, annotating in addition the writing normal (upright, turned left, turned right). Single digit patches were obtained by dividing the annotation boxes into equal segments corresponding to the number of individual digits in the annotated number. As the types for the individual digits from zero to nine vary in width, this introduces  some error in the single digit patches that can be the larger, the greater the number of digits in the annotated number. After annotating about a third of the selected pages we thus decided to restrict the annotation on the remaining pages to digit bigrams (adjacent digits, regardless whether they form a two digit number or a part of a longer number) but retained the annotations produced before.\par

\medskip
\paragraph*{Non-digit content}
In order to correctly model non-digit page content such as text, illustrations or layout geometry, we extracted patches from non-table pages as contrastive examples.
\par

\medskip
\paragraph*{Fully annotated numbers} We have selected 11 pages and annotated each single digit contained on the pages by a bounding box. In addition, we have marked if the individual digit is the first and/or the last digit of a number. With this information, all numbers and thus also all bigrams contained on these pages can be reconstructed. The annotated pages have been selected to cover a wide spectrum of different manifestation of numerical content in terms of writing direction, fonts, fonts' sizes, density of digit placement on the page, etc.\par

\medskip
\paragraph*{Sun zodiac pages} To evaluate to what extent our approach can reproduce the salient relations between the tables in our corpus, we have chosen the sun-zodiac tables, which give the positions of the sun into the signs of the zodiac in degrees for each day of the year. This table was printed in varying layouts, where the different layouts partition the full table differently and in some cases the entire table is comprised on one page, in other books it is distributed over as many as nine pages. The sun-zodiac tables are thus a well-suited example for evaluating our approach in dealing with heterogeneous source material.
Moreover, due to its content the table only comprises numbers from 1 to 31 (max. 31 days per month, 30 degrees per sign of the zodiac). The table thus only populates a subspace of the feature space that we exploit for our similarity assessments. Since this subspace is more densely populated than would be expected with a uniform distribution of the data over the similarity space, this table is particularly difficult to discriminate under our approach which makes it a good test case.\par

Two variants of the sun-zodiac table were identified: tables for the times of the
`ancient' poets (`veterum poetarum temporibus accommodata') where the sun is 16 degrees into Capricorn on the first of January, and tables for  'contemporary' times ('nostro tempori') where the sun on the first day of the year has advanced 3 degrees and is located 21 degrees into Capricorn. This difference amounts to a shift of the columns listing the days of the year with respect to columns giving the angular locations. From the perspective of our similarity model that pools the identified numerical features spatially, these two variations represent the same (more abstract) table.\par

Altogether, we have identified 68 instances of the sun-zodiac table that cover 250 pages in the corpus. A list of the pages containing the different versions of the sun zodiac tables is provided as  \texttt{sun\_zodiac\_pages.csv}. A ground truth histogram for the digit-features distribution of a prototypical, i.e.  noise-free and complete, sun-zodiac table is provided as \texttt{sun\_zodiac\_hist.csv}.\par

\medskip
\paragraph*{Clime table Pages}
We further collect a subset of material focused on  climate zone tables. These tables divide the surface of the ``inhabited'' world and that can be defined by the length of the solar day. This served as an indication of the overall meteorological conditions, which was in turn a determinant information in the framework of Medieval and early modern medicine. We find three different principle variants of climate zone tables that either use 7, 9 or 24 clime zones. The 225 pages containing these tables are provided as \texttt{clime\_tables.csv}. Each row of the csv file corresponds to one individual clime table, specifying its variant and providing metadata for the edition containing this table.\par

\subsection{Limits and advantages of the application of machine learning and XAI to historical analysis}
\label{supplement:sec:ML_limits}

The number of sources analyzed in historical studies is contingent upon the research question and the epistemological approach chosen by historians. Recent trends in historical research have shifted attention to sources that were largely overlooked in the past, such as university textbooks in our study. The sheer volume of these sources surpasses the human capacity for analysis using traditional methods, especially close reading. Consequently, we propose complementing traditional historical analysis methods with the application of ML techniques. While the need to employ ML thus arises from research questions within the historical disciplines, the application of ML methods might ultimately allow for and prompt new forms of research questions in the future and thus enrich historical research.\par

The approach adopted in this study encounters general challenges associated with the application of modern ML methods. In particular, the inherent data-dependent nature of models brings to question out-of-domain generalization abilities, while the high non-linearity of these models further poses challenges regarding model interpretability. However, as demonstrated in our research, these limitations can be effectively addressed so that, by utilizing ML to assist historical interpretation, we can also surmount constraints inherent in traditional approaches based on close reading and, specifically, the constraints related to human resources. 
In the same vein, there are appeals for combining computational approaches with traditional in-depth analysis in a productive manner \cite{brausch2023machine}.\par

In terms of data dependence, a significant limitation when applying ML methods to historical research lies in the availability of well-curated data. Although we have outlined methods to address the inherent heterogeneity and varied quality of digitized historical source material in this study, our research still depended on a corpus of sources furnished with high-quality metadata. The meticulous preparation of the Sacrobosco Collection took several years and involved two senior historians, two post-doc fellows, and three student assistants. International collaborations further enriched our dataset by bringing additional scholars into contact with it. To apply our methods to other historical inquiries, there is a presumption that historical source data must be similarly enriched and contextualized with metadata.\par

With respect to the generalization of our ML methods, we have evaluated them according to well-established standards in ML. For each phase of our process, we have presented the relevant evaluation metrics using appropriately selected test sets. For instance, we have quantified the performance of our digit recognition model (\ref{supplement:sec:fully_annotated}), the bigram recomposition (\ref{supplement:subsubsec:fully_annoated}), and the clustering performance based on our representation (\ref{supplement:cluster_performance}). This demonstrates that, within our corpus, the applied methods offer satisfactory performance for the intended task. Since ML models can take undesired strategies for making correct predictions \cite{lapuschkin-ncomm19}, we have further used XAI methods to ensure that our learned representations do indeed use task-relevant features.\par

If our approach were to be applied to a different corpus of numerical tables, its ability to generalize to this new dataset would need to be assessed in a similar manner. For example, digit recognition might decrease due to the presence of different printers using unique type fonts. Additionally, the discriminative capability of our representation with respect to tables might be compromised if the distribution of numbers in the tables of the new corpus varies significantly. Should our methodology be transferred not directly, but in a structural manner, to analyze other elements in the sources, such as illustrations, tests akin to the ones used here would be required to gauge the model's ability to generalize to the material at hand.\par

To deal with the limited availability of labels and overall data samples, our atomization-recomposition approach is designed to reconstruct the information content of the tables up to a sufficient level of representation, i.e., the level of bigrams instead of full numbers. As we have pointed out, achieving complete reconstruction would be nearly impossible due to the absence of annotated data for a fully supervised model. Moreover, the effort to generate such data would be disproportionate to its benefits. Consequently, 
we  only retain the necessary information for our specific objective of identifying similar tables. It is, in part, due to this limitation that our method does not replace but rather complements traditional historical analysis. Based on its  representation, our model will never discern the mathematical astronomical `meaning' of a table, such as for instance providing the right ascensions for a particular celestial object. However, it can aid a historian who,  examining such a table, wishes to locate similar instances amidst vast datasets, thereby facilitating studies on their spatial distribution or temporal evolution.\par

Fundamentally, for the reasons mentioned, ML models will never and are not intended to capture the complete richness of historical sources; they can only represent specific aspects. The choice of these aspects is ultimately driven by the research interests of the historians. With these models, however, historians can tackle questions that are otherwise unapproachable, primarily due to scale constraints. Consequently, these two methodologies must complement each other. In doing so, they can invigorate the historical disciplines with novel approaches, methods, and insights.\par

If the path outlined in this paper is consistently pursued, it holds the potential to unlock intricate historical analyses, such as understanding the long-term interplay between texts and images. More pressingly, there is the possibility of automatically generating genealogies between texts even before engaging in a thorough reading. The next ambitious goal, following the current research, is to achieve this using our atomization-recomposition method. This task holds significant relevance within the historical disciplines. A major hindrance so far has been that historical sources often come in languages or language variants for which no well-curated datasets exist. Our method might help bridge this gap, enabling the identification of pertinent phenomena at the corpus level. Once this is achieved, it paves the way for pinpointing the right clusters of texts that can then undergo a close reading—essentially a case study informed by a  selection made possible through the corpus-level analysis with the assistance of the ML model.\par

\section{Supplementary Text}
\subsection{Insights about the Sacrobosco Collection - Analysis of Numerical Content}

The potential of our approach is best displayed by the fact that it enables, for the first time, an automated investigation of the astronomic tables across the entire corpus of textbooks. It puts us in the position to analyze trends over the entire corpus or large parts thereof and to reveal geographical singularities or semantic shifts over time. In this way, we also gained the possibility to develop case studies and, as will be shown, to reveal unexpected historical findings. We will first present results concerning the general process of mathematization of astronomy as it was taught at the European universities between 1472 and 1650, then move to a corpus-level analysis and investigate its temporal and spatial dynamics, and finally move to two important case studies that could only be conducted based on our approach. Some meta-methodological considerations will complete this section on the historical investigations.\par

\subsubsection{Mathematization of Astronomy in the Framework of Teaching as a Result of Institutional Competition \textit{alias} Insights from Numerical Histograms Using t-SNE} \label{supplement:text:CorLev_TSNE}

\begin{figure*}[h!]
    \centering
    \includegraphics[width=0.95\textwidth]{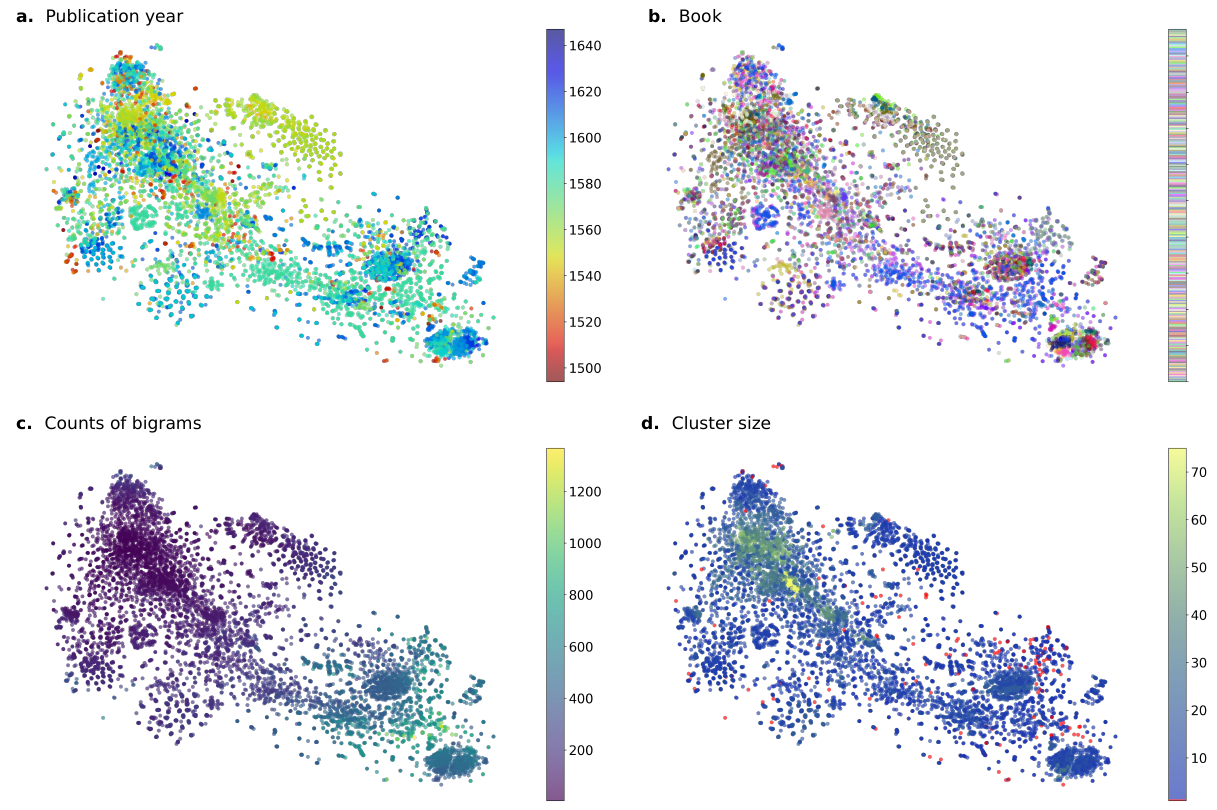}
    \caption{\textbf{t-SNE visualizations of the Sacrobosco Tables corpus.} Each data points corresponds to one page and color reflects additional meta-information (top rows) or model-based output (lower rows). We use the available data regarding (\textbf{a.}) the publication date and (\textbf{b.}) the book title to color each point. The extracted bigram histograms are used to visualize (\textbf{c.}) digit density of a page and (\textbf{d.}) the size of the assigned cluster for every page.
    }
    \label{fig:tsne_overview}
\end{figure*}

We start by inspecting the histogram embedding space of the Sacrobosco table pages with regard to additional  information about the collection. First, in Figure \ref{fig:tsne_overview} (top row), we use the meta-information regarding the publication year and the unique book identifier available for each book to color the t-SNE projected data points accordingly. As visible in Figure \ref{fig:tsne_overview}.a, the visualization using the publication year provides information about what pages were printed in close-by time periods \textit{and} are semantically similar, for instance the group of pages on the bottom right. We can further analyze this group by investigating from which editions these pages are extracted as indicated in Figure \ref{fig:tsne_overview}.b and find that these pages stem from multiple books. This allows domain experts to combine different layers of information and to gain corpus-level insights in order to develop hypotheses that can then be investigated further in a targeted analysis.\par

Second, we can add information from the automated analysis to the visualization. In Figure \ref{fig:tsne_overview}.c, we color code the bigram density on the pages and find that in the t-SNE projection the less dense tables are to be found in the top left corner and center whereas the very dense tables are predominant in the lower right corner. Finally, we show in Figure \ref{fig:tsne_overview}.d the size (number of cluster members) of the cluster that a page was assigned to in a $k$-means ($k=1500$) clustering. This tells us that most pages are contained in clusters of less than 30 members and only a small subset of pages is assigned to larger clusters of around 70 similar tables situated in the low bigram frequency domain.\par

\subsubsubsection*{Historical Interpretation Based on the t-SNE Projection}
\label{supplement:text:Histo_TSNE}

The region on the bottom-right of the embedding space shows a high number of semantically closely related table pages. From Figure \ref{fig:tsne_overview}.b, we see that these are from many different editions, and Figure \ref{fig:tsne_overview}.c tells us that they contain tables with a moderate density around 400 bigrams, and that are not assigned to great-size clusters as evident from Figure \ref{fig:tsne_overview}.d. Finally, against the background of the year-based projection (Figure \ref{fig:tsne_overview}.a), it is clear that this region hosts editions that were published starting from the mid of the sixteenth century and until the end of the historical time interval considered here, namely 1650. We exclude from this consideration those five early editions (the orange points on the left-bottom side of the region in Figure \ref{fig:tsne_overview}.a) as they are marginal in the projection.\par

\paragraph*{Mutual Awareness of Powerful Institutions}

By combining these different layers of information, we could identify a subgroup of editions that share a great number of semantically similar tables. Examining the group closer, we discover that the region is constituted by exactly forty editions (\texttt{data/corpus/Metadata\_year\_tsne\_bottom\_right.csv}). The printing dates range from 1551 to 1622. At first sight, a heterogeneity  of these editions seems to be implied by the fact that they were produced not only during a time interval of over 70 years but also in twelve different cities of Europe. However, upon examining the authors of these textbooks, we observe a peculiarity: thirty-six out of forty editions involve only four scholars. The first five editions, published between 1551 and 1556, are for instance five different texts all compiled by the French Royal Mathematician Oronce Finé and always published by the same printer and publisher Michel Vascosan in Paris \cite{Pantin2013}. Further twenty editions are commentaries on the original tract of Sacrobosco compiled by the then leader of the scientific section of the Collegio Romano (Christophorus Clavius), the center of the Jesuit Order where scientific knowledge was produced to sustain the innumerable Jesuit colleges all over Europe \cite{Grendler2022}. Starting in 1582 we find eight treatises compiled by Thomas Blebel in Wittenberg. The dominant role of Protestant  Wittenberg in producing and disseminating scientific knowledge between 1530s and 1560s has already been demonstrated \cite{mva19}. The scholar Thomas Blebel has not been hitherto investigated by historians of astronomy but the findings based on our new method strongly suggest  that Blebel's works represented an attempt of the Wittenberg community to cope with the works and the success of influential and institutionally powerful scholars such as Finé and Clavius.
Finally, we find further three editions written by the influential late Italian astronomer Francesco Giuntini, who distinguished himself thanks to the introduction of a series of scientific very long-lasting innovations \cite{SRN2020} under strong clerical patronage. In conclusion, this region of the embedding space represents clusters of editions generated in the frame of powerful institutions and communities and the fact that they contain similar tables means that they were observing and imitating each other, possibly due to the influence of cultural and institutional competitions among them.\par

As numerical tables in scientific textbooks are the external indication of the process of mathematization (Section \ref{text:table_meaning}),  we can hypothesize that one of the driving forces  of the mathematization of astronomy during the second half of the sixteenth century was an institutional competition that involved the Paris scientific institution expressed by the Royal power, the Protestant leading university of Wittenberg, the Jesuit order, and single scientists working in an institutional well-protected context.\par

From a methodological point of view and even without looking at the individual tables contained in  these editions (which we do in the next sections) our method allows us to draw historical inferences and reach historical relevant conclusions based on a collocation of material and metadata that otherwise would be non-accessible even to domain experts. Moreover, it can also be stated that our method helps generating specific historical micro research questions by identifying singularities in great corpora of sources.\par

\begin{figure*}[h!]
    \centering
    \includegraphics[width=1.\textwidth]{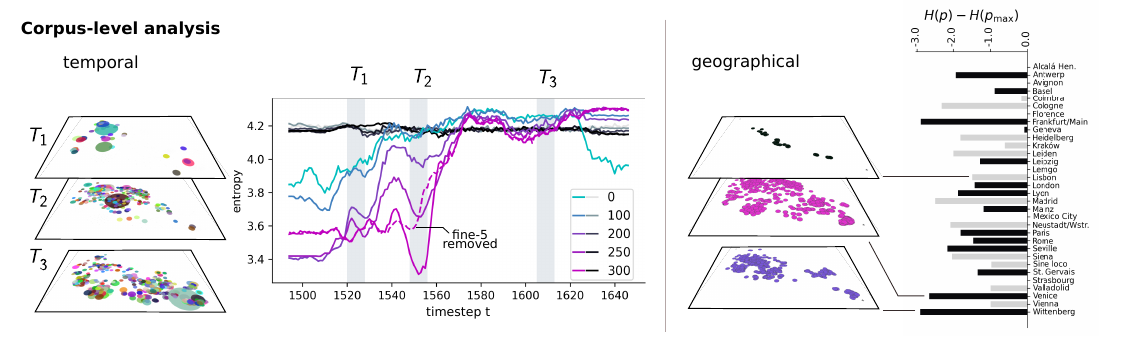}
    \caption{\textbf{Corpus-level analysis.} \textit{Left} Temporal evolution of knowledge displayed by computing the entropy of cluster membership vectors (number of tables in each cluster) for each time step. Gray to black lines correspond to a random embedding baseline, colored lines correspond to the data from our collection. Different colors indicate a filtering threshold on the digit density per page, e.g. all pages containing at least 100 digits. The clusters are shown as t-SNE visualization for three time intervals with the disk diameter of the active clusters set to be proportional to cluster size. We observe a marked drop in entropy for tables with extensive numerical content between 1540 and 1560. This drop disappears after removal of the \textit{Fine-5} group, a subset of tables that occur in the editions authored by Oronce Finé that we identified as the dominant factor driving the entropy change. \textit{Right} Geographical analysis of knowledge distribution for each print location in alphabetical order using relative entropy. Low-output cities (\textless=100 tables) are colored in light gray. For three selected cities a t-SNE visualization of the distribution of the printed tables is provided.}
    \label{fig:corpus_level_supplement}
\end{figure*}

\subsubsection{Corpus-level Analysis 1: Temporal Dynamics of Mathematization of Astronomy} \label{supplement:text:history_temporal}
Moving to the corpus-level analysis, we first investigate the temporal dynamics of the process of mathematization of astronomy during the early modern period by investigating the temporal dynamics of the entropy of the distribution of high-density numerical tables over clusters of similar tables. The editions of the Sacrobosco Collection that contain at least one numerical table were printed during a time span of 153 years (1494-1647). Over this time span  publication rates changed considerably. Thus, we apply a sampling based temporal analysis. For each time step $t_i$ we assign a sampling probability to each book page from a truncated normal distribution $\mathcal{N}(t_i, \sigma^2 )$ which sets probabilities for data points outside the interval $(t_i - \sigma, t_i + \sigma)$ to zero. At every step we sample $N=80$ data points, determine their cluster membership label, construct the cluster count histogram of size $1\times k$ and compute the  entropy $H(p_{cl}) = -\sum_k p_{cl,k}\log(p_{cl,k}) $ of the cluster probability vector $p_{cl} \in \mathbb{R}^{1\times k}$. We compute this for digit density thresholds of $\{0, 100, 200, 250, 300 \}$ and average entropy curves over 20 runs for each threshold. Results are shown in \ref{fig:corpus_level_supplement} (left).\par

\begin{figure*}[h!]
    \centering
    \includegraphics[width=0.7\textwidth]{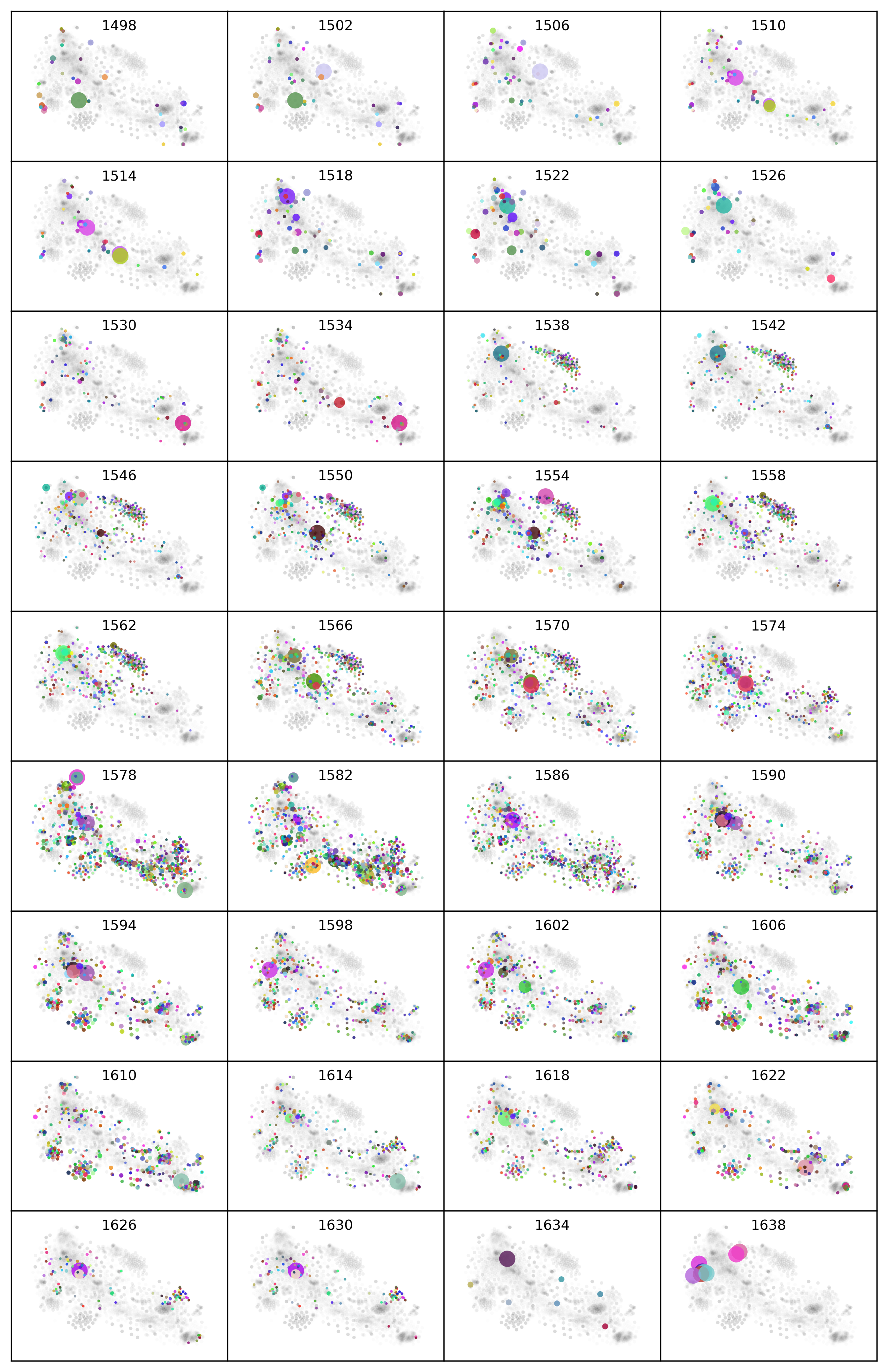}
    \caption{\textbf{Temporal dynamics of printing astronomical tables.} Each panel corresponds to one time point of the full corpus evolution. Clusters that contain published pages from this period are represented by a colored disk whose diameter is proportional to its cluster size. }
    \label{fig:evolve_all}
\end{figure*}

In Figure \ref{fig:evolve_all} we present additional cluster visualizations throughout the corpus evolution. Disk color codes for cluster membership and its size is proportional to number of cluster members at this time step.\par

\begin{figure*}[h!]
    \centering
    \includegraphics[width=0.9\textwidth]{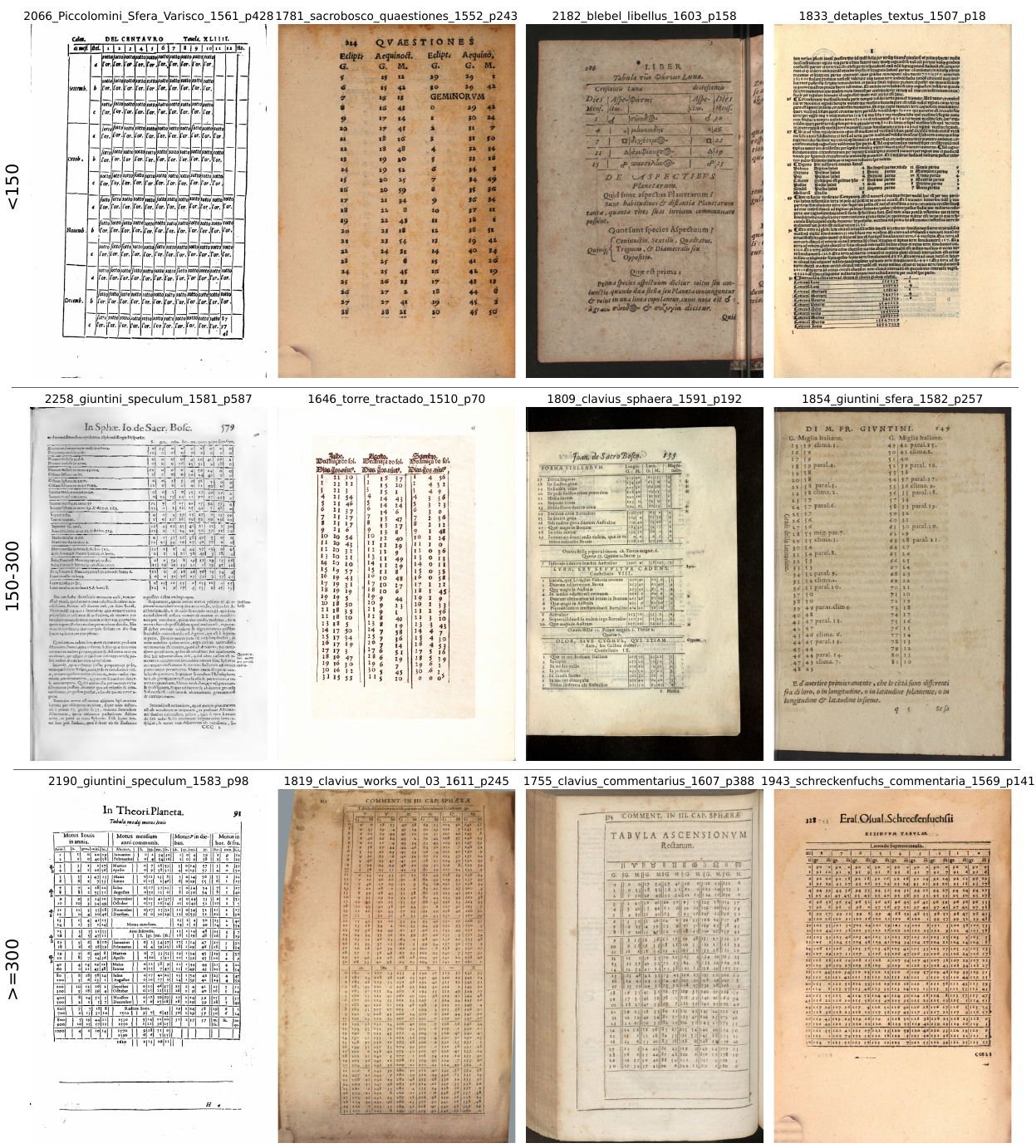}
    \caption{\textbf{Sacrobosco Table Pages grouped by digit feature density.} Top to lower rows  correspond to increasing digit feature density levels, i.e. the first rows shows pages that contain less than 150 digit features as extracted by the bigram network. }
    \label{fig:sphaera_density}
\end{figure*}

As shown in Figure \ref{fig:evolve_all}, it remains challenging to visually discern  whether significant changes occurred during the evolution of the corpus. However, we found that focusing on high-density tables reveals  significant changes in entropy over time over the full corpus. These changes are far less pronounced if all tables are taken into account. This can be explained by the fact that low-density tables carry less specific mathematical information and these do not vary greatly over time but, instead, often contain more basic information such as enumerated lists. We show exemplary pages grouped by different density levels in Figure \ref{fig:sphaera_density}.\par

We additionally validate our results against a baseline in which we randomly sample histogram representations $h_{rand} \sim \mathcal{N}(0,1)$. This serves as a model of a knowledge process that does not consider any evolution of information or knowledge transfer across printer locations and publication dates and thus is expected not to show any significant entropy changes. We confirm this as presented in Figure \ref{fig:corpus_level_supplement} (left).\par

\begin{figure*}[h!]
    \centering
    \includegraphics[width=0.8\textwidth]{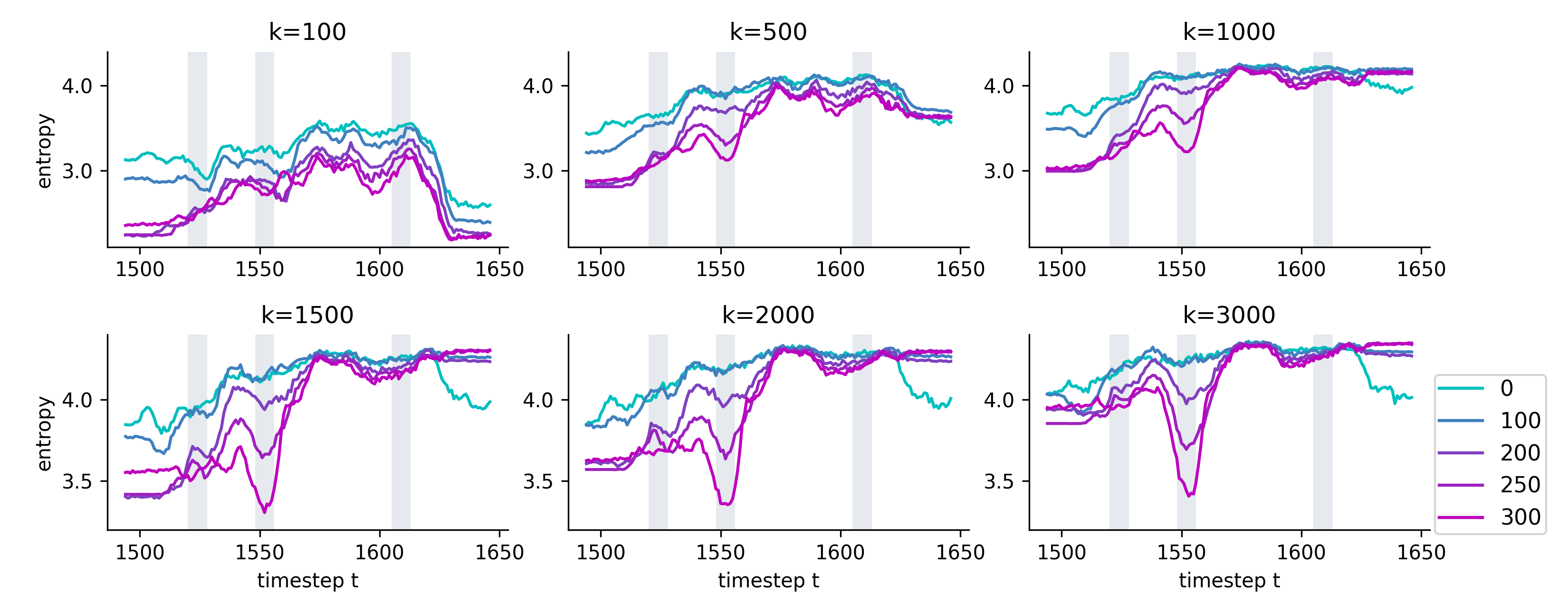}
    \caption{\textbf{Control study for different numbers of clusters.} Entropy evolution for different number of clusters $k=\{100,...,3000\}$. The different line colors correspond to a digit density filter of the pages, e.g. using all pages that contain at least 100 bigram features.}
    \label{fig:evolve_kstudy}
\end{figure*}

Since the entropy evolution analysis is based on the clustering assignment of pages, we want to control that our results are robust and consistent for a range of reasonable number of clusters $k$. In Figure \ref{fig:evolve_kstudy}, we repeat the analysis for  $k=\{100,500,1000,1500,2000,3000\}$ and observe that if the clusters are sufficiently large for small $k$, we are not able to observe strong temporal changes as visible for $k=100$ since the clusters are semantically too diverse. For, $k=500$ we start to observe the emergence of the entropy drop for high-density pages between 1550 and 1560 which becomes more and more visible for increasing number of clusters. Thus, we conclude that our observation of the entropy singularity is not an artefact of a specific clustering solution but can be observed for a reasonable k-means clustering solutions.\par

\begin{figure*}[h!]
    \centering
    \includegraphics[width=0.8\textwidth]{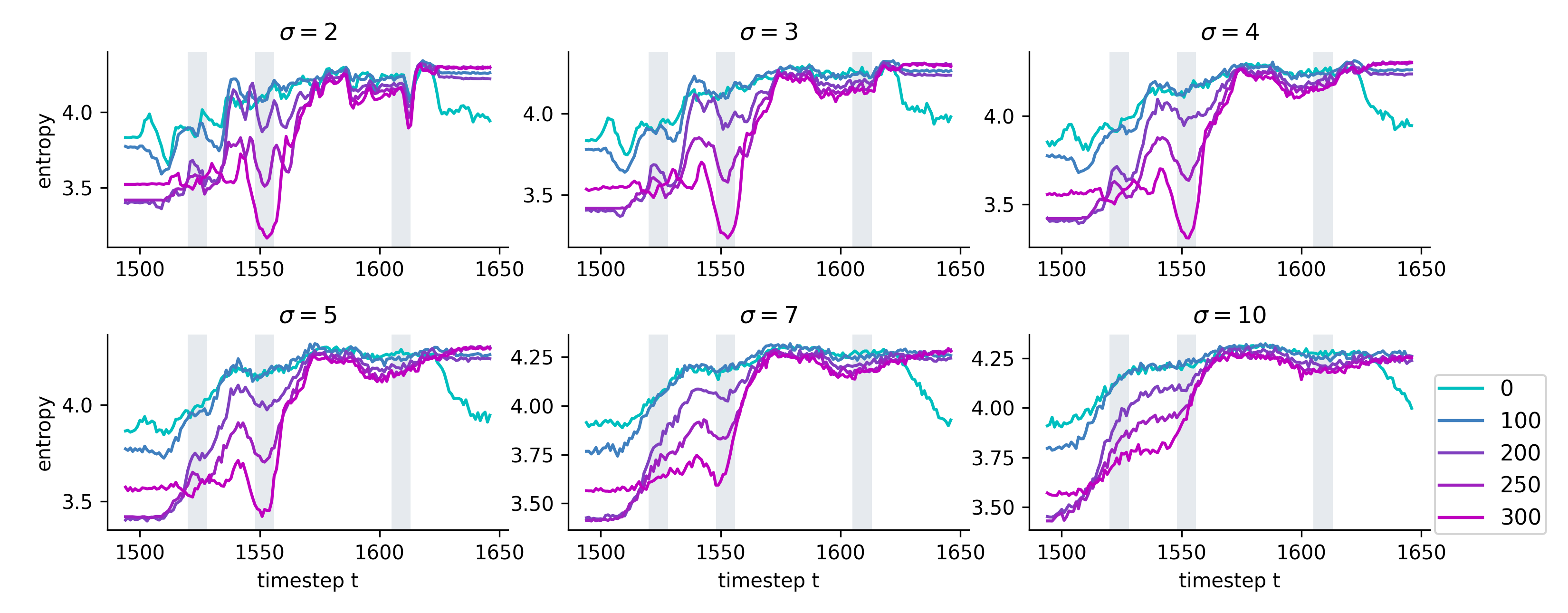}
    \caption{\textbf{Control study for varying time windows.}  Pages are sampled from a Gaussian distribution centered at current time step $t$ with a standard deviation of $\sigma=\{2,3,4,5,7,10\}$. }
    \label{fig:evolve_sigmastudy}
\end{figure*}

Next, we investigate the effect of the standard deviation $\sigma$ used to sample pages at each time step $t$.  A smaller $\sigma$ indicates a more narrow time window used to sample pages from the corpus for a given time step. For $\sigma = \{2,3,4,5,7,10\}$ we present the entropy evolution analysis in Figure \ref{fig:evolve_sigmastudy} and observe that for reasonably small $\sigma$ values the entropy drop is maintained. This is in line with the explanation for the drop which will be advanced below. Only for larger values of $\sigma 	\geq 7$ we can see that the effect of sampling temporally more distant pages results in a smoothing of the entropy curve and vanishes for  $\sigma=10$.\par

To consolidate the entropy drop observation we extract the Sacrobosco Table pages that are the main drivers of the entropy change. For this we look at the time between $t=\{1540,...,1560\}$ and compute for each time step the clustering distribution $p_{cl,t}$ and entropy $H(p_{cl,t})$. We rank time steps according to the strongest absolute change $|H(p_{cl,t}) - H(p_{cl,t+1})|$ and find that this occurs for $t^*=1553$. Next, we look at the change in clustering distribution $p_{cl,t^*} - p_{cl,t^*+1}$, rank which cluster has gained or lost the most relative members and historically investigate these relevant clusters and table pages respectively. The analysis reveals that during this period the same work,  Oronce Finé's  \textit{Sphaera}, has been repeatedly reprinted in five books to which we refer as the \textit{Fine-5} group\footnote{The group \textit{Fine-5} is constituted by three Latin editions (\cite{FineLat1551}, \cite{FineLat1552}, \cite{FineLat1555}) and two French ones (\cite{FineFrench1551}, and \cite{FineFrench1552}.}.\par

\subsubsubsection*{Historical Interpretation and Confirmation of the Temporal Corpus-Level Analysis} 
Lower entropy can suggests that scientific knowledge encoded in numerical tables stayed closer in its distribution to previously published material and/or the addition of semantically new tables while higher entropy can signal a more homogeneous distribution. The first aspect shown by the temporal analysis is the drop of entropy starting in 1551 in a rather short time window of 3 to 5 years. Historically to assume a rather drastic development in such a short time window becomes plausible if we take into account   the practice of the printers and publishers involved in the emerging academic book market during the sixteenth century. When a novelty was introduced, printers and publishers could obtain a so-called \textit{privilegium} upon request. A \textit{privilegium} is the forerunner of what is now called copyright and applied only to the book as a product and was therefore originally conceived to protect the work of print-shops's owners. Usually, however, a \textit{privilegium} was valid for only a few years, mostly only 2. This implies that once a new treatise was granted a \textit{privilegium}, the printer and/or publisher had every incentive to saturate the market with the same treatise. A normal practice, which also had the advantage to limit the financial damages caused by the second-hand market, was to produce a large print-run (which in itself was a way to lower the production costs per copy) and then to place portions of the same print-run on the market every year . Only a new title page, with a new date of publication, had to be printed anew in order to be allowed to claim that a new edition had been published. The new edition was therefore not really a re-print but more properly speaking  a re-issue. Moreover, because of the fact that books were not sold  bound, like nowadays, but as piles of printed sheets that were then folded and bound at the book shop, printers and publishers always had the possibility to replace or add just a few sheets in order to claim that the new edition was indeed ``really'' new. Because of these reasons, it was very frequent that a new treatise, with high potential for international success, was pushed into the market by means of a series of editions published during a relatively short time interval \footnote{For a comprehensive study of the economic rules of the academic book market during the early modern period and specifically related to the sources of which the Sacrobosco Collection is constituted, see \cite{VallerianiOttone2022} and, in particular \cite{VallerianiOttoneChap12022} and \cite{Maclean2022}.}.\par

By looking closer at the group \textit{Fine-5} that our corpus-level temporal analysis has identified as responsible for the entropy drop, it indeed turns out that the five editions are in fact, one Latin edition of 1551, a related re-issue in 1552, a slightly changed re-print of the same in 1552, a French edition of the same book also published in 1551 and a related re-issue in 1552 following the market mechanisms of printing outlined above. Moreover, these works also belong to the bigger cluster that involves forty editions identified by means of the t-SNE projection discussed in section  \ref{supplement:text:CorLev_TSNE}.
Bringing together these two findings enables us to conclude that the institutional competition mentioned earlier, which was a harbinger of the process of mathematization of astronomy, made use of commercial mechanisms developed within the context of early modern book marketing \cite{RN2923}. This has not adequately been comprehended before.\par

Moreover, the clear trend can be discerned for the entropy to rise until roughly 1570 when saturation sets in. We know that all the editions of the collection focus on the same core knowledge and these are printed in an increasing number of places and reach an ever widening audience in this period (knowledge homogenization). At the same time, however, the entropy trend means that novel content attaches to this common core in different ways (innovation) during the first 100 years of the period considered. Finally this implies that the process of mathematization and of diffusion of scientific innovations was going hand in hand with the process of homogenization of scientific knowledge. This important historical and epistemological result is be deepened by the successive corpus-level analysis presented in the next section that concerns the variance in the spatial distribution of the process of mathematization of knowledge as represented by computational tables.\par

\subsubsection{Corpus-level Analysis 2: Spatial Variance of Mathematization of Astronomy } \label{supplement:text:history_spatial}

In order to study the varying knowledge production expressed by the tables printed across 32 different printing centers, we compute for each its \textit{relative entropy} score as presented in Figure \ref{fig:corpus_level_supplement} (right).
This score captures the difference of entropy between the observed cluster distributions and an uninformed uniformly distributed production process \mbox{$H(p)-H(p_{\max})$} with  $p_k$ being the probability of assigning a table to cluster $k$ and $H(p_{\max}) = log(N_c)$ with $N_c$ denoting the number of tables printed in city $c$.\par

The latter quantifies for each city the entropy of a hypothetical print process that is unrestrained, i.e. without memory of its print history and without outside influences, a scenario in which none of the printed tables is expected to be similar to any other. The relative entropy can be understood as a measure of the redundancy created by the actual process of content production and distribution in print as compared to this hypothetical process for each location. While a certain degree of redundancy can be considered a necessary precondition of stable and successful knowledge transmission, a too high redundancy would mean stagnation as is does not leave room for novelty. Our analysis in Figure \ref{fig:corpus_level_supplement} (right) shows that relative entropy varies strongly between print locations and that the minimum is reached for the cities of Frankfurt am Main and Wittenberg. This indicates that many tables are formed around the same clusters in comparison to an unconstrained print output. This result means that astronomic tables printed in the treatises produced in Wittenberg and Frankfurt are more homogeneous and therefore that textbooks in general were more similar to each other than those produced in other regions.\par

\begin{figure*}[h!]
    \centering
    \includegraphics[width=0.7\textwidth]{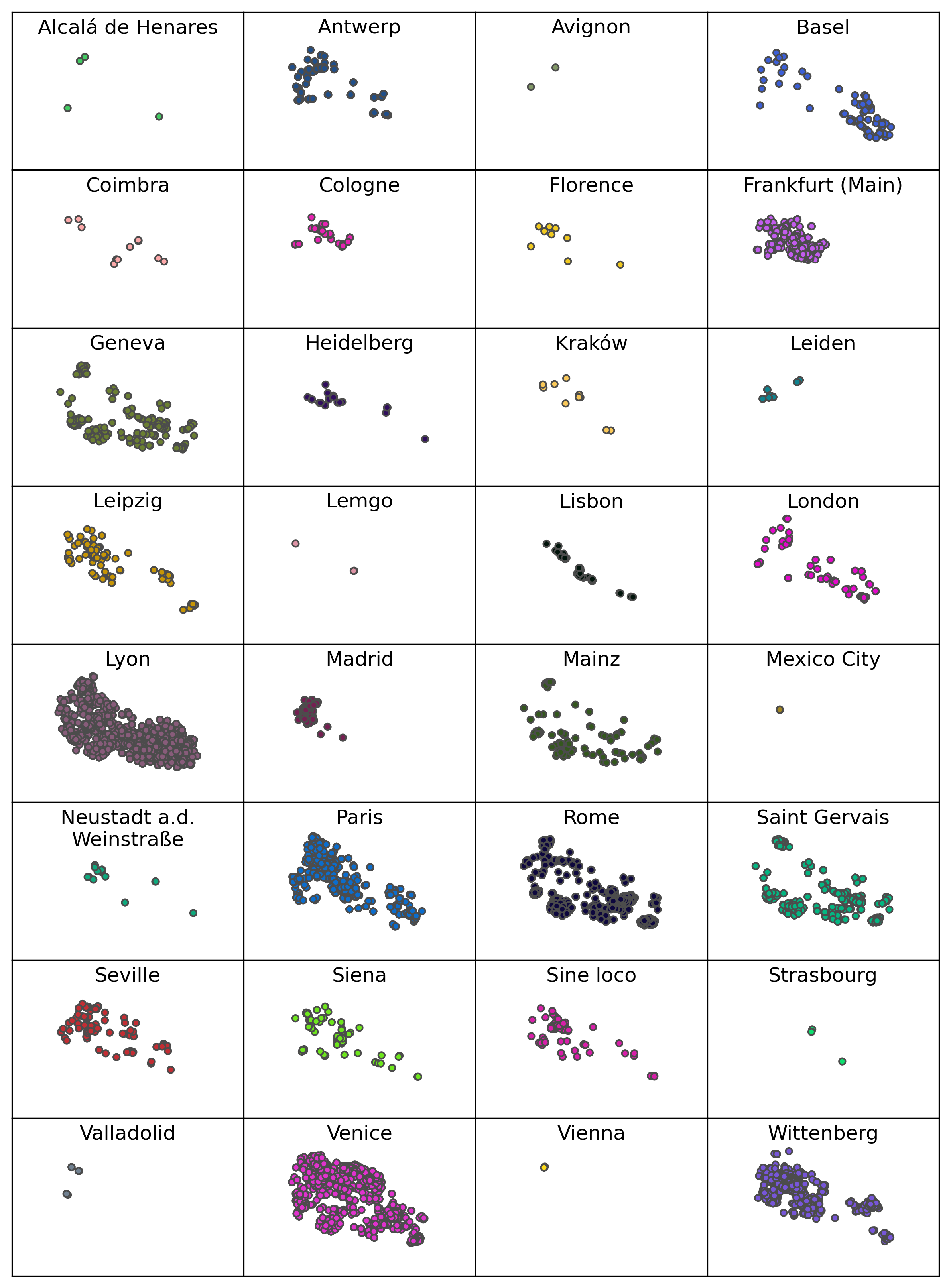}
    \caption{\textbf{Geographical distribution of printed table pages.} Each point corresponds to one page from the Sacrobosco Tables Collection and locations are sorted alphabetically.}
    \label{fig:supp_cities_tsne}
\end{figure*}

In Figure \ref{fig:supp_cities_tsne}, moreover, we provide a t-SNE visualization for each of the different print locations in the corpus, which allow further insights into the content variety and output quantity of the different cities. While `Alcalá de Henares', `Strasbourg', `Lemgo', `Vienna', `Mexico City' and `Avignon' have each printed less than five table pages, the most productive print centers have been `Lyon', `Venice', `Wittenberg', `Rome', `Frankfurt (Main)', `Paris' and `Saint Gervais' with at least 500 printed table pages. From the distributions of these pages once can moreover extract locations that exhibit a similar print program. For example `Rome', `Saint Gervais' and `Geneva' have printed content that covers comparable regions as visible in Figure \ref{fig:supp_cities_tsne}. In parallel, `Lyon' and `Venice' show the widest coverage across the embedding space.\par

\subsubsubsection*{Historical Interpretation and Confirmation of the Spatial Corpus-Level Analysis} 
The corpus-level analysis matched with further geographical data has provided two results. The first concerns the homogeneity of the output of tables in the treatises produced in Wittenberg and Frankfurt am Main. The second the similar t-SNE projections for some places of publication such as Geneva, Mainz, and Saint Gervais, as well as the dissimilar features for instance of the projection for Venice.
As for the first case, historical analysis can confirm that the drop of relative entropy in Frankfurt is due to the fact that a great part of its book production was constituted by many reprints of the same edition, the prime instance of a redundant production, which eventually contained a high number of tables. \par

In Wittenberg, however, the case is different. It is known that the main Protestant Reformers Martin Luther and especially Philipp Melanchthon meticulously designed and supervised the curriculum of study of the Wittenberg university \cite{Jackson2013,RN2847}. We also know that they worked in close contact with the numerous printers and publishers that had moved their businesses to this town after the Reformation \cite{Hennen2022}. Finally there is a text written by Melanchthon, the famous ``praeceptor Germaniae'' (Germany's instructor) to motivate students toward the study of mathematical disciplines and especially cosmology and astronomy. It  was first printed in 1531 and reprinted at least another 63 times until 1619 \cite{Melanchthon1531}. Based on all these facts we can surmise that the homogeneity of the mathematical apparatus of the treatises produced in Wittenberg, as discovered by the application of our model and our analysis, was possibly due to a political control of scientific knowledge executed by the Reformers, who were most certainly aware of the great influence that Wittenberg scientific treatises had all over Europe as mentioned above in Section \ref{text:knowledge_systems}. This interpretation is consistent with and backed by the fact that the scientific visual apparatus for astronomic studies developed in the same period mostly in Wittenberg remained constant for many decades and was also highly influential all over Europe \cite{Pantin2020, Limbach2022}. This suggests that the scientific output of the recently reformed Wittenberg had a significant influence on the linked processes of mathematization and homogenization of knowledge throughout Europe.\par

This cases of Frankfurt and Wittenberg, moreover, show  how easily our model and suggested analysis can identify singularities within a large volume of historical sources.\par

Coming to the second analysis, we focus on the projections for Mainz, Saint Gervais, and Geneve, as they clearly show a similar distribution. Also in this case, by means of a simple query of the data for those places of publications, we can easily have a closer look at those treatises that are mostly responsible for the pattern observed in the t-SNE projection. We immediately discover that the great majority of treatises produced in these locations actually are many different editions of the same commentary, possibly slightly changed over time and these are the treatises authored by Christophorus Clavius, namely those treatises that were also identified by means of a closer look at a specific cluster determined by the general t-SNE projection and discussed in section \ref{supplement:text:CorLev_TSNE}. This finally is a further confirmation of the overall picture of a series of centers and scientists in competition against each other to conquer the European academic book market by pushing the discipline toward a more sophisticated level of mathematization.\par

From the opposite historical perspective, we can briefly observe the projection of table pages produced in Venice, as its distribution is different from all the others. In this case, it is immediately evident that Venice's production of scientific treatises is not reducible to specific editions and their re-issues or re-prints. Rather, Venice's production of scientific treatises is clearly characterized by a great heterogeneity of scientific traditions which in turns fits very well with the widely accepted historical view that Venice became the most relevant international center of printing in Europe already at the end of the fifteenth century and maintained an extremely relevant position on the market for at least one century \cite{Nuovo2013}.\par

\subsubsection{Table similarity and Historical Case Studies} \label{supplement:text:case_studies}
While the primary goal of our approach is to obtain an overview of all the available materials, our model also allows for a different approach that focuses on specific and detailed interests. Thus, if for instance a historian wants to analyze the diffusion of  a specific table - identified either through close reading of the text or because of its position within a particular cluster in the embedding space - we have developed a tool to input the image of that table in order to identify all similar tables. This creates a group of tables that, once matched with metadata, provides all necessary information to the historian in order to conduct specific case studies. The tool enables users to query the corpus and to find all tables in the corpus similar to a query table they provide, either in original format (page scan) or as a ground truth histogram.
From a historical-methodological perspective, this approach implies that machine learning first enables us to conduct a corpus analysis, which serves as the backdrop against which case studies are then selected. In other words, the relationship between micro and macro history is further enriched by the possibility to move from one to the other level in both directions.\par

In the following, we present two case studies as examples of such an approach, one concerned with the \textit{Climate-Zones} tables and the other with the \textit{Sun-Zodiac} tables. At the end it will be shown how these two very different case studies together allow for a general historical contribution.\par

\subsubsubsection[]{Historical Case 1: Tables of Climate Zones} \label{supplement:text:case_study_climate}
In his \textit{Tractatus de sphaera} Sacrobosco's picks up on a topic that has its origins in ancient Greek astronomy and geography, namely the subject of the different climes or climate zones as they will be referred to in the following. The climate zones quite generally divide the surface of the ``inhabited'' world into regions or bands bordered by circles of the same latitude. Climate zones are formally defined by the length of the solar day used as an indication of the overall meteorological conditions which was in turn a determinant information in the framework of Medieval and early modern medicine. Sacrobosco's discussion of the climate zones is found at the end of chapter 3 of his treatise where he introduces the concept and provides the essential data defining seven climate zones. The original treatment by Sacrobosco is picked up in the majority of the books in our corpus. Thereby more often than not the data for the climate zones, which Sacrobosco himself renders as a text, is presented in form of a table. Our approach has allowed us to identify all climate-zone tables in the corpus  \footnote{For the complete list of such tables, computationally extracted from the corpus and amended by a human expert, see Subsection \ref{supplement:subsec:page_classifier}.}.\par

Due to the inherent characteristics of our model, we are able to identify not only tables that contain data on the seven climate zones, but also tables that are similar, for instance in that they contain additional information. These tables reveal the introduction of a novelty with respect to Sacrobosco's original treatise, which represented ancient and medieval knowledge. Besides tables listing the seven climate zones, the corpus contains instances of tables expressing a division into nine climate zones as well as tables representing a division into twenty-four climate zones. In the following we will analyze this finding and  the spread pattern of the occurrences of these three variants as evidenced by our collection and attribute historical reasons for those patterns.\par

\paragraph*{The Tradition of the Climate Zones}
\textit{Clima} in the ancient Greek tradition initially simply meant inclination and, if applied in geography, it specifically expresses latitude of a location on earth. In Ptolemy's time (1st cent. BCE), \textit{clima} was indeed predominantly related to terrestrial latitude. Latitude circles were usually referred to as parallels (i.e. circles parallel to the equator).\par

For places with the same latitude, numerous observable astronomical phenomena are the same. Thus for instance the length of the day is the same everywhere at the same latitude and thus in particular also the length of the longest day of the year at the summer solstice. Indeed before the introduction of a latitude grid, a common way to express the latitude of a place quantitatively was exactly to specify the length of the longest solar day at that place \cite[23]{Shcheglov:2004ul}.\par

In the second century BCE, Hipparchus had already furnished the mathematical relation between the length of the longest day specified in hours and the latitude specified in degrees. Initially, the expression \textit{clima} seems to have been used for any latitude expressed by length of longest day.\footnote{Otto Neugebauer has rightly remarked that measuring the day-length instead of simply the pole height is much more complicated and less precise. He speculates that specification of latitude by day-length was practiced nevertheless because of the greater practical relevance of the latter. Cf. \cite[p. 23]{Shcheglov:2004ul}.}

In the \textit{Almagest}, Ptolemy included a list of parallels for increasing day-lengths of the longest day from 12 to 24 hours, starting at the equator first in steps of 15 minutes and later in steps of half hours. Of these parallels he marked seven explicitly as \textit{climata}  and then he specified, in addition to their latitudes in degrees, the city or some notable geographical feature the respective parallel runs through. This system, which Ptolemy  later also included in his \emph{Geography}, was soon accepted as canonical and it ``radically changed the meaning of the term $\kappa\lambda\iota\mu\alpha$''.

\textit{Clima}, indeed, besides the association with the parallel, also came to acquire the meaning of ``region'', i.e a belt or zone of certain width around a particular latitude circle in which certain celestial phenomena ``do not change appreciably.'' This explains why the words `clime' could be used either for the parallel more specifically or, more generally, for the band around it, as under most perspectives this did not make a practical difference. It was moreover alleged that the climate and related phenomena such as vegetation were similar within these zones\cite{grasshoff2017living}. The seven zones subdivided that portion of the Earth surface that was considered habitable in antiquity, though it was already known that people were leaving also outside that zone. As life outside of the defined habitable surface portion was considered uncomfortable because of excessive heat or cold, the other zones were then just ignored. No climate zones were in particular specified for the southern hemisphere\footnote{The literature shows a dispute regarding the origin of the understanding of clime as climate zones. According to Ernst Honigman and Fuat Sezgin \cite{Honigmann:1992tt}, this idea traces to Eratosthenes or earlier. David R. Dick argues for it to have originated with Hipparchus in \textit{Posidionius} \cite{dicks_1955}}.\par 

Sacrobosco himself harks on a tradition of transmission of knowledge concerned with the climate zones as defined by Ptolemy: seven clime parallels as marking their centers and being boarded by parallels defined by the longest day being 15 minutes shorter or longer respectively.\footnote{In the \textit{Phases of Fixed Stars and Collection of Weather Changes}, Ptolemy himself only uses five \textit{climata} (cf. \cite{2014BlgAJ..20...68N}).} Thus, the first clime was understood as defined by the parallel for the maximum day-length of 13 hours as its center, the parallel for the day-length of 12 hours and 45 minutes as its southern and that of 13 hours 15 minutes as its northern confine. At the same time this was the southern confine of the second climate zone, i.e. the seven zones where perceived as as being directly adjacent. Primarily via their number, the seven climate zones were related to the seven planets and thus also assumed astrological and therefore medical significance. The climate zones understood in this way ``became one of the basic, canonical elements of late antique medieval European and Arabic geography'' and as such were also picked up by Sacrobosco \cite{Shcheglov:2004ul}.\par

\begin{figure}[h!]
    \begin{minipage}[b]{0.49\linewidth}
    \centering
    \includegraphics[width=\textwidth]{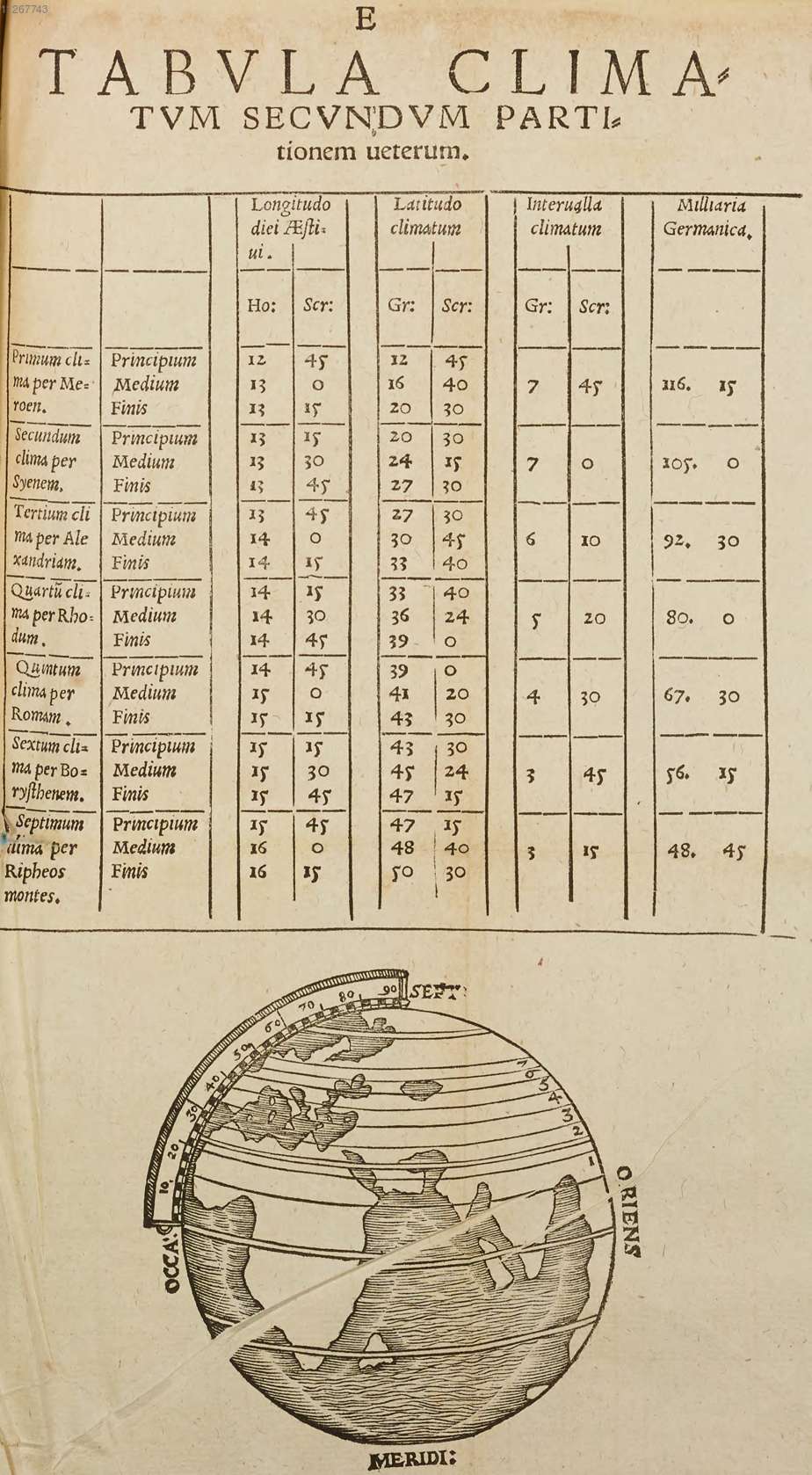}
    \caption{\textbf{Seven climes table.} \cite[Q4-8]{Peucer1558}. Augsburg, Staats- und Stadtbibliothek, urn:nbn:de:bvb:12-bsb11267743-1.}
    \label{fig:seven_sample_a}
    \end{minipage}
    \hspace{0.1cm}
    \begin{minipage}[b]{0.49\linewidth}
    \centering
    \includegraphics[width=\textwidth]{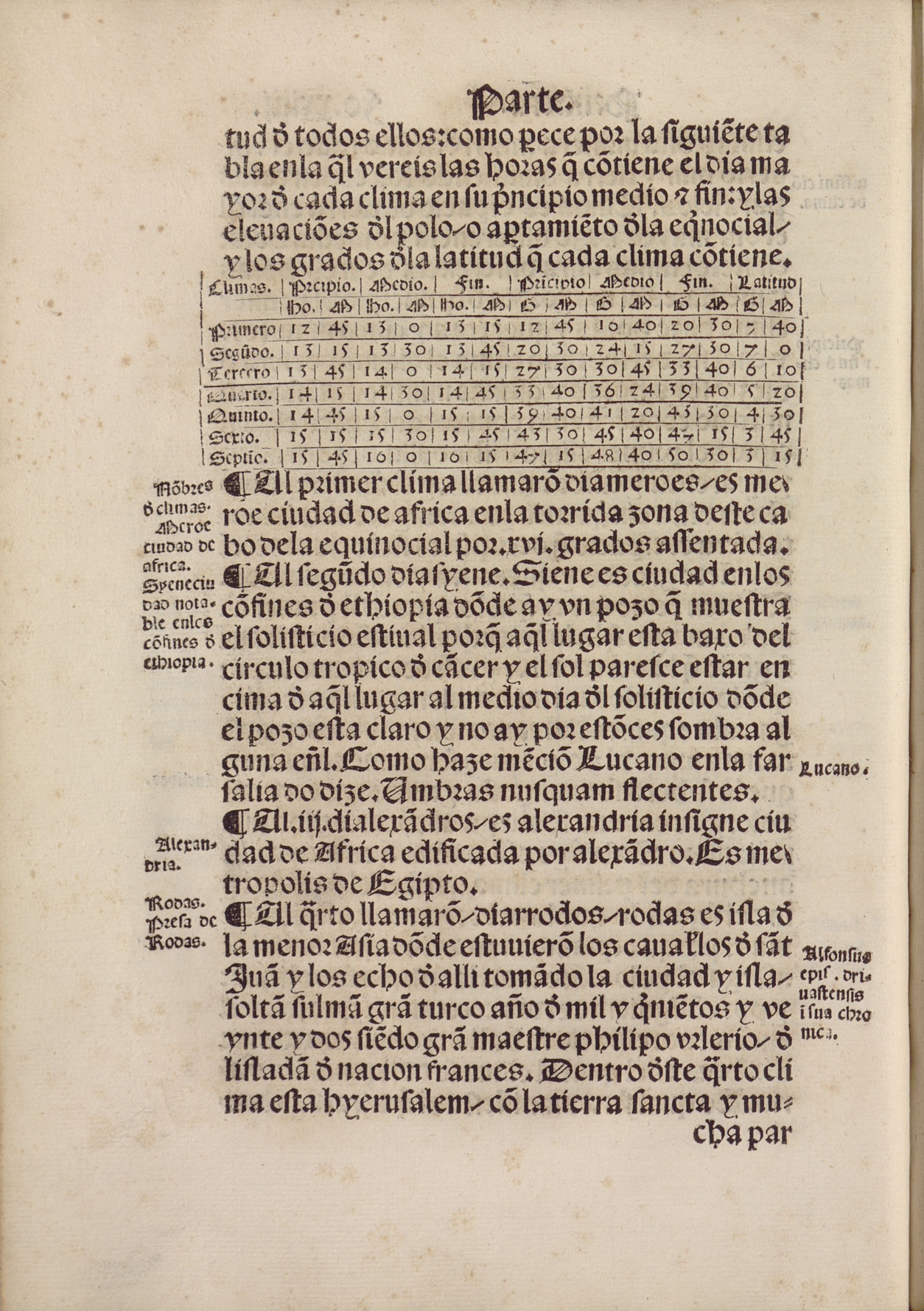}
    \caption{\textbf{Seven climes table.} \cite[XXIIIIv]{Cortes1556}. Biblioteca Nacional de España, bdh0000254979.}
    \label{fig:seven_sample_b}
    \end{minipage}
\end{figure}

\subsubsubsection*{The Climate-Zone Tables in the Sacrobosco Collection}
\subparagraph{\textbf{Seven climate zones:}} With the help of our approach we could identify 117 tables containing data concerned with only the seven climate zones. These tables show different attributes in various columns, though the sorts of attributes and the number of columns in a table can considerably vary, as the examples in figures \ref{fig:seven_sample_a} and \ref{fig:seven_sample_b} clearly show. Usually, a column is displayed for the latitude given in degrees and minutes. Almost always a column is present that shows data of the length of the longest solar day given in hours and minutes at a specific latitude. In the majority of cases there is also a column naming a place where the central parallel of the zone runs through. Moreover, there can be a column for the width of the zone given as an angle and/or arc length of the zone sector. If present, the arc length is specified with respect to different units in different tables (e.g. stadia, German miles etc.) and hence the numbers in this column, if present, can greatly vary.\par

For each climate zone, the parallel marking its southern confine, its center parallel and the parallel marking its northern confine are given. Sometimes the latter is omitted as it coincides with the beginning of the next climate zone. Usually, the parallels are given as rows, but there are other layout options working for instance with additional columns (Figure \ref{fig:seven_sample_b}). As mentioned above, the seven climate-zone table represents the main scientific tradition since antiquity and throughout the Western Middle Ages and the Islamicate culture.\par

\begin{figure*}
    \centering
    \includegraphics[width=0.7\textwidth]{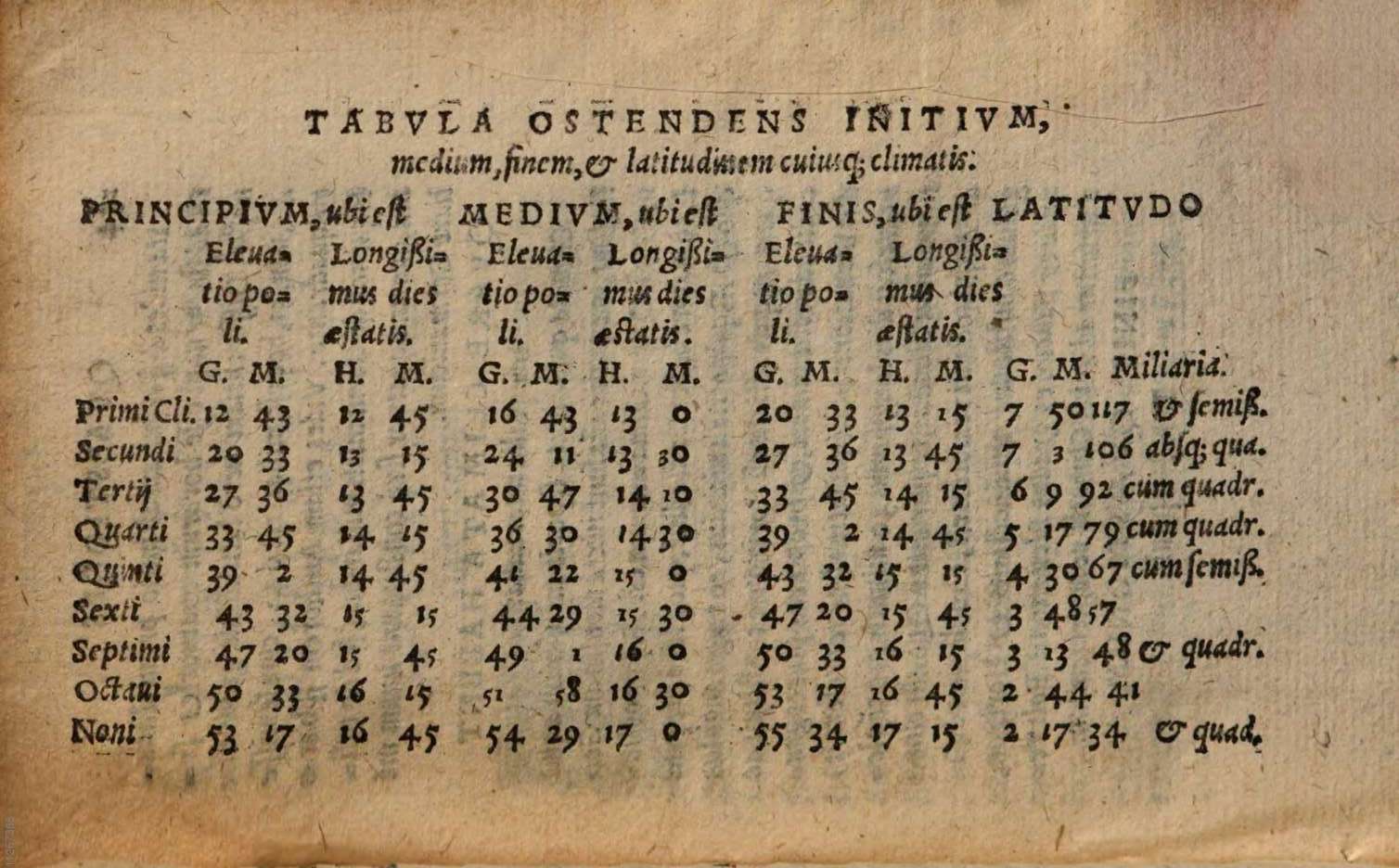}
    \caption{\textbf{Nine climes table.} \cite[283]{Beyer1560}. Augsburg, Staats- und Stadtbibliothek, urn:nbn:de:bvb:12-bsb11267368-7.}
    \label{fig:nine_sample_a}
\end{figure*}

\subparagraph{\textbf{Nine climate zones:}} These tables extend the type `seven' tables by two additional climate zones toward the north. They re-occur at a much lower rate, twenty-six instances in total in the Sacrobosco Collection, and the layout of these examples is much more homogeneous than in the case of the seven climate-zone tables. Figure \ref{fig:nine_sample_a} shows the typical layout: Initial medium and end parallel are defined by day-length and longitude and these pieces of information are listed in columns. Angle and arc length of the sectors for each climate zone are specified, no locations are usually listed. The addition of two climate zones toward north represents a rupture with the tradition. The reason for such change, however, was the evident fact that those zones were now inhabited at least as much as the seven defined by the tradition.\par

\begin{figure}[ht]
\centering
  \begin{subfigure}[b]{0.4\linewidth}
    \includegraphics[width=1\textwidth]{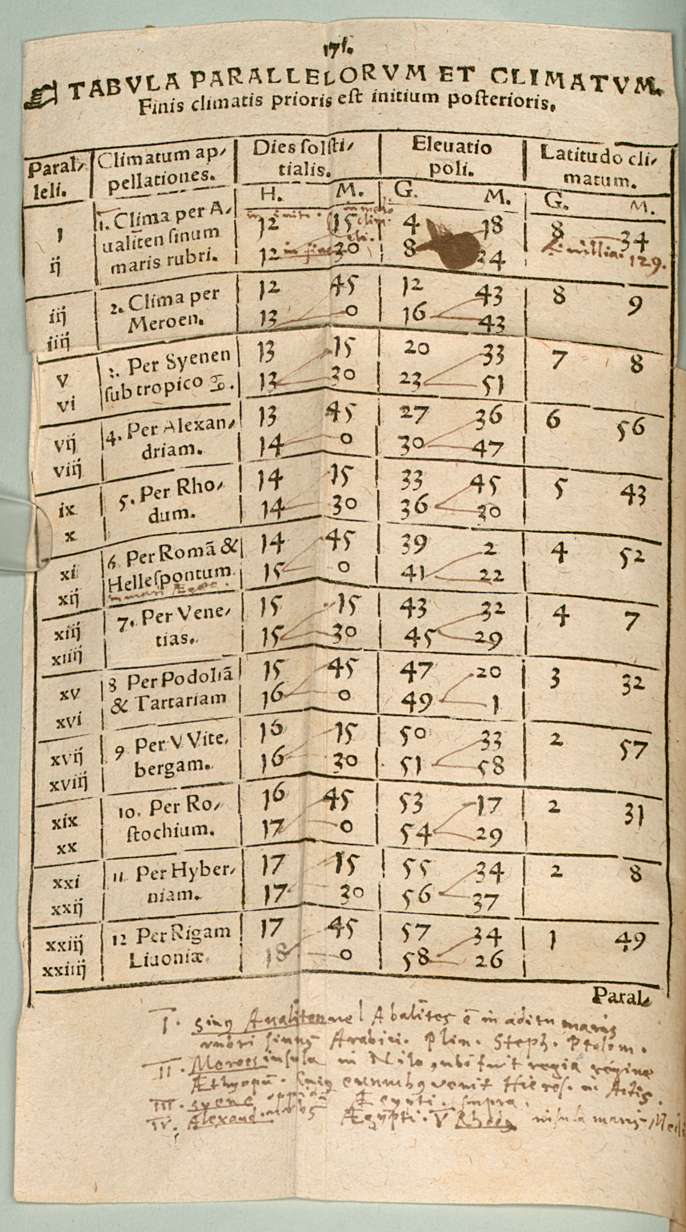}
    \caption{Left page}
    \label{fig:twentyfour_sample_a1}
  \end{subfigure}
  \hspace{0.1cm}
  \begin{subfigure}[b]{0.4\linewidth}
    \includegraphics[width=1\textwidth]{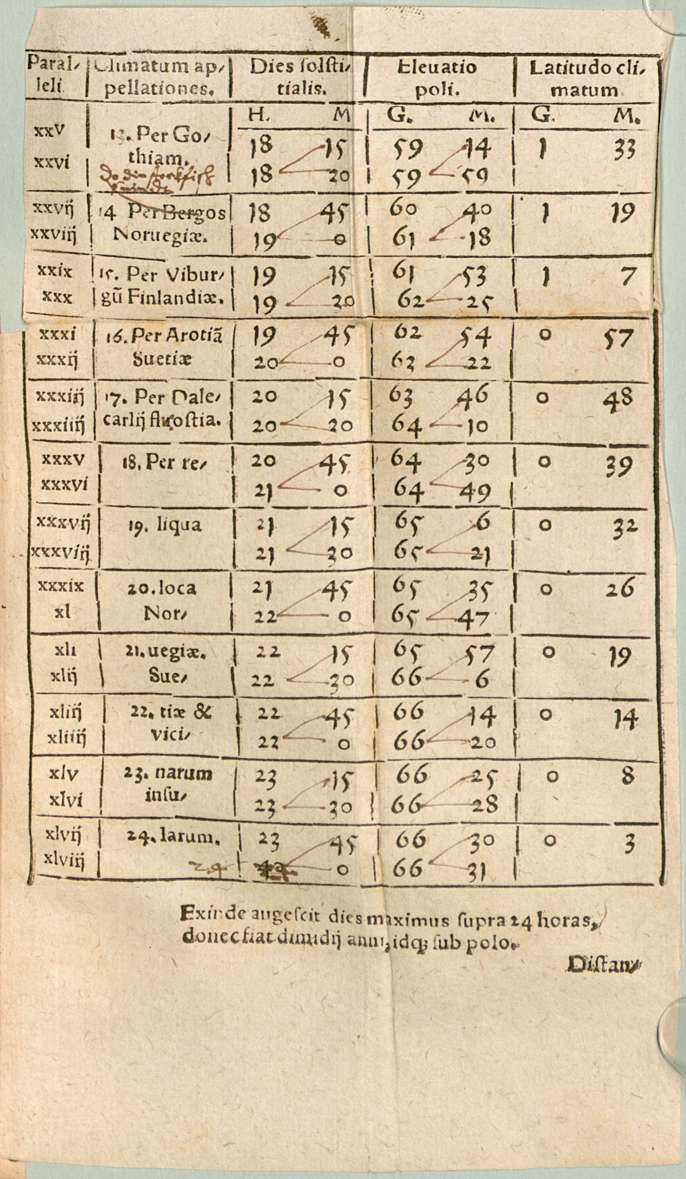}
    \caption{Right page}
    \label{fig:twentyfour_sample_a2}
  \end{subfigure}
  \caption{\textbf{Twenty-four climes table.} \cite[171r–171v]{Witekind1590}. München, Bayerische Staatsbibliothek, urn:nbn:de:bvb:12-bsb00021009-2.}
\end{figure}

\begin{figure}
\centering

\begin{subfigure}[t]{0.32\textwidth}
    \centering
    \includegraphics[width=\textwidth]{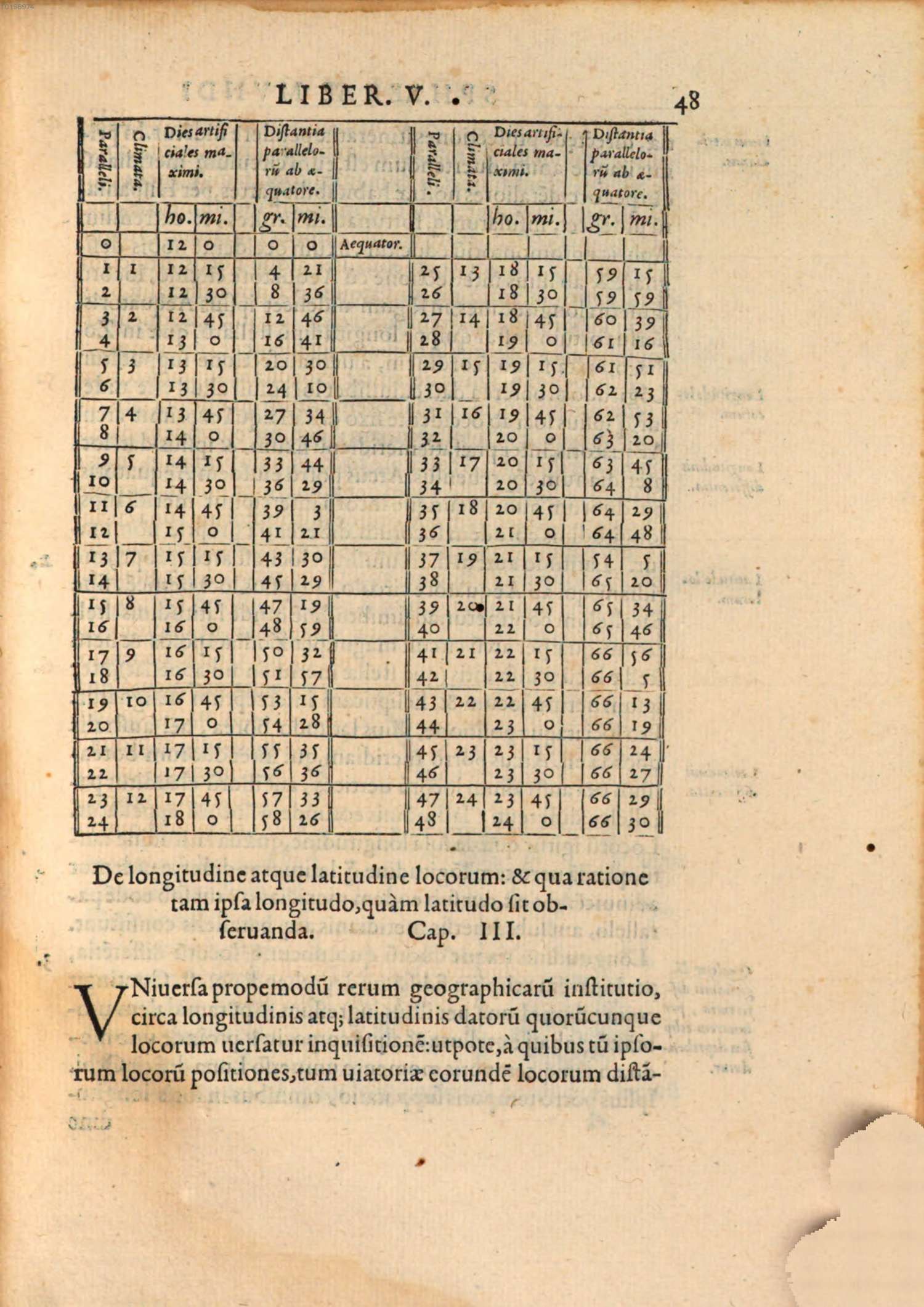}
    \caption{\textbf{Twenty-four climes table.} \cite[48r]{FineLat1555}. München, Bayerische Staatsbibliothek, urn:nbn:de:bvb:12-bsb10198974-9.}
    \label{fig:twentyfour_sample_b}
\end{subfigure}
\hfill
\begin{subfigure}[t]{0.32\textwidth}
    \centering
    \includegraphics[width=\textwidth, trim=5 30 0 5, clip]{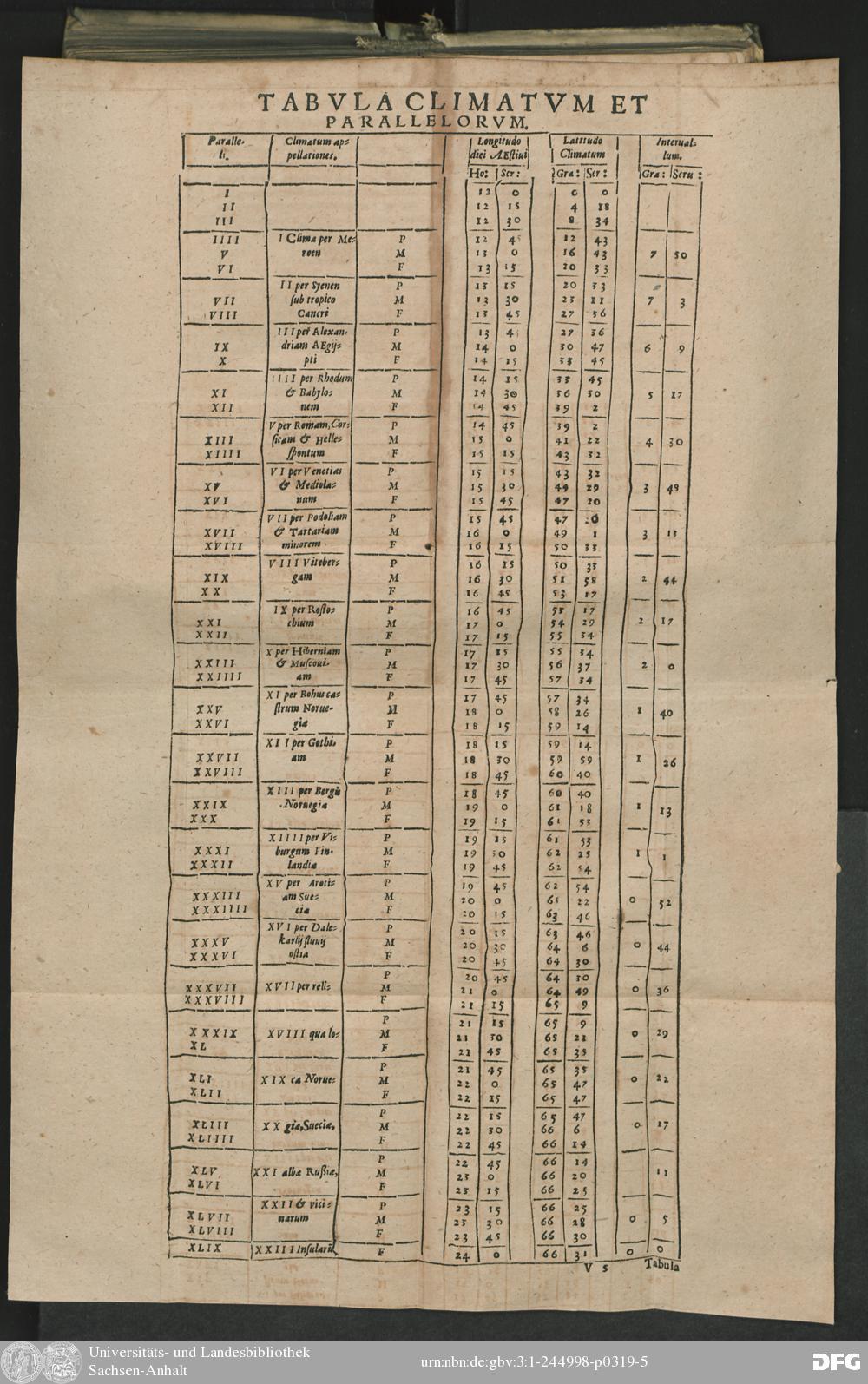}
    \caption{\textbf{Twenty-four climes table.} \cite[296]{Dietrich1591}. Universitäts- und Landesbibliothek Sachsen-Anhalt, \protect\href{http://dx.doi.org/10.25673/opendata2-7144}{http://dx.doi.org/10.25673/opendata2-7144}.}
    \label{fig:twentyfour_sample_c}
\end{subfigure}
\hfill
\begin{subfigure}[t]{0.32\textwidth}
    \centering
    \includegraphics[width=\textwidth, trim=35 40 5 70, clip]{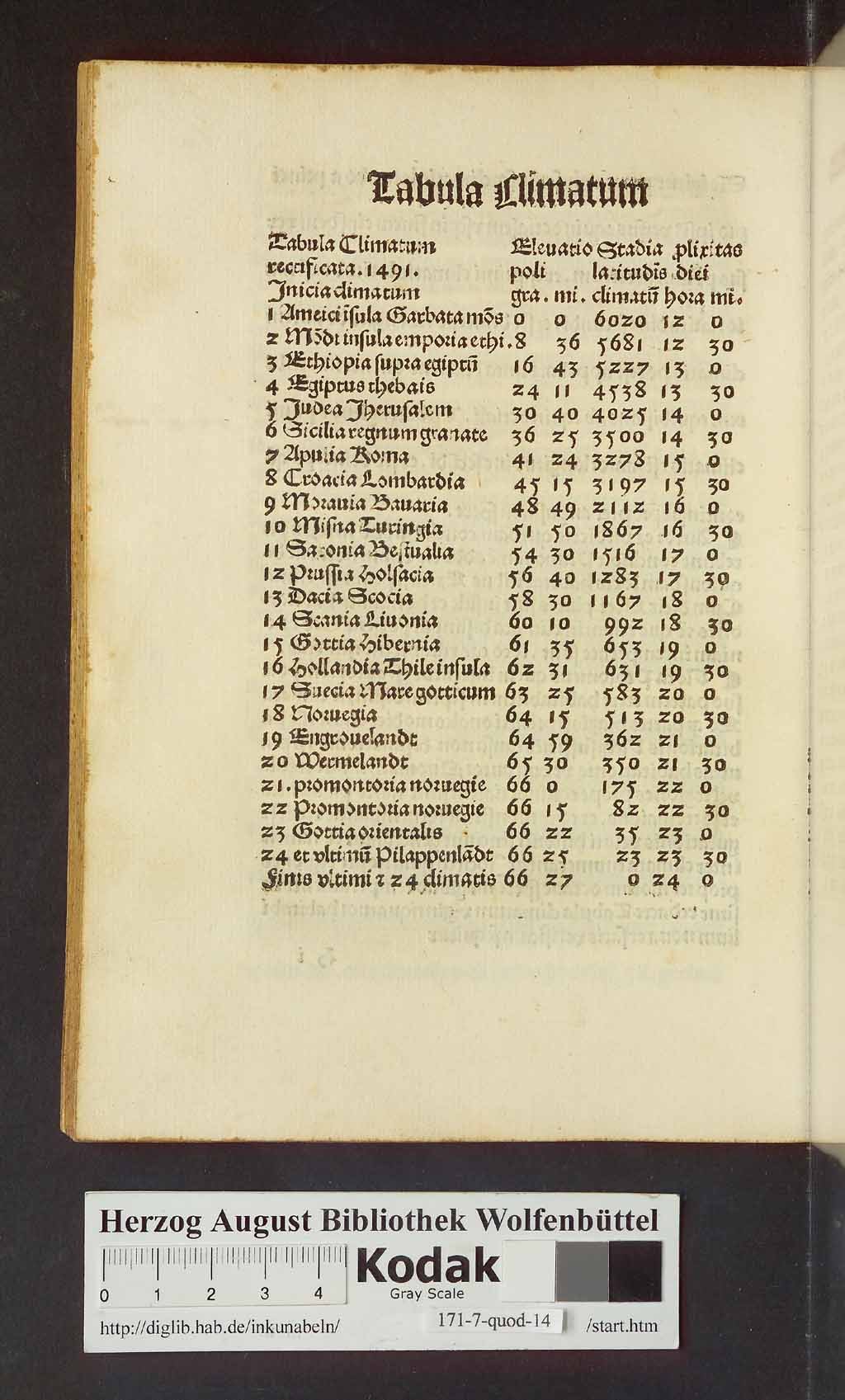}
    \caption{\textbf{Twenty-four climes table.} \cite[Hi-1]{Sacrobosco1495}. Herzog August Bibliothek Wolfenbüttel, \protect\href{http://diglib.hab.de/inkunabeln/171-7-quod-14/start.htm}{http://diglib.hab.de/inkunabeln/171-7-quod-14/start.htm}.}
    \label{fig:twentyfour_sample_d}
\end{subfigure}
\end{figure}

\subparagraph{\textbf{Twenty-four climate zones:}} These tables extend the schema of the seven and nine climate zones respectively even further north usually up to the polar circle where the length of the longest day is exactly 24 hours, resulting in either 24 or 23 (and sometimes even less) climate zones depending on how the extension is carried out concretely. Even though the actual number of climate zones listed in these tables can vary, they are here  subsumed under the rubric twenty-four climate zones. The Sacrobosco collection contains eighty-one such tables. They usually (but not always) stretch over more than one page. Their contents and layouts are even more variable than in the two previous cases. There is in particular a variability with respect to where the tables start in the south and how they count the parallels from there. In the traditional scheme the first climate zone has its southern confine where the longest day has 12 hours and 45, corresponding to a parallel at 12 degrees 45 latitude and this was counted as the first parallel. Of course this schema of defining parallels by day-length in hours of the longest day in increments of 15 minutes can be extended further south to the equator adding three more (including the the equator itself) parallels. This extension to south is indeed made in most of the tables in this group. It is however done in different manners resulting in different counts of the parallels (Figure \ref{fig:twentyfour_sample_a1} and \ref{fig:twentyfour_sample_a2}, \ref{fig:twentyfour_sample_b}, and \ref{fig:twentyfour_sample_c}). Sometimes the first zone is supposed to start at the earth's equator with the longest day (like every day there) measuring 12 hours, but sometimes also where the longest day measures 12 hours and 15 minutes. In both cases 24 climate zones result. As retained from the tradition, however,  the first zone often starts at day-length 12 hours and 45 minutes, in which case there are only 23 zones. Figure \ref{fig:twentyfour_sample_d} provides an example of an extreme variation. First it only gives the southern confines of the zones listed. Moreover, it starts the first climate zone at the equator and thus the zones listed do not correspond to the traditional ones (neither in their numbering nor in latitude of their confines). Among others, this results in an actual number of twenty-five climate zones, though we still maintain that it generally belongs to the group of tables showing twenty-four zones.\par

\begin{figure}[h!]
  \centering
  \begin{subfigure}[b]{0.4\linewidth}
    \includegraphics[width=\textwidth]{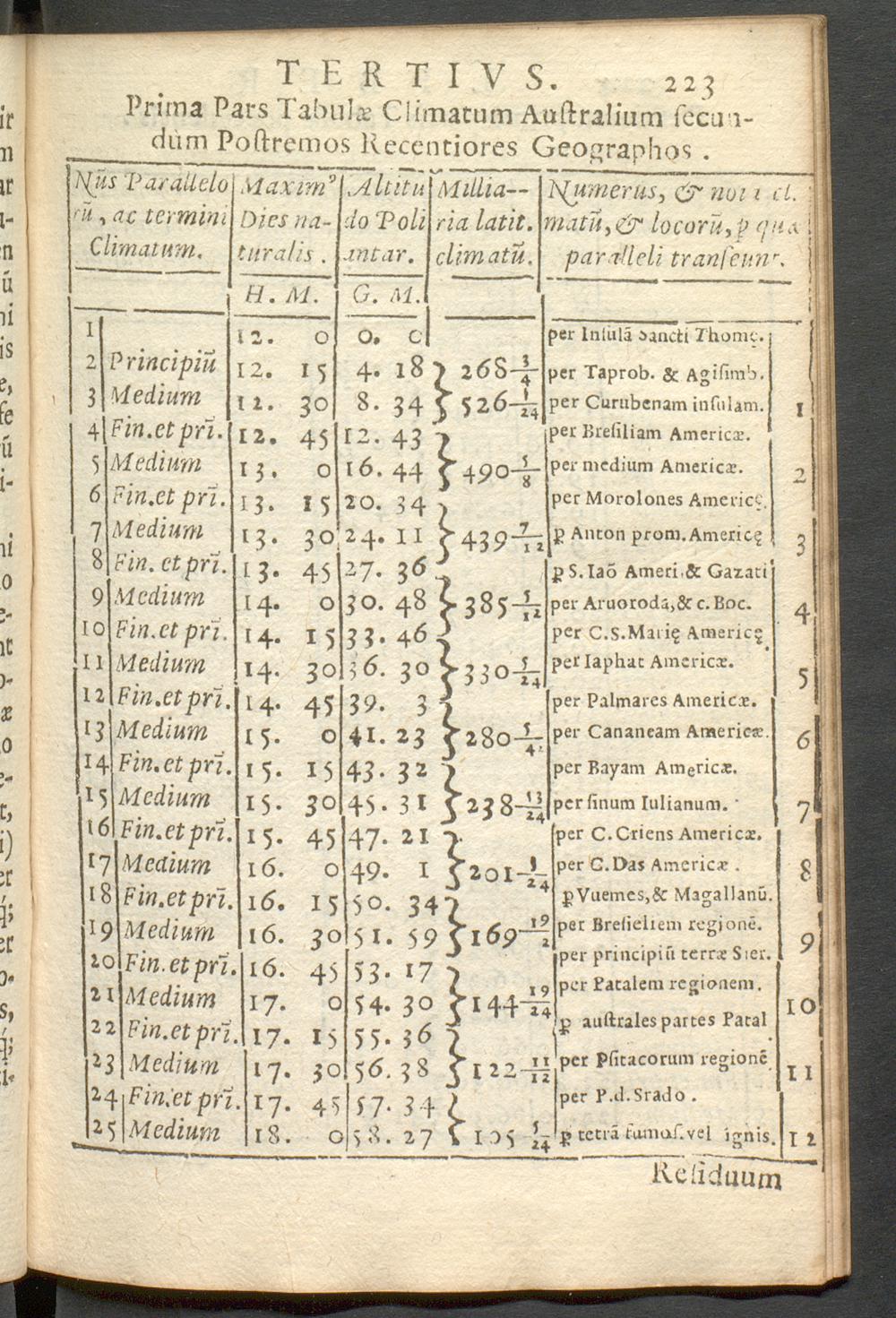}
  \end{subfigure}
  \hspace{0.1cm}
  \begin{subfigure}[b]{0.4\linewidth}
    \includegraphics[width=\textwidth]{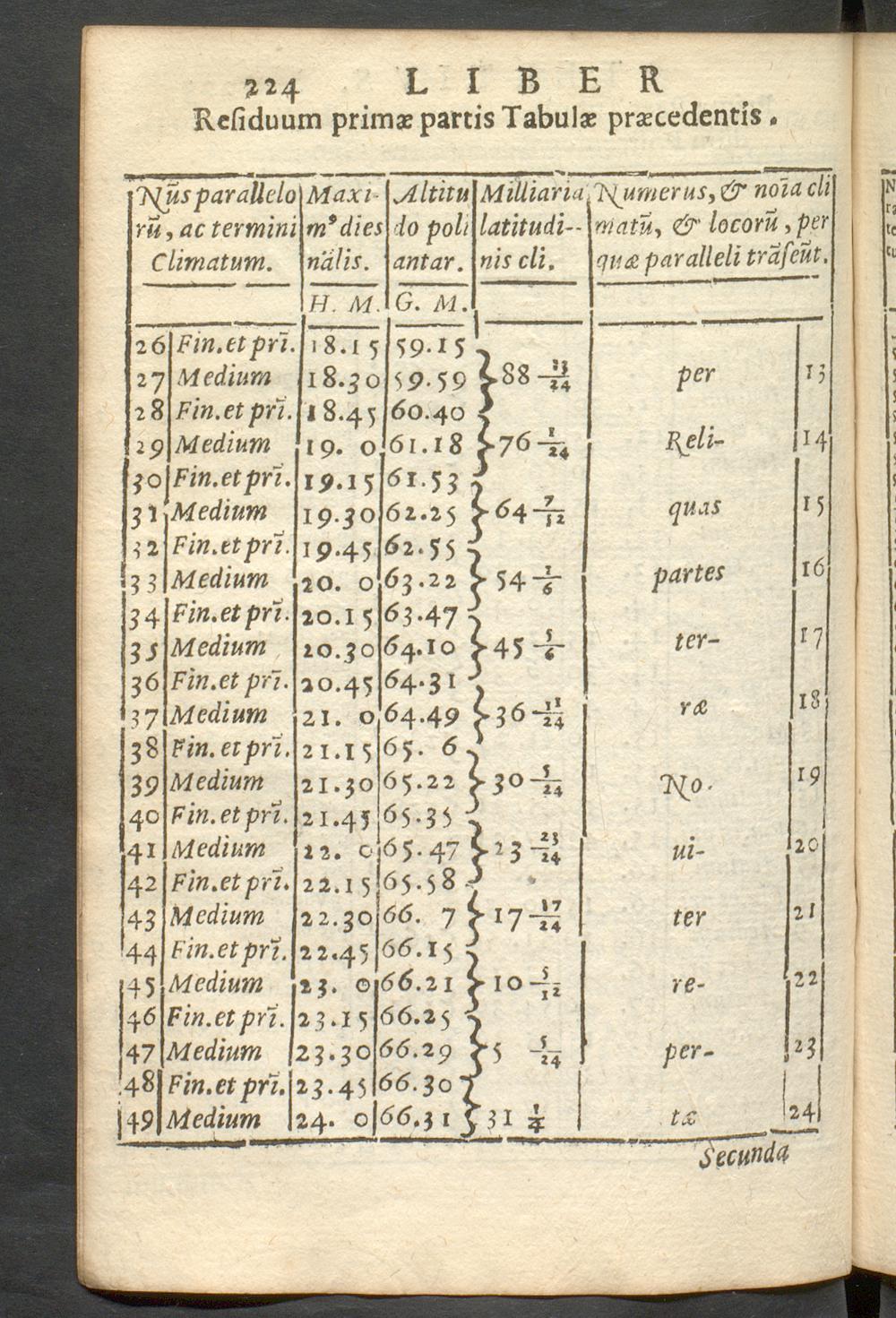}
  \end{subfigure}
  \caption{\textbf{Twenty-four climate  table  for the southern hemisphere.} \cite[223–224]{Barozzi1598}. ETH-Bibliothek Zürich, \protect\href{https://doi.org/10.3931/e-rara-17609}{https://doi.org/10.3931/e-rara-17609}.}
  \label{fig:clime_south_hemi}
\end{figure}

From a historical point of view, the appearance and diffusion of the twenty-four zones table can be considered as the consequence of the recognition that the entire globe, as it was becoming known through the journeys of exploration, was actually inhabited. This interpretation is supported by the fact that we indeed find one instance where the schema of the climate zones is applied to the southern hemisphere as testified by a table listing climate zones south of the equator (see Figure \ref{fig:clime_south_hemi}). \par

\paragraph{The Spread of Climate Zone Tables} \label{supp:link:climate_zones}
By means of our model, we were able to identify a group of 224 tables (out of about 10,000) that display data related to the climate zones and that can be distinguished into three sub-groups of tables as described above. The closer analysis of this hitherto unexplored historical material, furthermore, allowed us to formulate the hypothesis that the departure from the tradition, represented by the tables displaying nine and twenty-four climate zones respectively, is due to the increasing recognition that the concept of inhabitable portion of the Earth surface faced by the discoveries resulting from the journeys of exploration was loosing its scientific meaning. To try to determine the validity or at least the plausibility of this hypothesis, we  analyze the spatio-temporal distribution of these tables,  by matching the identified tables with the bibliographic metadata of the editions in which they were printed.\par

\begin{figure*}[h!]
    \centering
    \includegraphics[width=0.95\textwidth]{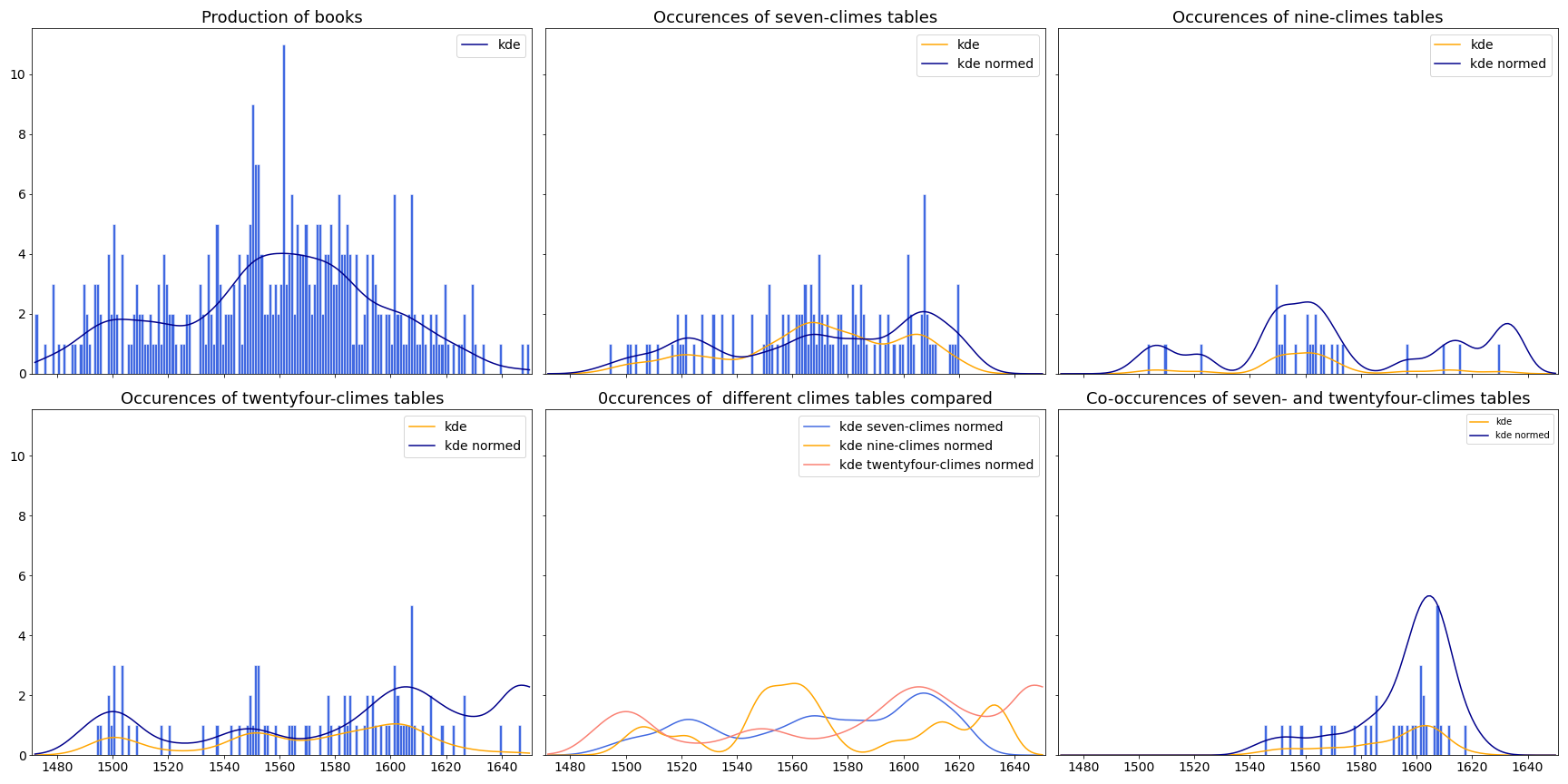}
    \caption{\textbf{Temporal spread of print production.} Analysis of the publication patterns of the different types of climate zone tables contained in the corpus.}
    \label{fig:clime_table_analysis}
\end{figure*}

First of all, we look at the temporal distribution of the three types of tables. To do so, we smooth the temporal distribution of the publication events as already provided in Figure \ref{fig:publication_rate} by a Kernel Density Estimation (KDE)\footnote{\href{https://docs.scipy.org/doc/scipy/reference/generated/scipy.stats.gaussian_kde.html\#scipy.stats.gaussian_kde}{https://docs.scipy.org/doc/scipy/reference/generated/scipy.stats.gaussian\_kde.html\#scipy.stats.gaussian\_kde}}, which provides an estimate of the probability of a book being interpreted as an averaged rate of appearance at any given time in the period considered (Figure \ref{fig:clime_table_analysis}-a.). We then single out the editions in which at least one table belonging for one of the three principal types of climate tables we have identified was published. We do this for all tree types and likewise calculate the KDEs for the three temporal distribution of the publication events of the editions thus singled out. In addition we  normalize the rate of appearance of the tables in the editions with regard to the KDE representing the rate of appearance of all the editions constituting the corpus and shown above (Fig. \ref{fig:clime_table_analysis}-b, c, d.) and directly compare the appearance rate of the three different types of tables (Fig. \ref{fig:clime_table_analysis} e.). All KDEs have been calculated using a bandwidth of 0.2.\par

Contrary to what one could expect at this point, the tradition, represented by the data concerned with the seven climate zones, remains rather robust for the entire period considered, which ends 158 years after the discovery of the New World. As expected, however, the rupture with the tradition represented by the nine climate-zones table spread about 40 years before the option of the twenty-four zones received same degree of of attention in print (Fig. \ref{fig:clime_table_analysis}-e.). This chronology of events supports the historical hypothesis that the emergence of new types of climate zone tables was associated with the increasing knowledge of the Earth's surface.\par

Moreover, if we directly compare the metadata for the editions containing more than one of the three types of tables, we discover a further interesting aspect. Firstly, many editions that contain the tables displaying the twenty-four climate zones also contain the traditional seven-zone table and secondly the peak in the production of the twenty-four zones table is due to exactly those editions that contain both types of tables (Fig. \ref{fig:clime_table_analysis} f.).\par

Considering the more abstract subject of the mode of production of scientific knowledge during the early modern period, finally, the present case study allows us to assert that scientific innovations as represented by some of the computational tables in our corpus were introduced and could become successful mainly by building upon traditional and well-accepted knowledge. This same pattern is also observed in the case of the textual apparatus of the treatises where new knowledge is often presented in form of a commentary on a old and authoritative text.\par

\begin{figure}[h!]
    \begin{subfigure}[b]{0.3\textwidth}
    \centering
    \includegraphics[width=\textwidth]{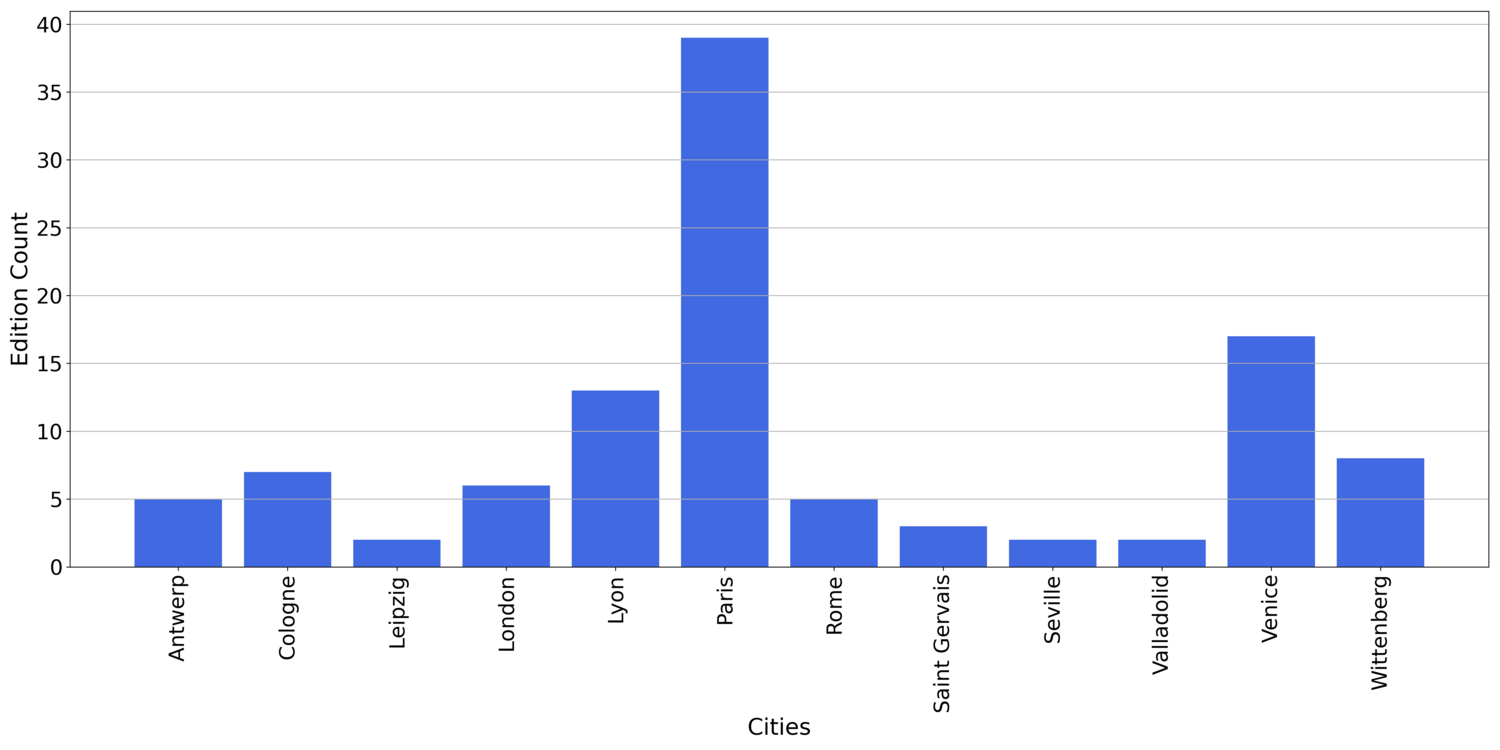}
    \caption{Seven climate zones}
    \label{fig:Cities_Seven}
    \end{subfigure}
    \hspace{0.1cm}
    \begin{subfigure}[b]{0.3\textwidth}
    \centering
    \includegraphics[width=\textwidth]{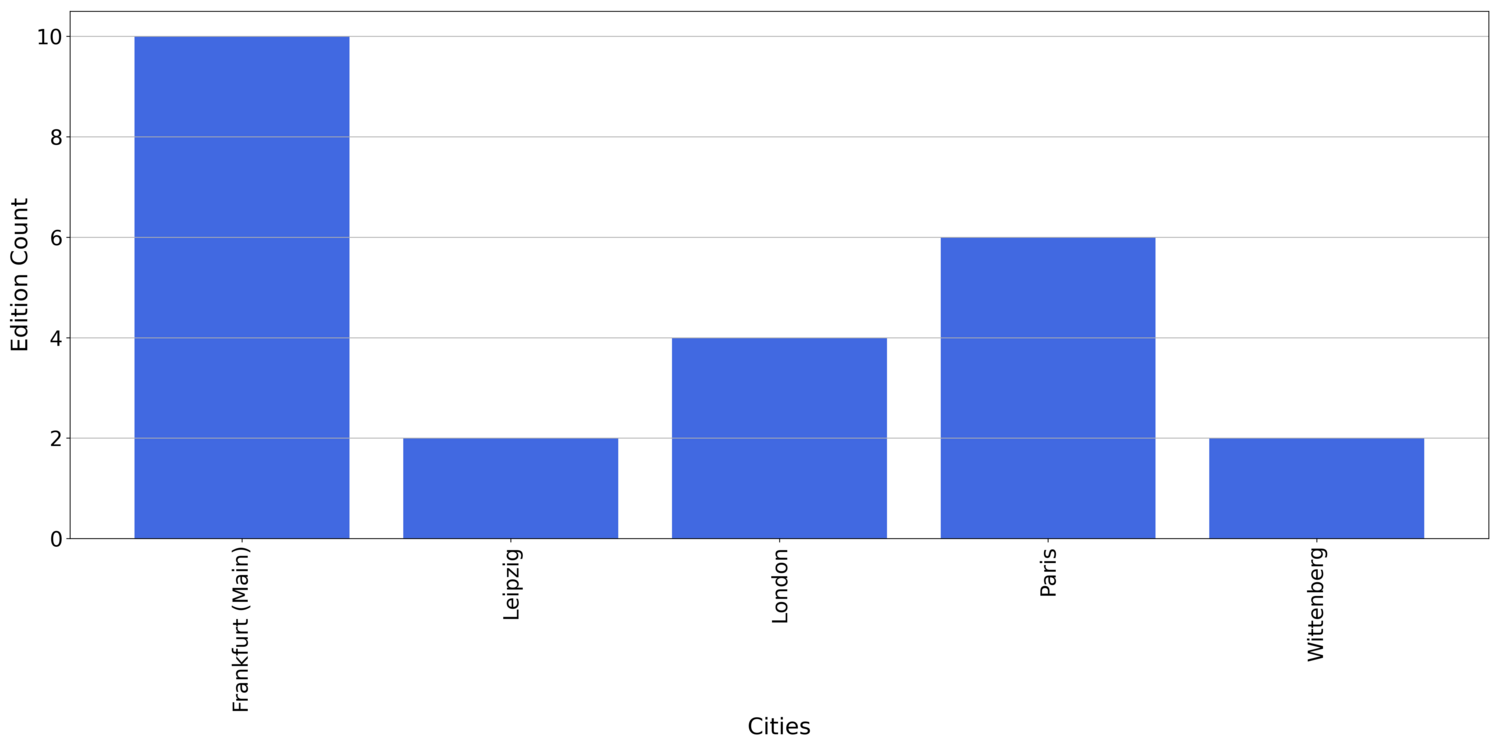}
    \caption{Nine climate zones}
    \label{fig:Cities_Nine}
    \end{subfigure}
    \hspace{0.1cm}
     \begin{subfigure}[b]{0.3\textwidth}
    \centering
    \includegraphics[width=\textwidth]{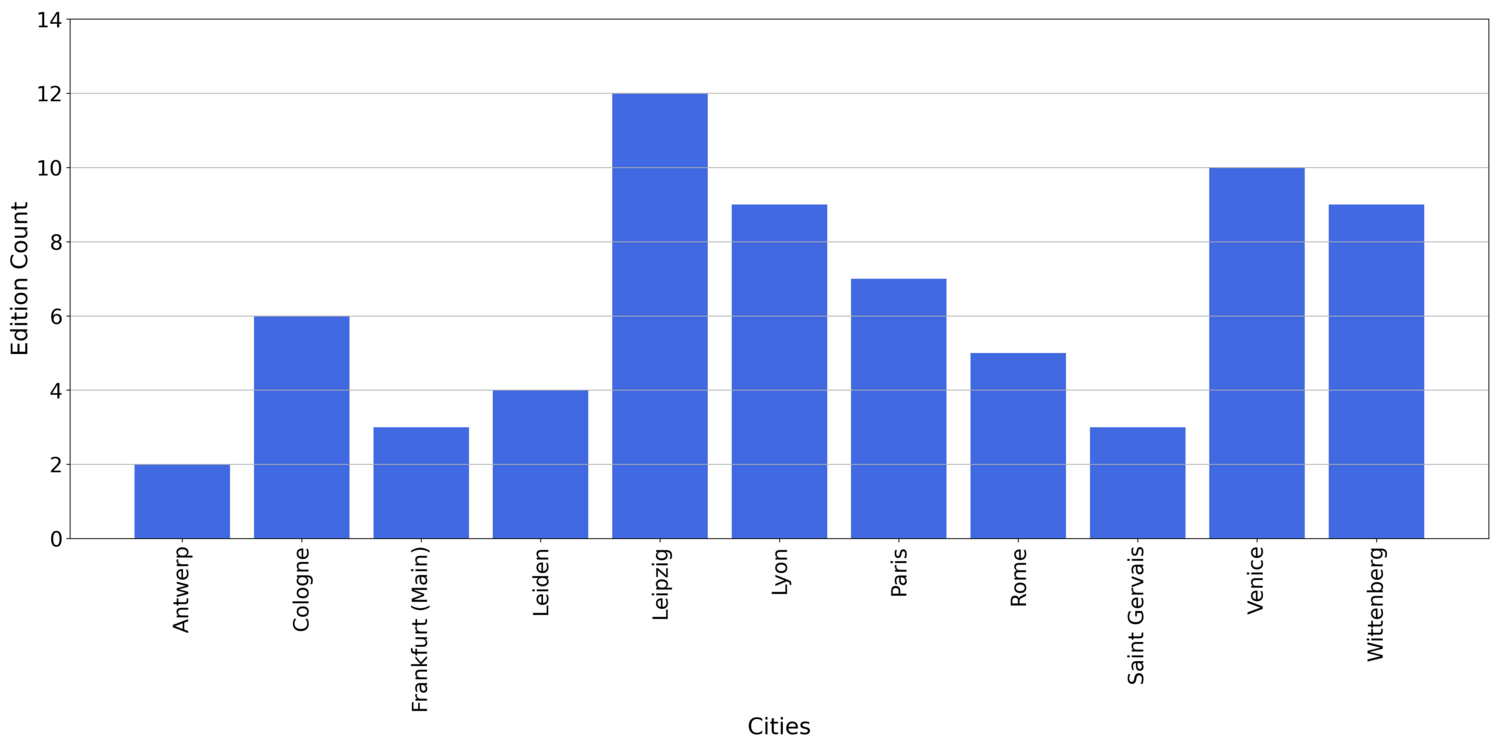}
    \caption{Twenty-four climate zones}
    \label{fig:Cities_twenty-four}
    \end{subfigure}
    
\caption{\textbf{Major centers of production for climate zone tables.}}
\label{fig:Cities_climate_zones}
\end{figure}

\begin{figure}[h]
    \centering
    \includegraphics[width=\textwidth]{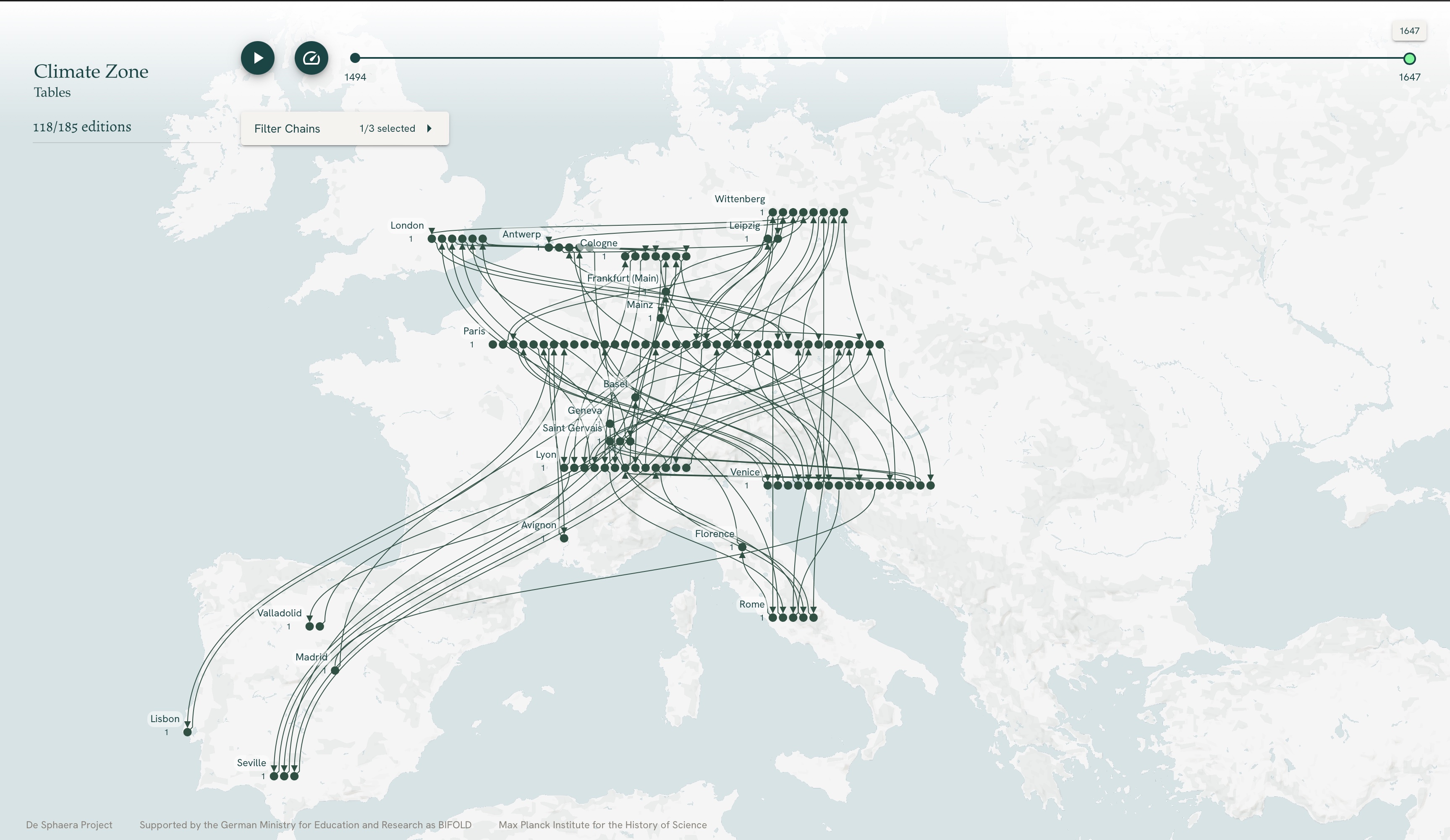}
    \caption{\textbf{Spread} of the table displaying the seven climate zones}
    \label{fig:Spread_Seven}
\end{figure}

\begin{figure}[h]
    \centering
    \includegraphics[width=\textwidth]{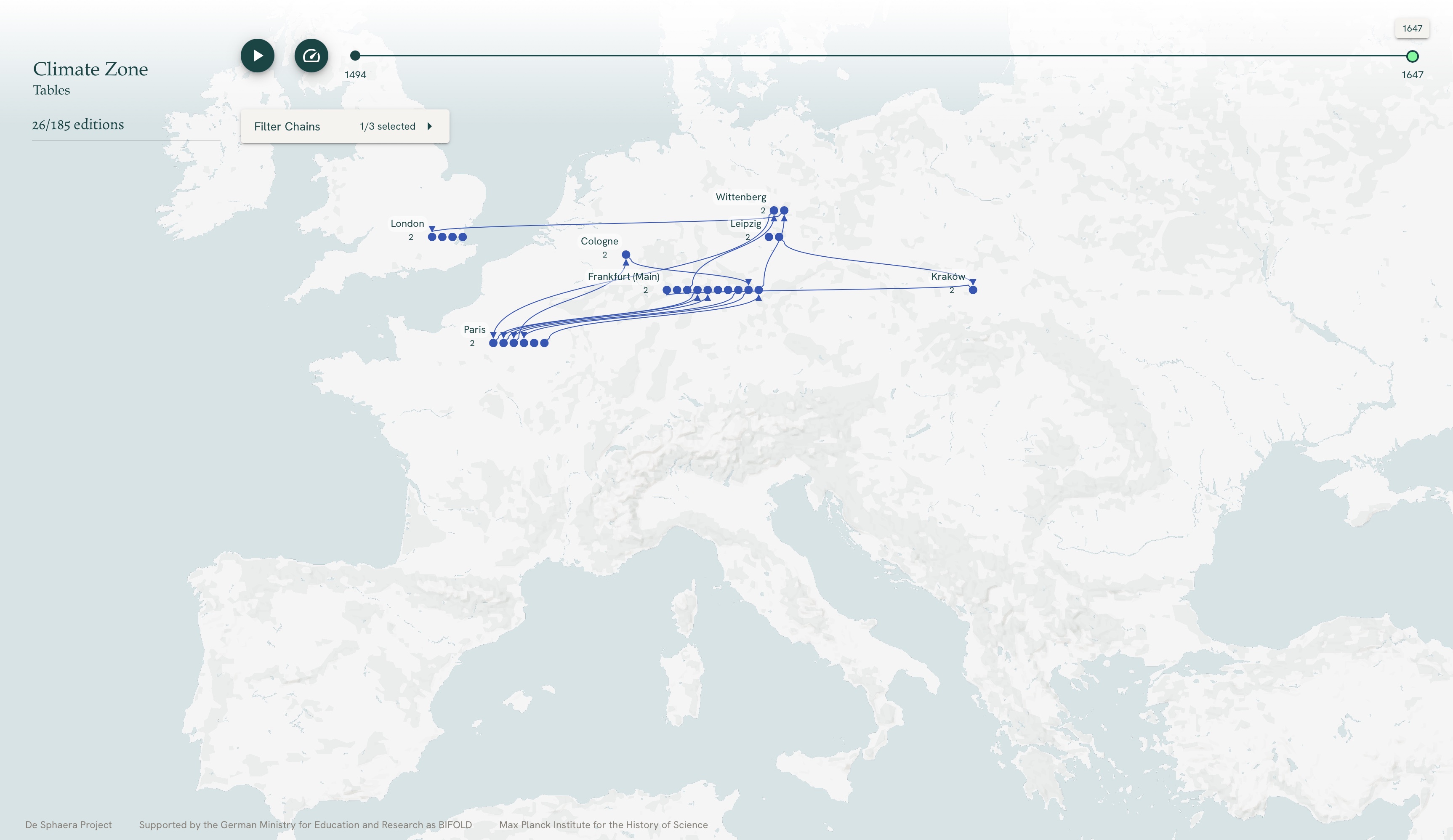}
    \caption{\textbf{Spread} of the table displaying the nine climate zones}
    \label{fig:Spread_Nine}
\end{figure}

\begin{figure}[h]
    \centering
    \includegraphics[width=\textwidth]{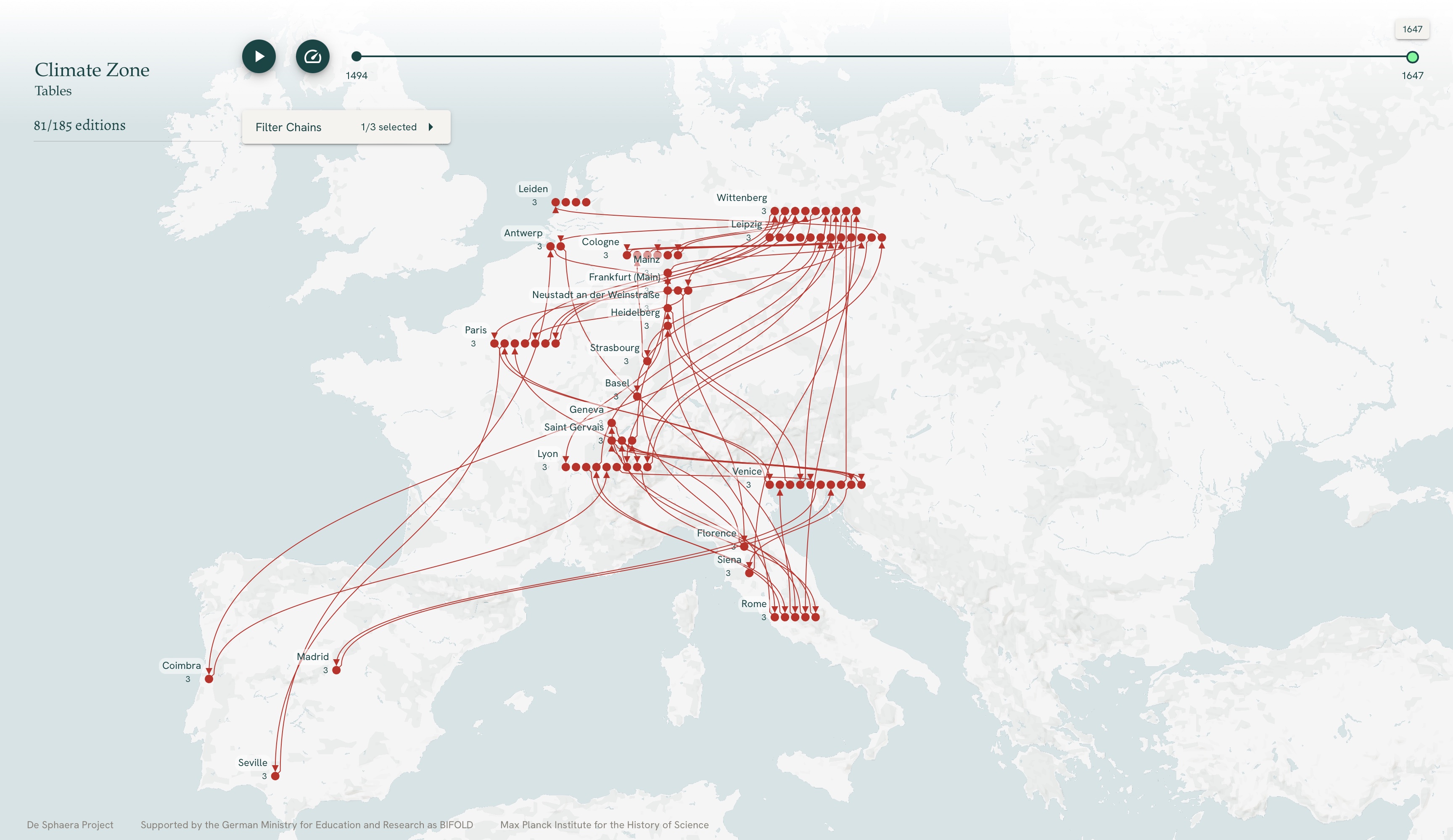}
    \caption{\textbf{Spread} of the table displaying the twenty-four climate zones}
    \label{fig:Spread_twenty-four}
\end{figure}

To look at the temporal and spatial dynamic of the geographic spread of the tables, we created and visualized chains of re-occurrences. Each occurrence of a climate table of a particular type is linked to the first occurrence of a table of the same type that came before it. This simulation enables us to visualize the process of diffusion of this specific type knowledge and calculate both the instant and average speed of diffusion.\par

The interactive visualization showing the evolution of the different table variants covering the different climate zones is available at 
\href{http://141.5.103.115/climes}{http://141.5.103.115/climes}\protect\footnote{\texttt{username: network | password: VIZ\_network\$\_61t50}}.
The dynamic of the spread of the three different tables can be visualized individually or together.

\begin{itemize}
    \item[--]  Chain 1:   Geo-temporal spread of the table displaying seven climate zones.
    
    \item[--]  Chain 2:  Geo-temporal spread of the table displaying nine climate zones.
    \item[--]  Chain 3:  Geo-temporal spread of the table displaying twenty-four  climate zones.
\end{itemize}

To display the metadata of the editions, each single edition can be selected on the visualization. This selection also displays the figures concerned with the speed of spread of that specific table selected.

Examining the first appearance of the tables, all three sorts of tables appear quite early in the collection for the first time, namely either before or just after the turn of the fifteenth century. Both the seven and the nine-zones tables disappear around 1620 while the twenty-four zones table re-occur until the very end of the period considered. Observing the spatial distribution of occurrences, further relevant aspects can be recognized, too (Figs. \ref{fig:Spread_Seven}, \ref{fig:Spread_Nine}, \ref{fig:Spread_twenty-four}, \ref{fig:Cities_Seven}, \ref{fig:Cities_Nine}, and \ref{fig:Cities_twenty-four})\footnote{To improve visibility, centers of production in which only one occurrence of the respective table took place have been suppressed from the plots.}. The table displaying seven zones is printed in nineteen different cities. In this case the major centers of production and spread clearly are Paris, Venice, Lyon, and Wittenberg. In the case of the table containing data for nine climate zones, this spread only covers seven different cities, and the major centers are Frankfurt am Main, London, and Paris. In the case of the twenty-four climate zones, finally, they were printed in twenty-two different cities, in spite of a lower number of occurrences. Their major centers of production were Leipzig, Venice, and Wittenberg.\par

In particular, we observe first that the geographic diffusion of the tables for both, the seven and the twenty-four climate zones is very similar, though not identical. Both chains of re-occurrences move from northern Europe toward the Iberian Peninsula and embrace northern Italy and South France. Toward East, they do not go beyond Wittenberg. In the case of the nine-zones table, however, the dynamic is very different as it fundamentally remains a northern European phenomenon, which, however, reaches as far east as the city of Krakow.\par

\subsubsubsection[]{Historical Case 2: Tables of Zodiac Signs} \label{supplement:text:case_study_zodiac}

As second historical case, we selected all printed instances of a particular table which gives the positions of the Sun into the signs of the zodiac in degrees for each day of the year. This table  occurs in varying layouts in our corpus, where the different layouts partition the full table differently. In some cases the entire table is comprised on one page, in other books it is distributed over as many as nine pages. \par

Due to the precession of the earth's axis, the positions of the Sun for a given point of time in the solar year (for instance the vernal equinox) gradually changes over time against the ecliptic. More importantly, before the calendar reform of 1582, the solar year was not in tune with the calendar year, such that the calendar date, for instance of the equinoxes, changed over time. In effect the position of the Sun against the ecliptic for a given calendar day depends on the year for which it is calculated.\par

Thanks to our model, we were able to investigate the diffusion of such table in the about 180 years considered here. First, we immediately recognized that, in our corpus,  two variants of the sun-zodiac table are present: first, tables designated as valid for the times of the `ancient' poets (\textit{veterum poetarum temporibus accommodata}) where the Sun is 16 degrees into Capricorn on the first of January and, second, tables valid for `contemporary' times (\textit{nostro tempori}) were the Sun, on the first day of the year, has advanced 5 degrees and is located 21 degrees into Capricorn. Historically, the contemporary table was inserted in the treatises to teach the students the correspondence between the calendar and the celestial phenomena. Such correspondence, then, could be used to date past events, should the observation concerning the position of the Sun have been made explicit in the sources. The table for the ancients is in fact called ``the table for the ancient poets'' because it refers to texts of classical literature of authors, such as Hesiod, Ovid, or Pliny.\par

In antiquity the correspondence between this celestial phenomenon and the calendar was common knowledge and, when they described particular events, those ancient authors rarely missed the chance to signalize the position of the Sun in the Zodiac, so that other and later readers could reconstruct the date of the event. This subject — the reconstruction of the temporal order of historical events described in ancient sources—became very important during the sixteenth century, especially in the cultural circles of the Protestant reformation. Initially, students could use the contemporary table and, through further calendric computations as well as astronomic calculations concerned with the precession of the equinoxes, they could date events described by the ancient authors. This method eventually revealed too complex and therefore lecturers of Wittenberg introduced a new table already adjusted for the ancient time. In fact this table is almost always directly accompanied by a another table specifically referring to and valid for Alexandria and Rome, the two places to which the ancient authors also commonly referred in their observations and descriptions. This table lists for the most prominent stars  the degree of the zodiac rising and setting respectively together with the corresponding star as observed from either of the two locations. Together with the sun-zodiac table one can thus effortlessly determine the cosmical risings and settings of the stars for these two locations (and thus eventually also the helical and acronical risings and settings), another type of data often given in the ancient literature.   \footnote{For the historical reasons of the introduction of a new computational table concerned only with the ancient classical time, see  \cite{MVBFON2022}. For examples on how past events were dated, see \cite{Pantin2021OP}.}\par 

\begin{figure}[h!]
\centering
  \begin{subfigure}[b]{0.4\linewidth}
    \includegraphics[width=1\linewidth, trim={0 20 0 0}, clip]{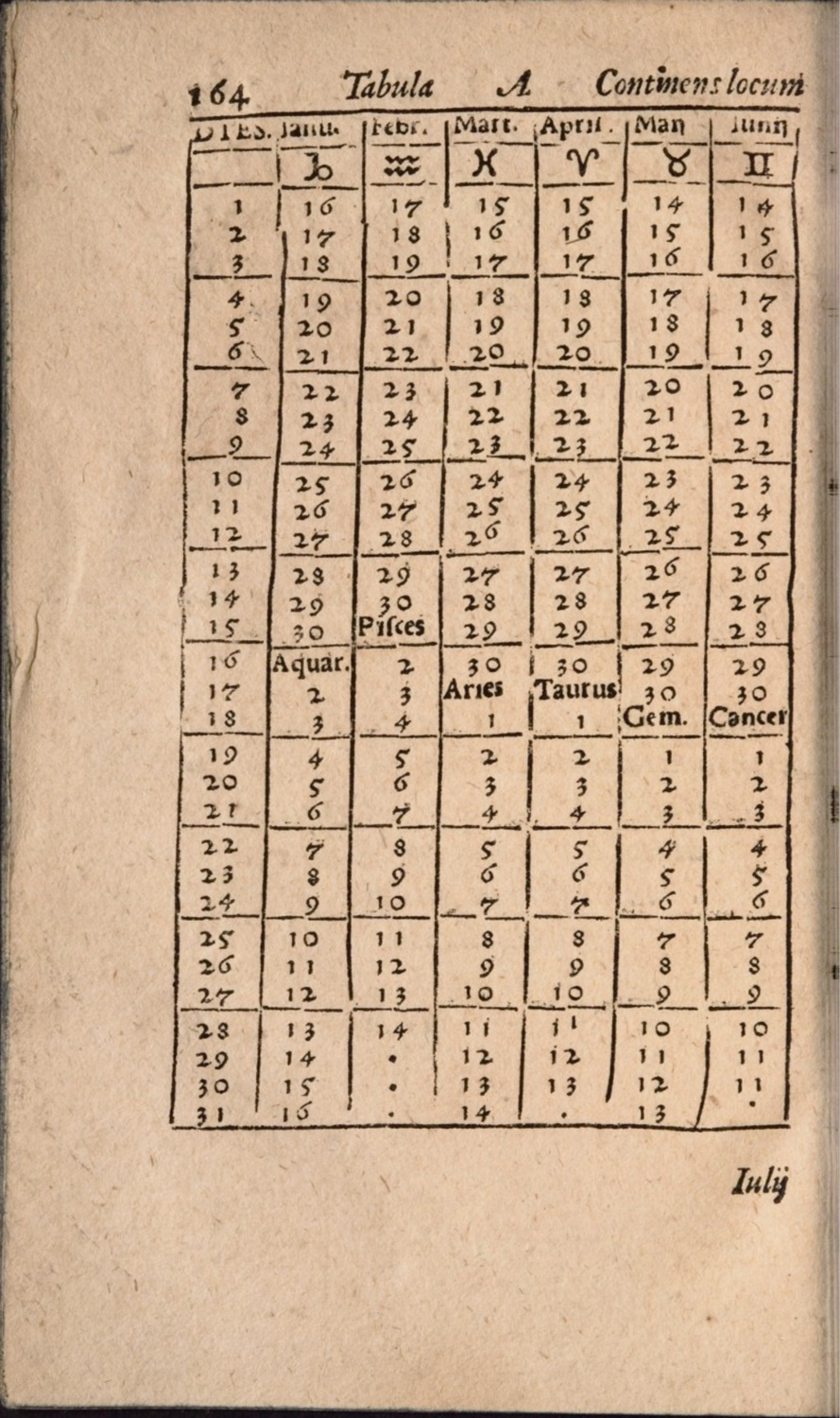}
    \caption{Left page}
    \label{fig:Sun_Zodiac_Nostro_Blebel_1582_01}
  \end{subfigure}
  \hspace{0.1cm}
  \begin{subfigure}[b]{0.4\linewidth}
    \includegraphics[width=1\linewidth, trim={0 20 0 0}, clip]{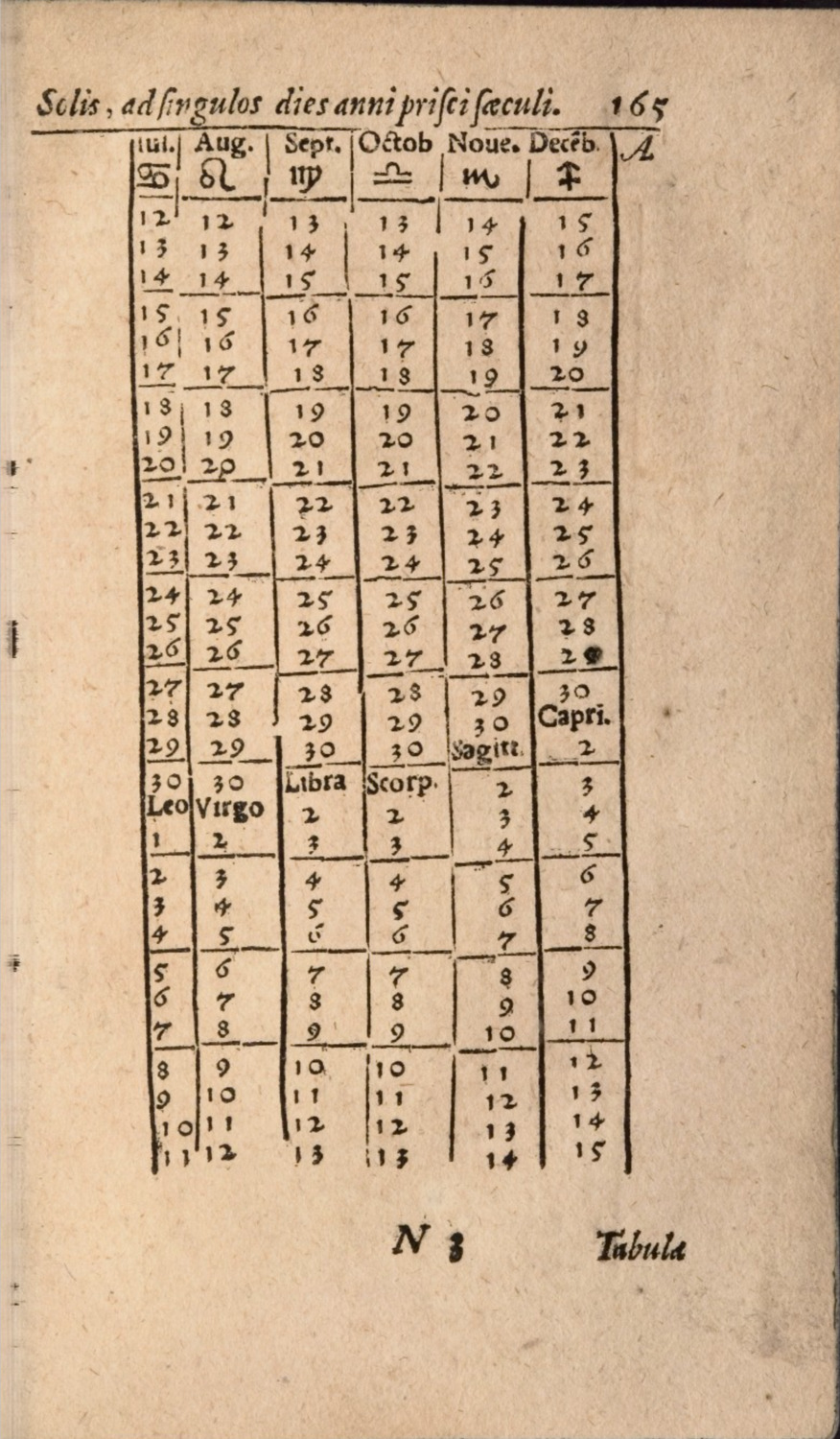}
    \caption{Right page}
    \label{fig:Sun_Zodiac_Nostro_Blebel_1582_02}
  \end{subfigure}
  \caption{\textbf{Sun-Zodiac table} computed in relation of the temporal position of the equinoxes in the contemporary time (Edition published in 1582). From \cite[164–165]{Blebel1582}. Courtesy of the Library of the Max Planck Institute for the History of Science.}
\end{figure}

\begin{figure}[h!]
\centering
  \begin{subfigure}[b]{0.4\textwidth}
    \includegraphics[width=\textwidth, trim={0 20 0 0}, clip]{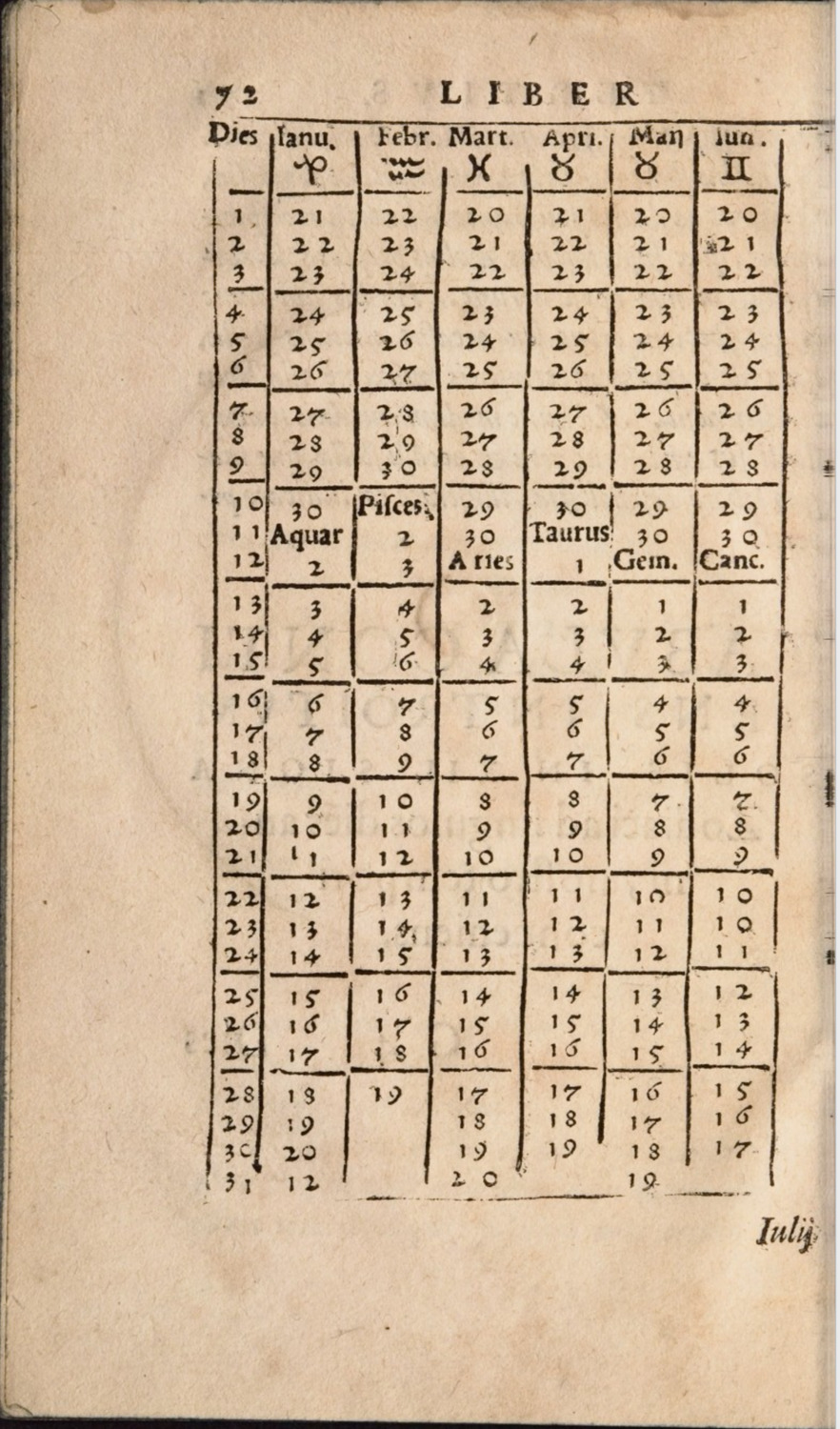}
    \caption{Left page}
    \label{fig:Sun_Zodiac_Veterum_Blebel_1582_01}
  \end{subfigure}
  \hspace{0.1cm}
  \begin{subfigure}[b]{0.4\textwidth}
    \includegraphics[width=\textwidth, trim={0 20 0 0}, clip]{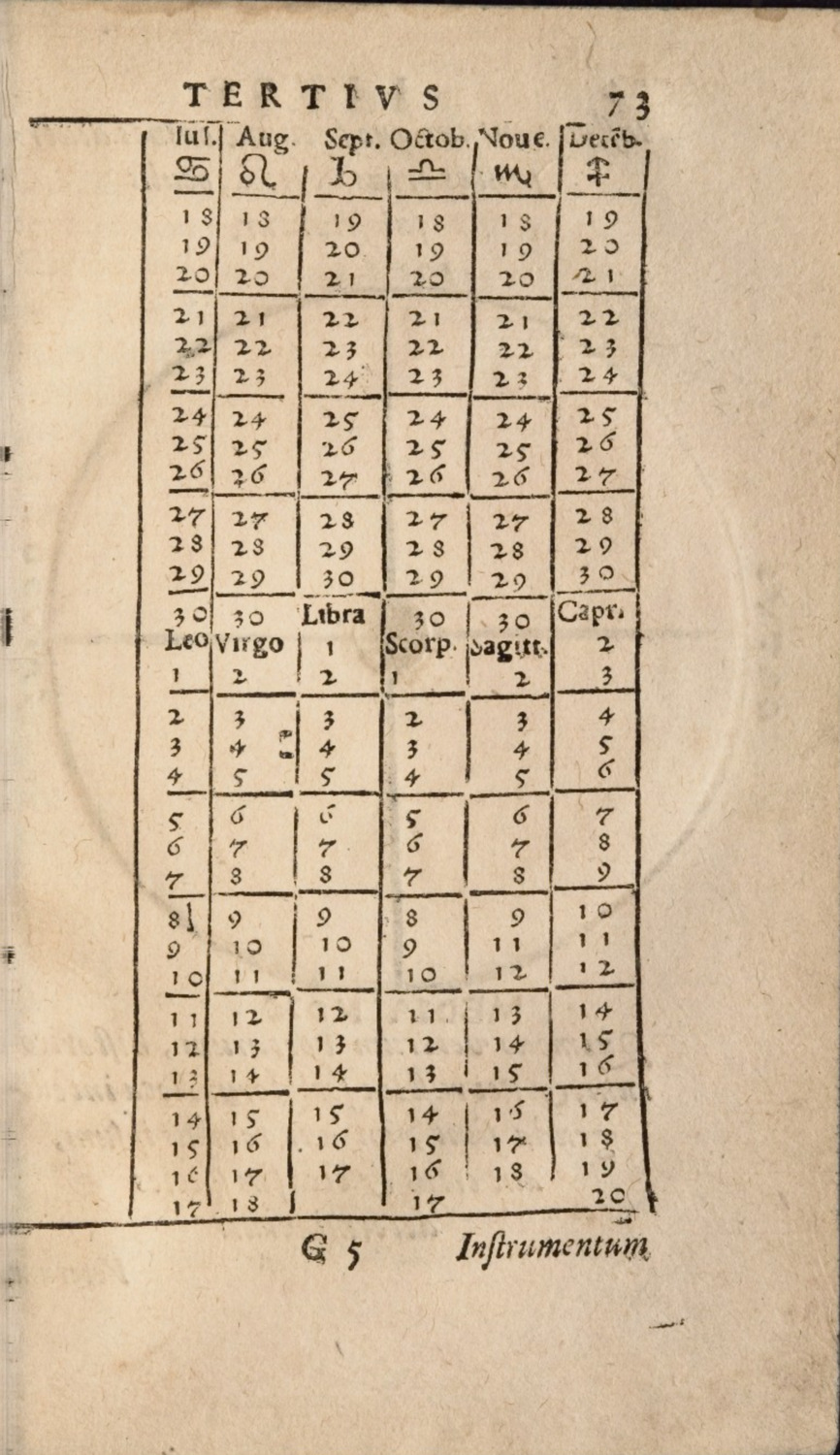}
    \caption{Right page}
    \label{fig:Sun_Zodiac_Veterum_Blebel_1582_02}
  \end{subfigure}
  \caption{\textbf{Sun-Zodiac table}  computed in relation of the temporal position of the equinoxes in ancient classical time (Edition published in 1582). From \cite[72–73]{Blebel1582}. Courtesy of the Library of the Max Planck Institute for the History of Science.}
\end{figure}

Essentially, the difference between the sixteenth-century and the ancient table amounts to a shift of the columns listing the days of the year with respect to columns giving the angular locations and thus, from the perspective of our similarity model, these two variations represent the same (more abstract) table. Examples of the two sorts of tables, taken from the same treatise published in Wittenberg in 1582 are given in Figs. \ref{fig:Sun_Zodiac_Nostro_Blebel_1582_01} and \ref{fig:Sun_Zodiac_Nostro_Blebel_1582_02} for the contemporary times and in Figs. \ref{fig:Sun_Zodiac_Veterum_Blebel_1582_01} and \ref{fig:Sun_Zodiac_Veterum_Blebel_1582_02} for the ancient classical time. Since the equinoxes drift westward along the ecliptic one degree in about 71 years, a five degree shift would correspond to a time difference of 350 years. Yet in the Julian calendar, valid before the calendar reform of 1582, dates of a fixed solar event like the equinoxes increases by three quarter of a day every century. Thus the time difference between the tables amounts to about 750 years, if modern parameters are used. This value of course depends on the value for the precision which was debated in the time in question. Indeed in the Medieval period the prevalent opinion was that the precession was not constant but changed over time, a theory known as trepidation. Ptolemy gives a value of 1/100 degrees \textit{per annum} for the precession \cite{10.2307/20488785}. Using this value a time difference for the two tables of approximately 2000 years results, which fits quite well with the distinction between the `old poets' and `our times.'\par

\vspace{\baselineskip}
\paragraph{The Spread of Zodiac Signs Tables}
\label{supp:link:sun_zodiac}
The identification of Sun-zodiac tables has been greatly facilitated by our approach. Indeed, without it, the task would be very laborious and almost impossible. Also in this case we can  visualize the spread of the tables in Europe during the early modern period.

A visualization is available at \href{http://141.5.103.115/zodiac}{http://141.5.103.115/zodiac} \protect\footnote{\texttt{username: network | password: VIZ\_network\$\_61t50}} and shows the spatio-temporal evolution of sun-zodiac tables. 
The spread is determined according to time and locations of appearance of editions containing the table. We have analyzed two variants of this table that we identified using our analysis:

\begin{itemize}
    \item[--]  Chain 1: Geo-temporal spread of the 16th century table displaying the contemporary (nostro) Sun-Zodiac table. 
    
    \item[--]  Chain 2: Geo-temporal spread of the table displaying the Sun-Zodiac table valid for ancient writers (veterum).
\end{itemize}

The dynamic of the spread of the different tables can be visualized individually or together. 
To display the metadata of the editions, each single edition can be selected on the visualization. This selection also displays the figures concerned with the speed of spread of that specific table selected. The following simulations concerns solely those editions that contain the tables discussed. This implies, however, that the visualizations do not show the spread of this historical phenomenon completely for two specific reasons. The first concerns the fact that a certain amount of treatises discuss the subject at length but furnish data in the flow of the text and could not therefore be identified by our model. The second relates instead to the preservation history of our historical sources. Especially the table for the ancient poets was often printed as a foldout bound at the end of the book. We have several cases in which we recognize that the foldout once existed but was then later torn away. For the result of the manual analysis of the treatises affected by one or both such limitations, see the data available through \cite{MVBFON2022}.\par

\begin{figure}[h!]
\centering
\includegraphics[width=.9\textwidth]{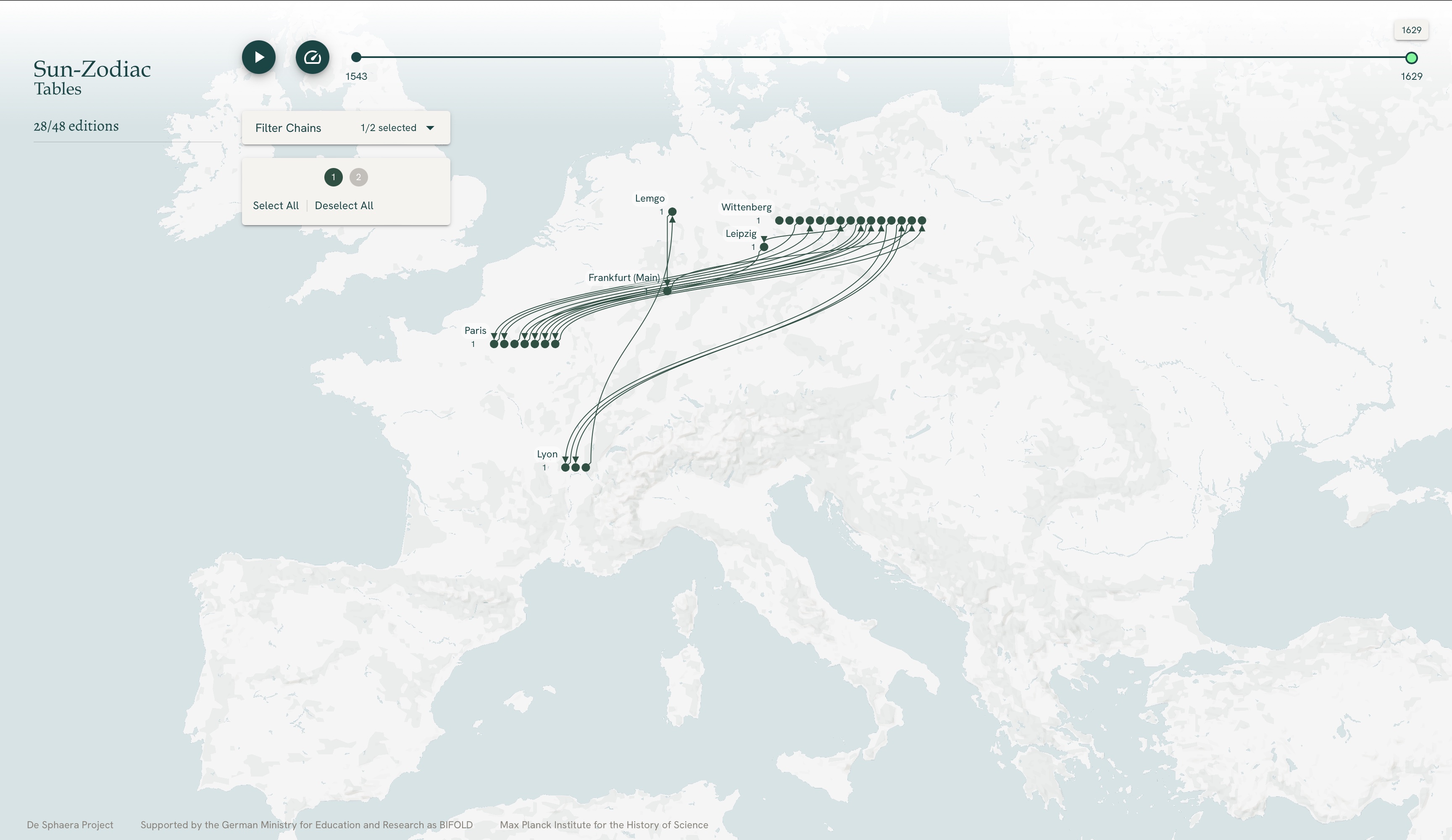}
\caption{\textbf{Spread} of the contemporary (16th cent.) Sun-Zodiac table.}
\label{fig:Spread_Contemporary}
\end{figure}

\begin{figure}[h!]
\centering
\includegraphics[width=.9\textwidth]{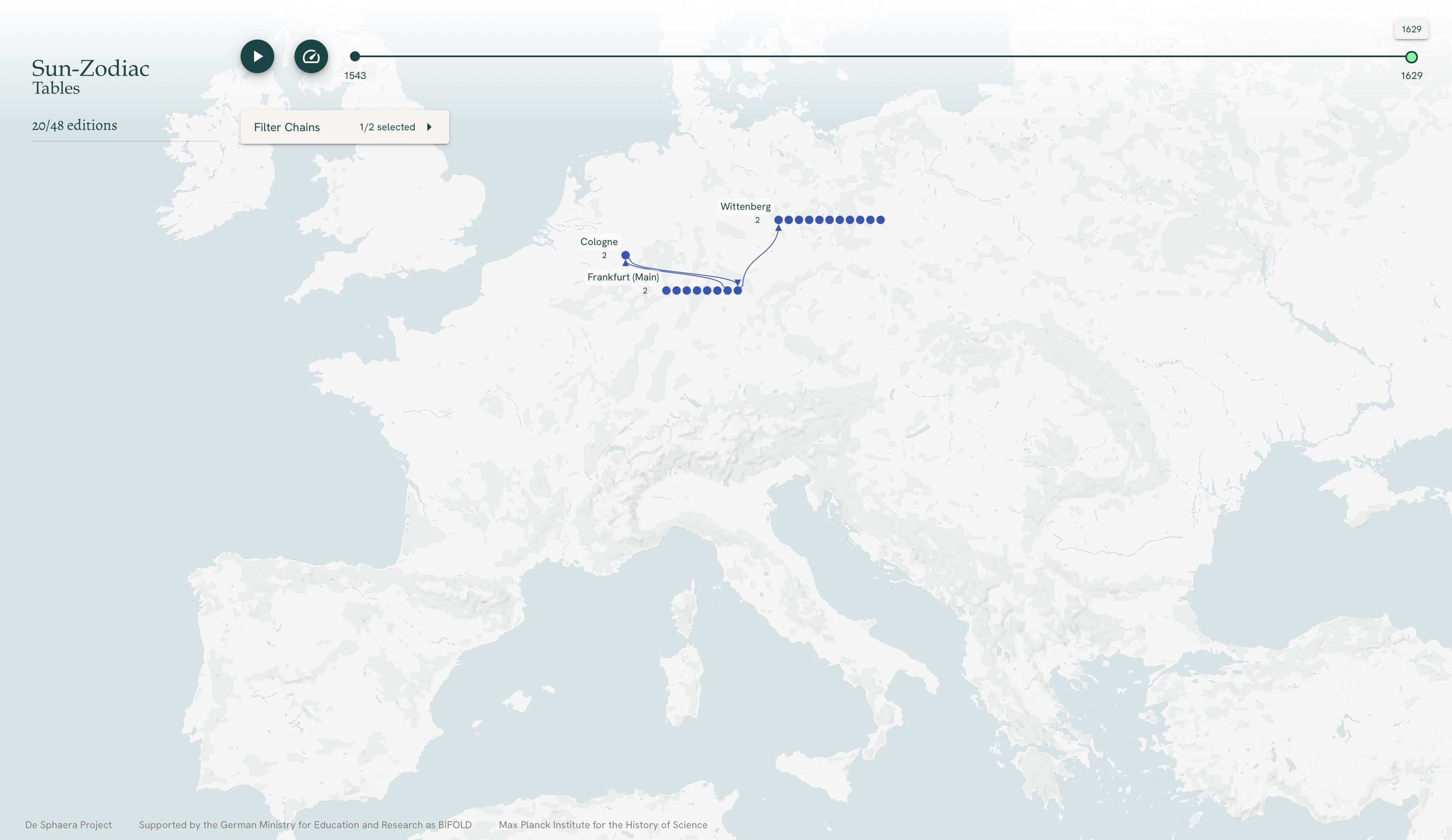}
\caption{\textbf{Spread} of the ancient Sun-Zodiac table.}
\label{fig:Spread_Ancient}
\end{figure}

In the case of the contemporary table, its diffusion begins in 1545 in Wittenberg and forty-six different editions could be identified as containing this table. The overall geographic spread remains limited to the German and French speaking regions. In total, only printers and publishers of six different cities printed this table. The two major centers of production were Wittenberg and Paris (Fig. \ref{fig:Spread_Contemporary}). The table concerning the position of the Sun in the Zodiac as observable in ancient classical time has a more limited diffusion pattern. Only twenty editions in the corpus contain it. The table appears for the first time in 1549 and was produced in only  four different locations, where Wittenberg clearly maintains the primary role. The diffusion of such table is clearly a restricted northern German phenomenon (Fig. \ref{fig:Spread_Ancient}). \par

\subsubsubsection[]{Mathematization and Identity} \label{supp:Identity}
The dataset underlying this research is obtained from the collections provided by the historical research project "The Sphere. Knowledge System Evolution and the Shared Scientific Identity of Europe." As the project's title suggests,   its primary goal  is to investigate whether and how  the development and dissemination of scientific knowledge have served as an element in shaping European identity during pre-modernity.\par

The two case studies provide an initial affirmative response to this question. The first one, concerned with the tables displaying climate zones, has shown how the journeys of exploration and therefore the discovery of an inhabited world beyond Europe and the Ecumene led to an enrichment of an existing knowledge structure, namely the division of the inhabitable zone into climate zones. This implies that the new geographic discoveries were not perceived as contradicting established scientific knowledge. Instead, they were seen as a reason for expanding the knowledge base, and from this viewpoint, they served as confirmation. This interpretation is supported by the observation that the success of the new table displaying 24 climate zones was largely dependent on the production of circulation of editions and treatises that contained not only this new table but also the old, classic one displaying the seven climate zones.\par

Especially the discovery of the "New World" (the American continent), with its diverse populations, cultures, and traditions, was particularly significant. Despite the brutality that ensued, the ability to incorporate the new discoveries in an established knowledge system must have been perceived as a confirmation tout court of the science that, at that time, was already  seen as the result of collaborative efforts  that were taking place at a continental level in Europe. Although this incorporation came at the expense of other cultures, it likely played a role in shaping science as a factor in identity formation.  This idea appears all the more plausible if it is considered that the most relevant identity-shaping factor of Europe had previously been  religion, particularly through the external representation of authority embodied by the Pope and the Holy See of Rome. In the period considered in this research, the unity of Europe could however no longer be guaranteed by religion due to the fragmentation of the church and the consequent political and military conflicts. On the basis of our results, we therefore propose the working hypothesis that while encountering "otherness," science began to emerge as a new identity-shaping cultural aspect.\par

The second case study further supports this hypothesis. If science plays a role in shaping identity, its roots need to be investigated as well. During that era,  these roots were clearly identified  with the philosophical and scientific cultures of classic antiquity, particularly those of the Greeks and Romans.  It was one of the most important fathers of the Protestant Reformation, Philipp Melanchthon, who  in 1531 and 1538 urged the youth to  study  astronomy, stating that without it,  history would merely be a chaotic collection of unordered pieces of information \cite{MVBFON2022,RN1480}. Melanchthon published this call in form of an open letter used as preface to an astronomy textbook. This text soon became  the most republished text-part of the entire collection examined in this research, circulating widely across Europe independently from the religion of the countries where the editions containing this text were printed and distributed.\par 

The significance of this call was twofold. On one hand it underscored the importance of understanding the history of Western civilization, a crucial step in any process of identity formation. On the other hand, it further contributed to the process of diffusion of mathematical culture as it was mathematical astronomy that was called upon to support history writing. Dating past events by means of astronomic calculations and investigations became a specialized field, creating a self-reinforcing loop. This process not only promoted scientific development but also positioned science as an identity-shaping factor.\par

Finally, it is worth considering that the processes unveiled  by this research both on corpus level and at the level of case studies was taking place in what was becoming the most relevant educational institutional setting -- the universities.   In conclusion, we would like to highlight that this path has paved the way to the formulation of an important new working hypothesis within the realm of political epistemology of science: the relationship between identity and domination as pivoted around scientific knowledge.\par 

\end{document}